\documentclass{egpubl}

\JournalSubmission    % uncomment for submission to Computer Graphics Forum
\usepackage[T1]{fontenc}
\usepackage{dfadobe}
\usepackage{amssymb}
\usepackage{amsmath}
\usepackage{algorithm}
\usepackage{algorithmic}
\usepackage{booktabs}
\usepackage{svg}
\usepackage{subfig}
\usepackage{placeins}
\usepackage{threeparttable}
\usepackage[commandnameprefix=always,final]{changes}
\usepackage{float}
\usepackage{bbding}
\usepackage[strings]{underscore}
\usepackage{makecell}
\usepackage[labelfont=bf]{caption}
\DeclareMathOperator*{\argmax}{arg\,max}
\DeclareMathOperator*{\argmin}{arg\,min}
\setaddedmarkup{\color{blue}{#1}}
\setdeletedmarkup{\color{red}{\sout{#1}}}
\newcommand{\add}[1]{\chadded{#1}}
\newcommand{\remove}[1]{\chdeleted{#1}}
\newcommand{\replace}[2]{\chreplaced{#2}{#1}}

\usepackage{cite}  % comment out for biblatex with backend=biber
\BibtexOrBiblatex
\electronicVersion
\PrintedOrElectronic
\usepackage{egweblnk}
\title[GNNRL-Smoothing: A Prior-Free Reinforcement Learning Model for Mesh Smoothing]%
{GNNRL-Smoothing: A Prior-Free Reinforcement Learning Model for Mesh Smoothing}

\author[Z.\,C. Wang \& X.\,C.  Chen \& C.\,Y. Gong et al.]
{\parbox{\textwidth}{\centering Z.\,C. Wang$^{1,2}$,  X.\,C. Chen$^{1,2}$\thanks{Corresponding author}, C.\,Y. Gong$^{1,2}$, B. Yang$^{1,2}$, L. Deng$^{3}$, Y.\,F. Sun$^{4}$, Y. F. Pang$^{3}$ and J. Liu$^{1,2}$
}
\\
{\parbox{\textwidth}{\centering $^1$National University of Defense Technology, Science and Technology on Parallel and Distributed Processing Laboratory, China\\
$^2$National University of Defense Technology, Laboratory of Digitizing Software for Frontier Equipment, China \\
$^3$China Aerodynamics Research and Development Center, China \\
$^4$Nankai University, College of Software,  China
}
}
}
\begin{document}

% uncomment for using teaser
% \teaser{
%  \includegraphics[width=0.9\linewidth]{eg_new}
%  \centering
%   \caption{New EG Logo}
% \label{fig:teaser}
%}

\maketitle
%-------------------------------------------------------------------------
\begin{abstract}
Mesh smoothing methods can enhance mesh quality by eliminating distorted elements, leading to improved convergence in simulations. To balance the efficiency and robustness of traditional mesh smoothing process, previous approaches have employed supervised learning and reinforcement learning  to train intelligent smoothing models. 
However, these methods heavily rely on labeled dataset or prior knowledge to guide the models' learning. Furthermore, their limited capacity to enhance mesh connectivity often restricts the effectiveness of smoothing.
In this paper, we first systematically analyze the learning mechanisms of recent intelligent smoothing methods  and propose a prior-free reinforcement learning model for intelligent mesh smoothing. Our proposed model integrates graph neural networks with reinforcement learning to implement an intelligent node smoothing agent and introduces, for the first time, a mesh connectivity improvement agent. We formalize mesh optimization as a Markov Decision Process and successfully train both agents using Twin Delayed Deep Deterministic Policy Gradient and Double Dueling Deep Q-Network in the absence of any prior data or knowledge. 
We verified the proposed model on both 2D and 3D meshes. Experimental results demonstrate that our model achieves feature-preserving smoothing on complex 3D surface meshes. It also achieves state-of-the-art results among intelligent smoothing methods on 2D meshes and is 7.16 times faster than traditional optimization-based smoothing methods. Moreover, the connectivity improvement agent can effectively enhance the quality distribution of the mesh.
%-------------------------------------------------------------------------
%  ACM CCS 1998
%  (see https://www.acm.org/publications/computing-classification-system/1998)
% \begin{classification} % according to https://www.acm.org/publications/computing-classification-system/1998
% \CCScat{Computer Graphics}{I.3.3}{Picture/Image Generation}{Line and curve generation}
% \end{classification}
%-------------------------------------------------------------------------
%  ACM CCS 2012
%The tool at \url{http://dl.acm.org/ccs.cfm} can be used to generate
% CCS codes.
%Example:


\end{abstract}
%-------------------------------------------------------------------------

\section{Introduction}\label{sec:intro}
 Numerical simulation  is commonly employed to simulate and analyze fluid dynamics phenomena. It finds widespread applications in various fields such as automotive engineering, aerospace, biomedicine, and nuclear physics \cite{yadigarogluComputationalFluidDynamics2005, spalartRoleChallengesCFD2016, faizalComputationalFluidDynamics2020, aultmanEvaluationCFDMethodologies2022}.
In mesh-based  numerical simulation methods, mesh generation is a pre-processing step preceding numerical simulations
\cite{bakerMeshGenerationArt2005}. This step involves partitioning the computational domain into non-overlapping geometric elements, such as 2D polygons or 3D polyhedra. Mesh quality, characterized by orthogonality, smoothness, and density distribution, has a significant impact on simulation errors, stability, and efficiency \cite{knuppAlgebraicMeshQuality2001,knuppRemarksMeshQuality2007,katzMeshQualityEffects2011}. Low-quality mesh elements can increase the stiffness of the matrix, leading to divergence in the simulation process. However, despite the advancements in automated mesh generation techniques, the initially generated mesh quality may still fail to meet the requirements of simulation, especially for complex geometries. To address this issue, mesh optimization methods are widely utilized to improve mesh quality \add{in the pre-processing step},  such as smoothing \cite{parthasarathyConstrainedOptimizationApproach1991, freitagCombiningLaplacianOptimizationbased1997, zhouAngleBasedApproachTwoDimensional2000, jonesNoniterativeFeaturepreservingMesh2003}, edges or faces flipping \cite{freitagTetrahedralMeshImprovement1997a, prasadComparativeStudyMesh2018a}, and \replace{node insertion/deletion}{mesh refinement} \cite{freitagTetrahedralMeshImprovement1997a,prasadComparativeStudyMesh2018a,guoAdaptiveSurfaceMesh2021}, among others.

\begin{figure}[tbp]
\centering
\includegraphics[width=0.95\columnwidth]{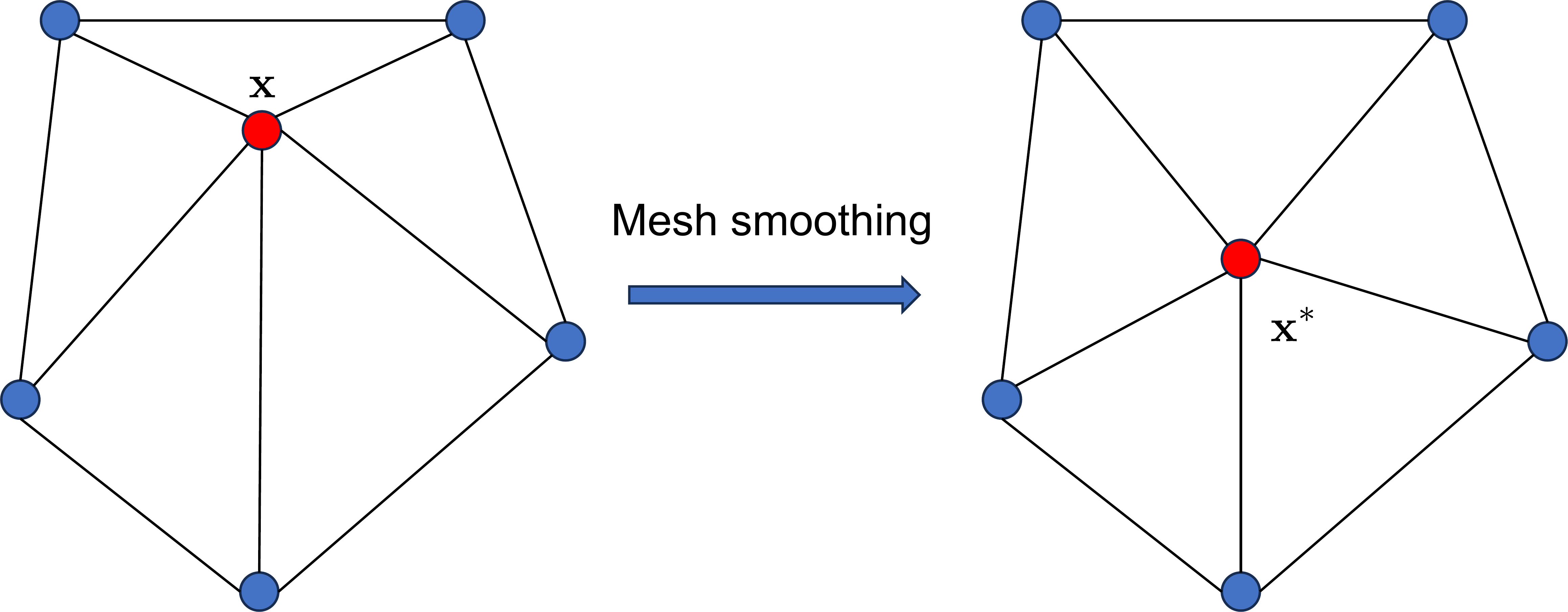}
\caption{\textit{A illustration of the mesh smoothing method. By adjusting the positions of central nodes from $\mathbf{x}$ to $\mathbf{x}^*$, the quality of adjacent mesh elements can been improved.}}
\label{fig:introsmoothing}
\end{figure}

Improving mesh quality during the pre-processing step is an effective way to ensure successful numerical simulations. Compared to adaptive mesh refinement \cite{wallworkE2NErrorEstimation2022, yangReinforcementLearningAdaptive2023, yangMultiAgentReinforcementLearning2023, zhangUniversalMeshMovement2024}, which dynamically adjusts mesh density during simulations, mesh smoothing is a more efficient and straightforward method to accelerate convergence.
The smoothing methods enhance the quality of mesh elements by relocating the positions of mesh nodes while preserving mesh \replace{topology}{connectivity}, which is shown in Figure~\ref{fig:introsmoothing}.  Traditional mesh smoothing methods can be broadly categorized into two main types: heuristic-based and optimization-based methods. Heuristic-based methods adjust the positions of mesh nodes using manually designed heuristic functions. Laplacian smoothing and its variants are the most representative methods in this category \cite{fieldLaplacianSmoothingDelaunay1988a, vollmerImprovedLaplacianSmoothing1999, freitagCombiningLaplacianOptimizationbased1997,loFiniteElementMesh2014}. Although heuristic methods are simple, their smoothing performances are limited due to their reliance on manual heuristic function design, which may lead to the introduction of inverted elements. To improve the robustness of smoothing process, optimization-based methods compute new mesh node locations by solving a constrained optimization problem \cite{parthasarathyConstrainedOptimizationApproach1991, canannApproachCombinedLaplacian1998a, freitagLocalOptimizationbasedSimplicial2000a, gargallo-peiroSurfaceMeshSmoothing2014a}. They generally offer superior mesh quality compared to heuristic-based methods. Nevertheless, implementing these methods requires defining an objective function (often derived from mesh quality metrics), optimization algorithms, and stopping criteria, which requires a substantially greater effort and computational resource. Moreover, the quality of optimization is often subject to the choice of the objective function \cite{chenConstructionObjectiveFunction2004, daiComparisonObjectiveFunctions2014}. As mentioned above, balancing the robustness and efficiency of smoothing through intelligent methods is a worthwhile problem to address.

Recently, Artificial Intelligence (AI) methods have been introduced into mesh-related fields to achieve efficient and high-quality mesh processing,  such as mesh generation \cite{zhangMeshingNetNewMesh2020a, daroyaREINFlexibleMesh2020,  zhangMeshingNet3DEfficientGeneration2021a,  papagiannopoulosHowTeachNeural2021b, chenMGNetNovelDifferential2022}, mesh quality evaluation \cite{chenDevelopingNewMesh2020, chenMVENetAutomatic3D2021, wangEvaluatingMeshQuality2022a}, and mesh optimization \cite{bohnRecurrentNeuralNetworks2021, paszynskiDeepLearningDriven2021, tingfanMeshOptimizationMethod2022a}. A great deal of works have also utilized AI methods to enhance the efficiency of optimization-based smoothing methods \cite{guoNewMeshSmoothing2022a, wangUnstructuredSurfaceMesh2023a, wangProposingIntelligentMesh2023}. NN-Smoothing  \cite{guoNewMeshSmoothing2022a} first employed supervised learning and multilayer perceptron  (MLP) to predict the optimal positions of mesh nodes for smoothing. Despite achieving satisfactory smoothing results on both 2D and 3D surface meshes, its main drawbacks lie in the requirement for prepared high-quality mesh data and limited scalability, and the inability to handle mesh nodes with different degrees. GMSNet \cite{wangProposingIntelligentMesh2023}  addressed the data dependency issue of the NN-Smoothing model by introducing an unsupervised loss function, which avoids reliance on labeled high-quality meshes. It has also realized a lightweight mesh smoothing model, addressing the scalability issues of NN-Smoothing. However, it introduces unverified prior assumptions into the smoothing process, which leads to an inferior performance. Additionally, the method  was only validated on 2D meshes, limiting further applications. DRL-Smoothing \cite{wangUnstructuredSurfaceMesh2023a} combined traditional mesh smoothing methods with reinforcement learning (RL) algorithms, DDPG \cite{lillicrapContinuousControlDeep2019a}, to achieve high-quality mesh smoothing, demonstrating its effectiveness in practical simulations. Despite its effectiveness,  the performance of the method is largely derived from traditional smoothing methods. Specifically, it requires the output of traditional smoothing methods as part of the results for training and inference, creating a strong dependency on traditional methods.
In summary, the aforementioned intelligent smoothing methods heavily rely on prior data (\add{high-quality} mesh datasets) or knowledge (prior assumptions or prior knowledge from traditional methods). Additionally, they focus solely on adjusting node positions and neglect mesh \replace{topology}{connectivity} adjustments, making them difficult to handle low-quality meshes with poor topological structures.

To address the aforementioned challenges, we systematically analyze existing intelligent smoothing models from the reinforcement learning perspective, revealing their dependency on prior data or knowledge through theoretical analysis or experiment. Building upon this, we further integrate GNNs with RL, and introduce a priori-free model GNNRL-Smoothing.
In the proposed model, a continuous-action agent and a discrete-action agent are coupled to perform node smoothing and topological structure improvement, respectively. We successfully applied reinforcement learning to mesh optimization tasks, defining the elements of a reinforcement learning-based mesh optimization problem, and  training the two agents using the Twin Delayed Deep Deterministic Policy (TD3)  \cite{fujimotoAddressingFunctionApproximation2018a} and the Double Dueling Deep Q-Network (D3QN) \cite{wangDuelingNetworkArchitectures2016, vanhasseltDeepReinforcementLearning2016}. Compared to previous intelligent smoothing models, GNNRL-Smoothing does not need supervised data from high-quality meshes, nor does it depend on traditional methods as part of its output.
We tested the proposed model on both 2D and 3D meshes. The experimental results indicate that our model successfully preserves features while smoothing complex 3D surface meshes. Additionally, it achieves state-of-the-art performance among intelligent smoothing methods for 2D meshes and is 7.16 times faster than conventional optimization-based smoothing techniques. Furthermore, the connectivity improvement agent significantly enhances the quality distribution of the mesh.

The remainder of this paper is structured as follows. Section~\ref{sec:rw} provides a concise overview of traditional mesh smoothing methods and offers a reinforcement learning perspective on recent intelligent mesh smoothing methods, highlighting their limitations and dependencies on prior data or knowledge. Section~\ref{sec:method}   presents a detailed description of the proposed GNNRL-Smoothing model, including the design of the mesh node smoothing agent and connectivity improvement agent, as well as the reinforcement learning framework. In Section~\ref{sec:exp}, we conduct comprehensive experiments on both 2D meshes and 3D surface meshes, including ablation studies on the model's design. Finally, we conclude our work, discuss current limitations of the proposed methods, and outline potential directions for future research.

%-------------------------------------------------------------------------

\section{Related works}\label{sec:rw}

\subsection{Traditional mesh smoothing methods}
\textbf{Laplacian smoothing.}
Laplacian smoothing and its variants \cite{fieldLaplacianSmoothingDelaunay1988a, vollmerImprovedLaplacianSmoothing1999, freitagCombiningLaplacianOptimizationbased1997,loFiniteElementMesh2014} are the most representative methods for mesh smoothing. The simplest version achieves mesh smoothing by computing the algebraic average of adjacent nodes. Let \(\boldsymbol{x}\) denote the coordinates of the mesh node to be smoothed, \(\mathbf{x}_{i}\) be the adjacent mesh nodes, and \(N\) be the number of adjacent nodes. The smoothed position of the mesh node, \(\mathbf{x}^{*}\), can be expressed as:
\begin{equation}
\mathbf{x}^{*}=\frac{1}{N} \sum_{i=1}^{N} \mathbf{x}_{i} .
\end{equation}
Despite its simplicity and effectiveness, its drawback lies in the potential generation of inverted elements when the local region around the mesh node is non-convex. Other methods, such as area- or volume-weighted Laplacian smoothing \cite{vollmerImprovedLaplacianSmoothing1999}, Smart Laplacian smoothing \cite{freitagCombiningLaplacianOptimizationbased1997}, and Angle-based smoothing \cite{zhouAngleBasedApproachTwoDimensional2000}, have been proposed to mitigate this issue. Area- or volume-weighted Laplacian smoothing gives additional consideration to the size or volume of the adjacent elements. This approach achieves a more balanced and effective redistribution of node positions by computing the geometric characteristics of adjacent elements. This contributes to improved mesh quality and helps avoid the generation of inverted elements. Smart Laplacian smoothing employs the same node update method as vanilla Laplacian smoothing but with an additional check before node updates. If the node movement results in lower mesh quality or invalid elements, such movements are canceled. Angle-based smoothing focuses on adjusting the angles associated with the nodes of these elements, which also helps prevent the creation of inverted elements.

\textbf{Geometric element transformation method.}
Unlike individual mesh node movement, the Geometric element transformation method (GETMe) \cite{vartziotisMeshSmoothingUsing2008a, vartziotisImprovedGETMeAdaptive2017a, vartziotisGetmeMeshSmoothing2018} performs mesh smoothing by transforming the shape of mesh elements. Let \(\mathbf{z} = \left(z_{0}, \ldots, z_{n-1}\right)^{\mathrm{t}} \in \mathbb{C}^{n}\) denote node coordinates of the mesh element, where \(n\) is the number of nodes in the element, \(z_k = x_k + iy_k \in \mathbb{C}\) represents the coordinates of node \(k\)  \add{ in complex field}. It can be proven that through an iterative transformation  $ \mathbf{M} $, where \(\mathbf{z}^{i+1} = \mathbf{M}\mathbf{z}^{i}\), the mesh element gradually approaches its regular counterparts. This transformation not only enhances mesh quality but also preserves good geometric properties, such as maintaining the centroid of the polygon. However, this method has a drawback: the transformation can cause adjacent elements to become invalid due to changes in mesh element size. Scaling and post-check adjustments are needed to ensure the validity of the elements. Additional techniques are also required to ensure the effectiveness of the method during the smoothing process, making the implementation of this method relatively complex.

\textbf{Optimization-based smoothing.}
Optimization-based smoothing methods formalize the mesh smoothing problem as a constrained optimization problem. They then apply standard optimization techniques such as Newton's algorithm, conjugate gradient approach, and gradient descent method \cite{boydConvexOptimization2004} to achieve mesh smoothing. When implementing this method, it is necessary to define mesh quality metrics, objective functions, and solving algorithms. Commonly used mesh quality metrics include maximum and minimum angles, aspect ratio, distort ratio, and others \cite{parthasarathyConstrainedOptimizationApproach1991, xuHexahedralMeshQuality2018}. The objective function is typically constructed from the mesh quality metrics, with the goal of maximizing or minimizing it. While explicit solutions are possible when the objective function is simple, iterative solutions are required in most cases. Despite being able to achieve higher mesh smoothing quality, optimization-based methods often have lower efficiency compared to heuristic smoothing methods. Moreover, their smoothing effectiveness depends on the construction and solution of the optimization problem.

\subsection{Analyze of the intelligent smoothing method}\label{subsec:review}
In this section, we summarize and analyze the learning mechanisms of current intelligent smoothing methods from the perspective of reinforcement learning, revealing their reliance on prior data or knowledge. We also highlight the limitations of these methods in the field of mesh smoothing. \\
\textbf{Reinforcement learning.}
Reinforcement Learning is a machine learning paradigm that involves an agent learning to make decisions through interactions with an environment. The agent receives feedback in the form of rewards, guiding it to learn optimal behavior through trial and error  \cite{wieringReinforcementLearningStateoftheArt2012, liDeepReinforcementLearning2018}. In reinforcement learning, the agent observes the state \(\mathbf{s}_t\) from the environment at time step \(t\), decides on an action \(a_t\) based on its policy \(\pi_\mathbf{\theta}\) with parameters $\mathbf{\theta}$. The environment transitions to the next state \(\mathbf{s}_{t+1}\) according to the current state and the agent's action, and provides the corresponding reward \(r_t\) to the agent. A transition in reinforcement learning is typically denoted as a tuple \( \left( \mathbf{s}_t, \mathbf{a}_t, r_t, \mathbf{s}_{t+1}\right)  \), which represents the state, action, reward, and next state at time step $t$. An episode is a sequence of transitions, denoted as
\( \mathbf{\tau}=\left(\mathbf{s}_{0}, \mathbf{a}_{0}, r_{0}, \mathbf{s}_{1}, \mathbf{a}_{1}, r_{1}, \ldots, \mathbf{s}_{T-1}, \mathbf{a}_{T-1}, r_{T-1}, \mathbf{s}_{T}\right) \), from the initial to the final time step $T$. The aforementioned components constitute a \textit{Markov Decision Process} (MDP), which serves as the foundation of reinforcement learning. The agent's objective is to learn a policy that maximizes the cumulative reward (return) $R(\tau)=\sum_{i=0}^{T} \gamma^tr_t$, where  $\gamma \in \left( 0,1 \right) $ is the discount factor that determines the importance of long-term rewards.

\begin{figure}[tbp]
\centering
\includegraphics[width=0.75\linewidth]{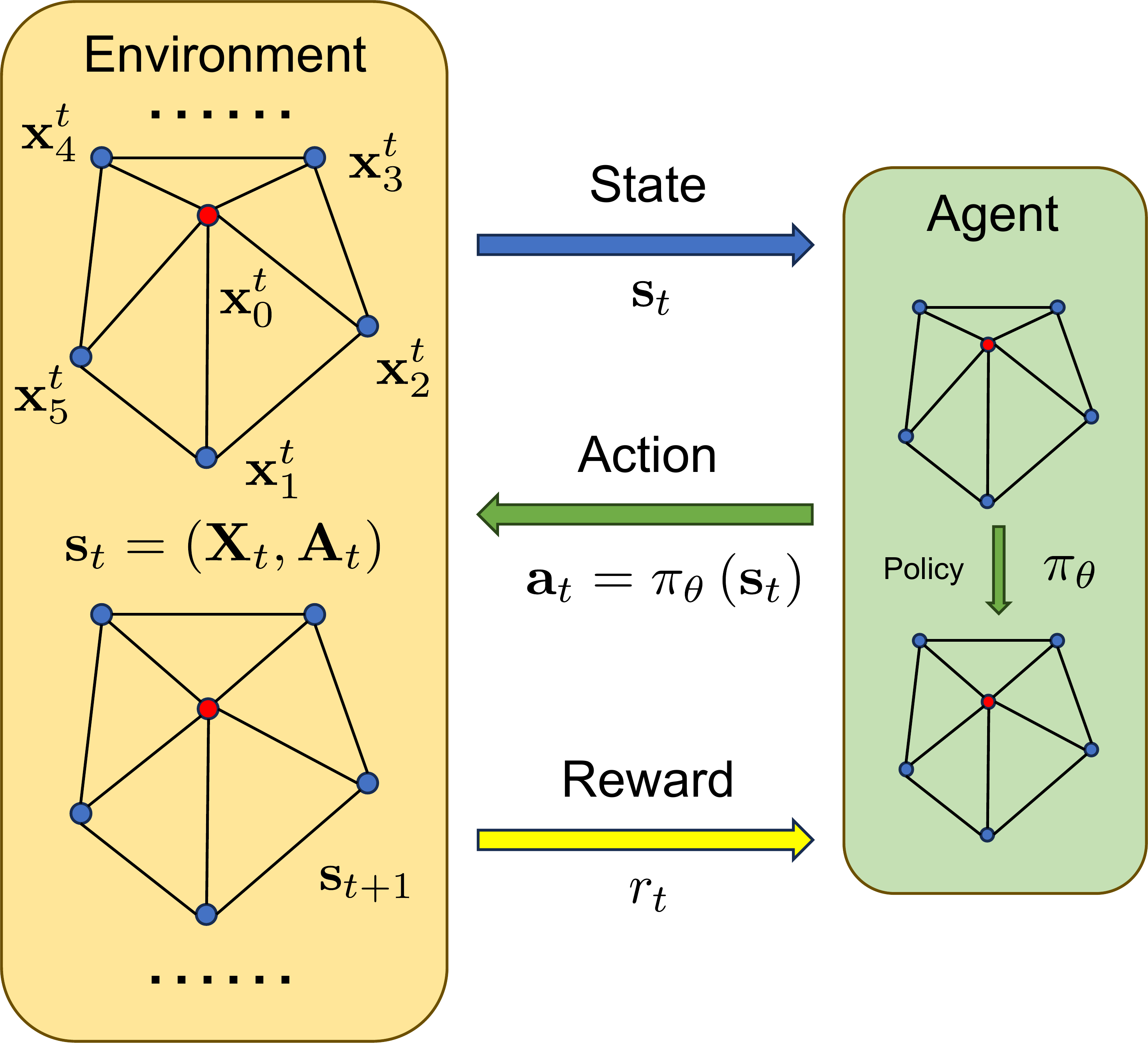}
\caption{\textit{Definition of reinforcement learning concepts in mesh smoothing task. At time step $t$, given a mesh node (red node) and its neighboring nodes (blue nodes), the agent of smoothing method observes the state $\mathbf{s}_t$, provides the action $\mathbf{a}_t$ to update the position of the mesh node, and receives the corresponding reward $r_t$. The corresponding mesh state is then transformed into \( \mathbf{s}_{t+1} \).}}
\label{fig:RLDefine}
\end{figure}
To provide a summary of recent intelligent smoothing methods from the perspective of reinforcement learning, we first \replace{present}{provide } the following definitions within the context of mesh smoothing and reinforcement learning, as illustrated in Figure~\ref{fig:RLDefine}:
\begin{itemize}
\item \textbf{Environment}: Given a mesh node, the environment consists of the node itself and \( N \) neighboring nodes.
\item \textbf{State}: Environment state at time $t$ is defined as $\mathbf{s}_t=\left(\mathbf{X}_t, \mathbf{A}_t \right) $, where $\mathbf{X}_t=\left[  \mathbf{x}_0^t, \mathbf{x}_1^t, \mathbf{x}_2^t, \cdots , \mathbf{x}_{N-1}^t,  \mathbf{x}_{N}^t \right] ^T  \in \mathbb{R}^{(N+1) \times 2}$ represents mesh nodes' position,  $\mathbf{A}_t=\left[a_{ij}^t\right] \in \mathbb{R}^{(N+1) \times (N+1)}$ represents the connection among nodes, \add{and time \( t \) is defined as the number of iterations smoothing the mesh node}. Each row in $\mathbf{X}_t$ represents the coordinate of mesh node and its neighbors, and $a_{ij}^t$ denotes whether node $i$ is connected with node $j$. It's worth noting that the order of nodes in $ \mathbf{X}_t $ is arbitrary, and it can be transformed into each other through permutation operations. The node positions here are regularized using the normalization method employed in previous works, constraining the coordinates of the nodes within a fixed range (such as \( [0,1]  \)  ) \cite{guoNewMeshSmoothing2022a, wangUnstructuredSurfaceMesh2023a, wangProposingIntelligentMesh2023}.
\item \textbf{Action}: The action $\mathbf{a}_{t}=\pi_{\theta}\left( \mathbf{s}_{t} \right)$ is the movement of the mesh node to achieve smoothing.
\item  \textbf{Reward}: The reward $r_t$ is defined as some metric related to mesh quality. It may involve traditional mesh quality metrics or the validation of the mesh elements.
\end{itemize}
These definitions are similar to those in DRL-Smoothing. However, our environment is defined on an individual node rather than all mesh nodes, and the state of the environment further includes the connectivity between mesh nodes. We will further discuss the differences between the two in Section \ref{sec:modeltraining}. Based on the above definitions, we elucidate the various forms of dependency inherent in current intelligent methods: dependency on prior data, dependency on prior assumptions, and dependency on prior knowledge. The analysis of these three forms of dependency is detailed as follows.

\textbf{Prior data dependency.}
In our framework, the dependency on prior data refers to relying on supervised high-quality meshes for model training.
NN-Smoothing learns how to output the optimal positions for smooth nodes by \textit{imitating} the behavior of optimization-based smoothing. From the perspective of reinforcement learning, this method can be viewed as a type of \textit{Behavior Cloning} in imitation learning \cite{husseinImitationLearningSurvey2018}. Behavior cloning in reinforcement learning involves training an agent to mimic the behavior of an expert by learning a mapping from observations to actions. In NN-Smoothing, the expert refers to optimization-based smoothing, and the dataset \(\mathcal{D}=\left\{\left(\mathbf{s}^i, \mathbf{a}^i \right) \right\}_{i=1}^{N_d}\) consists of the pre-smoothed mesh state \(\mathbf{s}^i\) and the corresponding action \(\mathbf{a}^i\) ( $i$ represents data index rather than a time step, and $N_d$ is the number of data samples). Here, the episode length is 1, which means the method directly outputs the optimal smoothing positions. It is evident that this method relies on prior data, which makes it face challenges: on the one hand, training data must be obtained through an optimization method, and on the other hand, the inherent solving errors of the optimization method introduce bias into the trained model.

\textbf{Prior assumption dependency.}
Dependency on prior assumptions refers to the use of simplified models during the learning process, limiting the expressive power of the model. Here, we will show that the GMSNet is essentially a simplified form of \textit{policy gradient method} \cite{petersPolicyGradientMethods2010, silverDeterministicPolicyGradient2014}.
GMSNet introduced an unsupervised loss for model training. Let $\mathbf{s}$ and $\mathbf{s}'$ be the states of the mesh node before and after smoothing, respectively,  $q_i$ represents the aspect ratio of the adjacent mesh element $i$ to the smoothed mesh node, and $\mathbf{W}$ is the parameter of GMSNet model. The proposed unsupervised loss can be expressed as:
\begin{equation}
\label{eq:GMSNetLoss}
\mathcal{L}\left(\text{GMSNet}_{\mathbf{W}} \left( \mathbf{s} \right) \right)=\mathcal{L}(\mathbf{s}')=\frac{1}{N} \sum_{i=1}^{N}\left(1-\frac{1}{q_{i}}\right),
\end{equation}
which is only dependent on the node state after smoothing. The training process of GMSNet can be formulated as solving the following optimization problem:
\begin{equation}
\begin{aligned}
    \mathbf{W}^{*}&=\arg \min _{\mathbf{W}} \mathrm{E}_{\mathbf{s} \sim \mathcal{M}}\left(\mathcal{L}\left(\mathbf{s}' \right)\right) \\
    &=\arg \min _{\mathbf{W}} \sum_{\mathbf{s}} \mathcal{L}\left(\mathbf{s}' \right) p\left( \mathbf{s} \right)  ,
\end{aligned}
\end{equation}
where $\mathcal{M}$ is node state distribution on mesh. Further details about GMSNet training are provided in the Appendix \ref{app:GMSNet}.

Policy gradient method aims to find the optimal policy $\pi^*$ by maximizing the expection of the cumulative rewards (expected return), which can be formalized as:
\begin{equation}
\label{eq:PG}
\begin{aligned}
    \pi^* & =\argmax_{\pi_\theta}\mathrm{E}_{\mathbf{\tau} \sim \pi_\theta} \left( R(\tau) \right) \\
    & =\argmax_{\pi_\theta} \sum_\tau R(\tau)p(\tau| \theta)\\
    & = \argmax_{\pi_\theta} \sum_\tau \sum_{t=0}^{T-1} \gamma^t r_t p(\tau| \theta),
\end{aligned}
\end{equation}
where $p(\tau| \theta)$ is the probability of the episode $\tau$ occurring given the model parameter $\theta$. Let $r_t=-\mathcal{L}(s_{t+1})$,  episode length $T=1$, and assume episode distribution is independent on policy parameter $\theta$, we can rewrite Equation \ref{eq:PG} as:
\begin{equation}
\begin{aligned}
    \pi^* & = \argmax_{\pi_\theta} \sum_\tau \sum_{t=0}^{T-1} \gamma^t r_t p(\tau| \theta) \\
    & = \argmax_{\pi_\theta}  \sum_{\tau=\left (\mathbf{s},\mathbf{a},\mathbf{s}' \right)} - \mathcal{L}(\mathbf{s}') p(\mathbf{\tau}| \mathbf{\theta})\\
    & = \argmax_{\pi_\theta} - \sum_{\mathbf{s}} \mathcal{L}(\mathbf{s}') p(\mathbf{s})\\
    & = \argmin_{\pi_\theta}  \sum_{\mathbf{s}} \mathcal{L}(\mathbf{s}') p(\mathbf{s}),
\end{aligned}
\end{equation}
which is the same as GMSNet's optimization problem. It can be seen that GMSNet introduces two prior assumptions in policy gradient: episode length assumption, $T=1$,  and the assumption of independent node state distribution. However, these priors limit the expressive power of the model and have yet to be validated for their effectiveness.

\textbf{Prior knowledge dependency.}
\begin{figure*}[tbp]
\centering
\subfloat[]{\includegraphics[width=0.96\columnwidth]{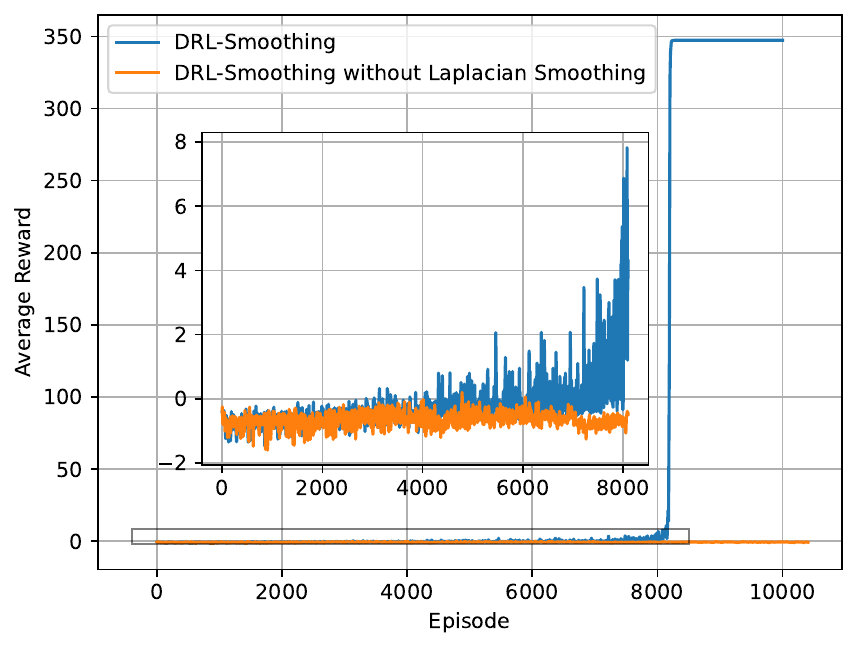}}\hspace{5pt}
\subfloat[]{\includegraphics[width=0.96\columnwidth]{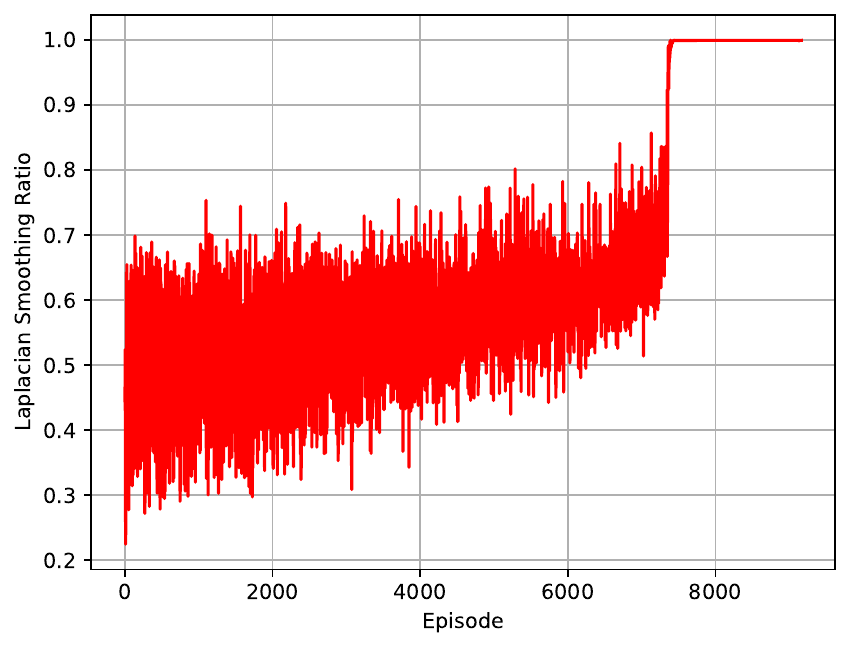}}\\
\caption{\textit{\add{The drawbacks of DRL-Smoothing.} The DRL-Smoothing models were trained with and without Laplacian smoothing. It can be observed that including Laplacian smoothing as part of the output allows convergence at around 8000 steps, while excluding it leads to non-convergence. When Laplacian smoothing is included, the majority of smoothing actions are determined by the traditional method, with the intelligent model's action contribution being relatively small (only 0.05\%).}}
\label{fig:RWDRLwoLap}
\end{figure*}
Dependency on prior knowledge refers to the reliance on knowledge from traditional methods to achieve intelligent smoothing. Such dependency can be seen in DRL-Smoothing, which combines traditional smoothing methods with intelligent smoothing methods. Specifically, denoting \(\textbf{x}_{\text{trans}}\) as the node positions output by traditional smoothing methods and \(\textbf{x}_{\text{ai}}\) as the node positions output by intelligent methods, the final node positions are obtained by summing the two: \(\textbf{x} = \textbf{x}_{\text{trans}} + \textbf{x}_{\text{ai}}\).
However, through experiments we show that this approach introduces a prior dependency on traditional methods. As illustrated in Figure~\ref{fig:RWDRLwoLap}, we conducted model training with and without the inclusion of the traditional Laplacian smoothing method as part of the output. It can be observed that without the traditional Laplacian smoothing model, the convergence of the method is not achieved. On the other hand, although the model including Laplacian smoothing can converge,  the proportion of decisions made by the traditional method ($ \frac{\Vert \textbf{x}_{\text{trans}} \Vert_2}{\Vert \textbf{x}_{\text{trans}}\Vert_2 + \Vert\textbf{x}_{\text{ai}}\Vert_2} $) is significantly higher than that made by the intelligent method, rendering the intelligent model trivial in the smoothing process. \remove{In this paper, we aim to address the challenge of mitigating the model's dependency on traditional methods.  }

\section{Methodology}\label{sec:method}

\begin{figure*}[tbp]
\centering
\includegraphics[width=\linewidth]{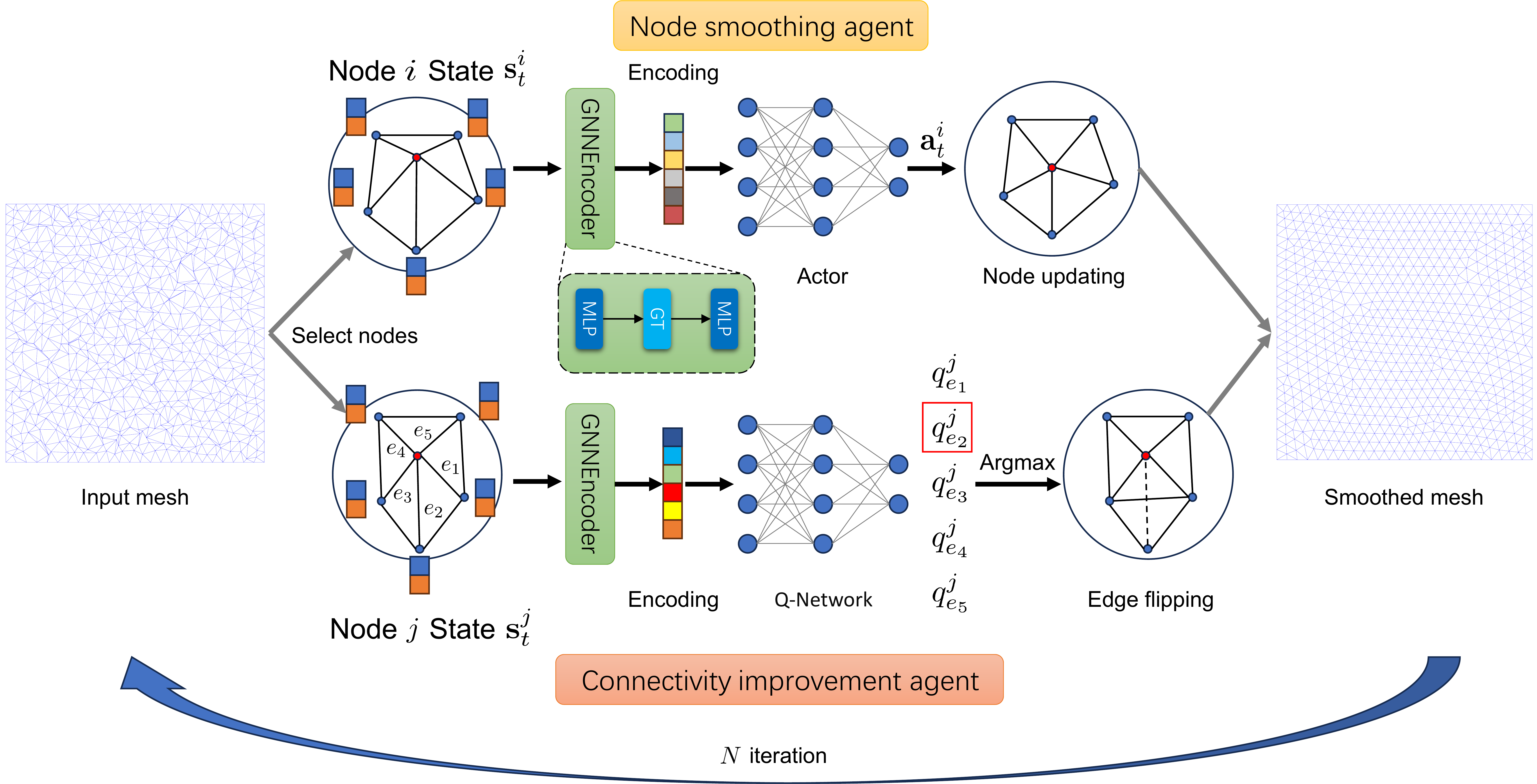}
\caption{\textit{GNNRL-Smoothing model. During the smoothing process, the agent takes mesh nodes and their adjacent nodes as input and performs state encoding through graph convolution. The encodings are then used by either the actor network or the Q-Network to compute corresponding continuous or discrete actions. Smoothing all nodes on the mesh allows us to obtain high-quality meshes.}}
\label{fig:MethodGNNRL}
\end{figure*}
To address the current dependency of intelligent smoothing methods on prior data or knowledge, we further leverage the learning capabilities of graph neural networks on unstructured data and reinforcement learning methods to train intelligent agents, proposing a new mesh smoothing model, GNNRL-Smoothing, as illustrated in Figure~\ref{fig:MethodGNNRL}. During mesh smoothing, we initially take the mesh node and its neighboring nodes as inputs, encoding the node states through graph convolution to generate state encoding.
In the node smoothing agent, we use an Actor-Critic algorithm, TD3, to train our continuous action agent. The actor outputs the displacement of the mesh node to smooth the mesh elements. In the connectivity improvement agent, we employ a value-based algorithm to determine which edges need to be flipped. The Q-Network outputs the Q-value for each edge, and the agent selects the edge with the highest Q-value for flipping to enhance the mesh connectivity. Through the collaboration of these two agents, we can achieve optimization operations for smoothing low-quality meshes. Notably,  GNNRL-Smoothing does not require labeled expert data or prior knowledge of traditional \add{smoothing} methods during training, making it a more efficient and flexible approach.

\subsection{Mesh state encoding}
Unlike images and natural language, mesh data is a typical example of unstructured data. Using traditional neural networks often encounters issues with varying numbers of adjacent nodes and input data sequence order problems. To address these challenges, we employ GNNs to encode the state of the mesh, as illustrated by the GNNEncoder module in Figure~\ref{fig:MethodGNNRL}. As defined in Section \ref{subsec:review}, the state of the mesh node $i$ at time step $t$ consists of two components: the coordinates of the mesh nodes, represented by $\mathbf{X}_{t}^{i}$, and the connectivity information, represented by $\mathbf{A}_{t}^{i}$. For notational convenience, we omit the node index $i$ and time step $t$, and the following calculations are performed with respect to \textit{a specific node and time step}.
\begin{figure}[tbp]
\centering
\includegraphics[width=0.75\columnwidth]{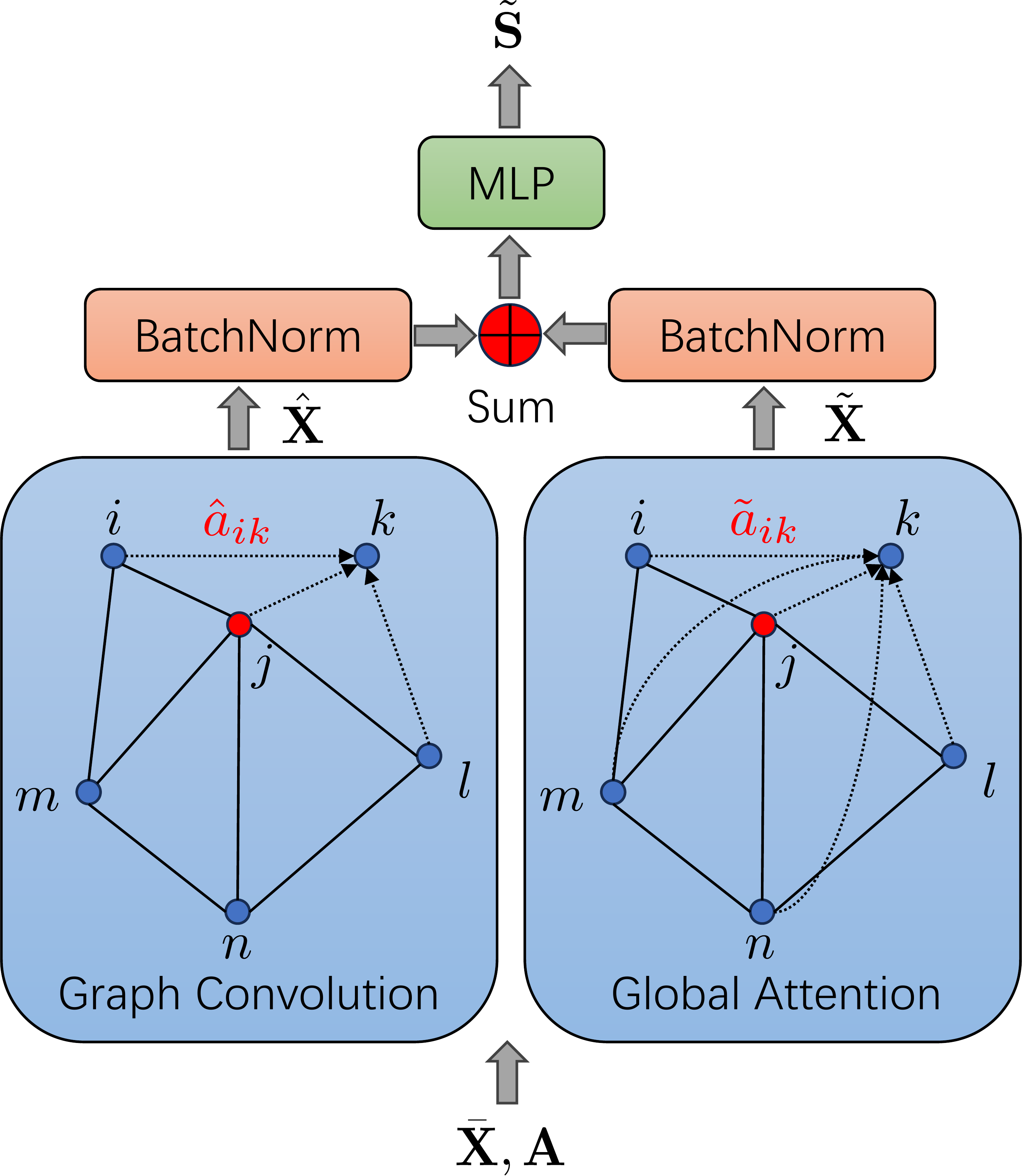}
\caption{\textit{The Graph Transformer block. The block takes the $\bar{\mathbf{X}} $ and $\mathbf{A} $ features of the graph as input, adopting a graph convolution and a global self-attention block to compute the hidden features, and finally obtains the state embedding through an MLP.}}
\label{fig:GT}
\end{figure}
To encode the mesh state, we employ an Encoder-Processor-Decoder architecture, which consists of two MLPs as the encoder (\( \text{MLP}_e \)) and decoder  (\( \text{MLP}_d  \)) for node features, and utilized Graph Transformer (GT)  \cite{shiMaskedLabelPrediction2021a, rampasekRecipeGeneralPowerful2023} to encode the features of nodes and the connectivity of the mesh. The entire process of encoding the mesh feature points can be represented as follows:
\begin{equation}
\begin{aligned}
    \bar{\mathbf{X}}=\text{MLP} \left( \mathbf{X}\right),
    \tilde{\mathbf{S}}=\text{GT}\left(\bar{\mathbf{X}} , \mathbf{A} \right),
    \mathbf{S}= \text{MLP}_d \left(  \tilde{\mathbf{S}}  \right).
\end{aligned}
\end{equation}
The encoded mesh state $\mathbf{S}$ is a tensor of shape $\left(N+1\right) \times d_{f}$, where $N$ is the number of neighboring nodes of mesh node, and $d_{f}$ is the feature dimension.

The GT block consists of a self-attention layer and a global attention layer to learn the graph representation of the state, which is shown is Figure~\ref{fig:GT}. Let $\bar{\mathbf{X}}= \left[ \mathbf{\bar{x}}_0, \mathbf{\bar{x}}_1, \dots, \mathbf{\bar{x}}_{N-1}, \mathbf{\bar{x}}_{N} \right]^T \in \mathbb{R}^{(N+1) \times d_f}$, the node features after graph convolution, $\hat{ \mathbf{X}}=\left[ \hat{\mathbf{x}}_0, \hat{\mathbf{x}}_1, \dots, \hat{\mathbf{x}}_{N-1}, \hat{\mathbf{x}}_{N} \right]^T$, are computed as:
\begin{equation}
\hat{\mathbf{x}}_k = \mathbf{W}_1 \mathbf{\bar{x}}_k+ \sum_{j \in \mathcal{N} \left( k \right) } \hat{a}_{kj} \mathbf{W}_2 \mathbf{\bar{x}}_j,
\end{equation}
where $\mathcal{N}(k)$ is the set of neighboring nodes connected to node $k$, $\mathbf{W}_1$ and $\mathbf{W}_2 $ are the model parameters, and $\hat{a}_{kj}$ is the attention coefficient computed via dot product attention. The node features obtained via global attention are:
\begin{equation}
\tilde{\mathbf{X}}=\textrm{GlobalAtten} \left( \bar{\mathbf{X}} \right) .
\end{equation}
The final node features are computed via a MLP with batchnorm \cite{ioffeBatchNormalizationAccelerating2015} (BN), which can be expressed as:
\begin{equation}
\tilde{\mathbf{S}}=\textrm{GT}(\mathbf{X}, \mathbf{A}) = \textrm{MLP} \left(  \textrm{BN}\left( \hat{ \mathbf{X}}\right) + \textrm{BN}\left( \tilde{\mathbf{X}} \right) \right).
\end{equation}

To further enhance the model's expressive power, we utilize different mesh state encoders in the node smoothing agent and the connectivity improvement agent, respectively.

\subsection{Actions}
\textbf{Node smoothing agent.}
In DRL-Smoothing, the model employs Laplacian smoothing as part of its output to achieve mesh smoothing (in fact, almost 100\%). In contrast, our proposed node smoothing movement model directly outputs the displacement of mesh node \(\Delta \mathbf{x}_t^i\), and update the node coordinate of node \( i \) by  \(\mathbf{x}^i_{t+1} = \mathbf{x}_t^i + \Delta \mathbf{x}_t^i\). Given the encoded mesh features $\mathbf{S}_t^i$, we map it to the displacement vector through an MLP:
\begin{equation}
\Delta \mathbf{x}_t^i = \textrm{MLP}_{actor} \left( {\mathbf{S}_t^i} _{[mapping_i, :]} \right),
\end{equation}
where   $ mapping_i$ is the index of the centering node (red node in Figure~\ref{fig:RLDefine}).

In standard continuous action reinforcement learning algorithms, the agent's output is typically scaled to be within the range of -1 to 1, requiring subsequent rescaling to a specific range. However, an inappropriate range can lead to the generation of invalid training samples (for example, when the range is large, only a small range of action is valid), which reduces the sample efficiency of reinforcement learning. To mitigate this issue, we adopt a small action range of -0.25 to 0.25 for both model training and inference, which means that the absolute value of the increments in the x and y directions will not exceed 0.25. We will validate the effectiveness of our chosen action range in Section \ref{sec:ExpReward}.

Another network related to action computation is the critic network, which evaluates the quality of actions under the current state. We simply implemented it using an MLP: it takes concatenation of  $ {\mathbf{S}_t^i }_{[mapping_i, :]}$ and $ \Delta \mathbf{x}_t^i$, and outputs the corresponding Q-value.

\textbf{Connectivity improvement agent.}
In the connectivity improvement agent, the Q-Network takes the current state of the mesh as input, and outputs the expected Q-values for each action (edge to flip), as illustrated in Figure~\ref{fig:MethodGNNRL}. Since the number of adjacent edges for each node is not fixed, we use action masking \cite{huangCloserLookInvalid2022} in the output of Q-values to remove invalid actions. Specifically, we set the output layer dimension of the Q-Network to 15, which is sufficient for almost all mesh nodes. Let \( N_ j\) denote the number of adjacent nodes for node \( j \), and \( \mathbf{S}_t^j \) represent the state encoding of the mesh node. Then, the output of the connectivity improvement agent is given by:
\begin{equation}
\begin{aligned}
    \mathbf{Q}_t^j &=\textrm{Q-Network} \left( \mathbf{S}_t^j \right) ,\\
    \tilde{\mathbf{Q}}_t^j &= \textrm{Softmax}\left( {\mathbf{Q}_t^j}\right)  \odot \mathbf{I}_{N_j}, \\
    e_t^j &= \argmax \tilde{\mathbf{Q}}_t^j,
\end{aligned}
\end{equation}
where \( \mathbf{I}_{N_j} = [ \overbrace {1,1,\ldots ,1}^{N_j+1},\underbrace{0,0,\ldots ,0}_{14-N_j}]^T \), Softmax function converts the $\mathbf{Q}_t^j$ into a probability distribution, $ \odot$ is the element-wise multiplication,  and \( e_t^j \) indicates the edge connected to which node will be flipped. If \( e_t^j \) equals to $mapping_j$, then no flipping operation is executed.

In the proposed method, we adopt an improved version of the architecture proposed in D3QN as the Q-Network, which can be represented as:
\begin{equation}
\begin{aligned}
    \textrm{Q-Network} (\mathbf{S}_t^j )  = & \textrm{V-Network}\left(  \mathbf{S}_t^j \right) \mathbf{1}_{N_j} + \textrm{A-Network} \left( \mathbf{S}_t^j \right)  \\ & - \max_i \left( \textrm{A-Network} \left( \mathbf{S}_t^j \right)_{[i]}  \right) \mathbf{1}_{N_j} , \\
    \textrm{V-Network}\left(  \mathbf{S}_t^j \right) = & \text{MLP}_{value} \left( \frac{1}{N_j}\sum_{i=1}^{N_j} \mathbf{S}^j_{t[i,:]}\right) ,\\
    \textrm{A-Network} \left( \mathbf{S}_t^j \right) = & \text{MLP}_{adv} \left( \mathbf{S}_t^j \right),
\end{aligned}
\end{equation}
where A-Network (Advantage Network) models the relative importance of each action,  V-Network (State-Value Network) outputs the estimated state value, and $ \mathbf{1}_{N_j} $ is the vector of all ones with length $N_j$. The detailed model architecture parameters are provided in Appendix \ref{app:network}.

\subsection{Reward design}
\textbf{Node smoothing agent.}
Reward design is crucial in reinforcement learning, as it directly influences agent behavior. Well-designed rewards guide the agent toward desired goals, whereas poorly designed rewards lead to suboptimal behavior. In the proposed model, we have made the following adjustments to the reward function in DRL-Smoothing: firstly, the introduction of an advancement-based reward function term; secondly, modification of the penalty term for invalid elements.

At time step $t $, let \( q_{t, j}^i \) denote the aspect ratio of the adjacent mesh element \( j \) of node $ i $, and \( \tilde{q}_{t, j}^i = \frac{1}{q_{t, j}^i} \) represent the mesh quality metric. When \( \tilde{q}_{t, j}^i \) is 1, the shape of the element is optimal, and when it is 0, the element is degenerated. This transformation \( \tilde{q}_{t, j}^i = \frac{1}{q_{t, j}^i} \) is necessary because the original aspect ratio ranges from 1 to positive infinity, which would lead to non-convergent training. Despite the difference, this mesh quality metric is equivalent to that defined in DRL-Smoothing. Then ,a potential-based function can be defined as:
\begin{equation}
\phi(\mathbf{s}^i_t)= \min_{j \in \{1,\ldots,N_i \}}(\tilde{q}^i_{t, j}),
\end{equation}
which calculates the minimum quality of the mesh elements adjacent to the mesh node. In DRL-Smoothing, the reward is defined as $r_t^i=\phi(\mathbf{s}^i_{t+1})$. In our proposed method, we modify it to $r_t^i=\gamma \phi(\mathbf{s}^i_{t+1}) - \phi(\mathbf{s}^i_{t})$.
The agent receives a positive reward even if it takes no action in DRL-Smoothing; in contrast, our proposed reward grants positive rewards only for actions that enhance mesh quality. Furthermore, we can fine-tune the reward threshold by adjusting the value of $\gamma$. With a small $\gamma$,  the agent must achieve substantial quality improvements to earn a reward.

DRL-Smoothing penalizes invalid outputs with a constant value of -1. In contrast, we propose a more refined approach to penalizing invalid elements. By assigning a negative quality metric to invalid elements, we can convey the severity of bad actions to the agent. We define an indicator function $\mathbb{I}(q_j^i)$,  which equals 1 for valid elements and -1 for invalid elements. Our revised potential-based function is defined as follows:
\begin{equation}
\phi(\mathbf{s}^i_t)= \min_{j \in \{1,\ldots,N_i \}}\left( \mathbb{I}\left( \tilde{q}_{t, j}^i\right)  \tilde{q}^i_{t, j}\right). \label{eq:rewardfunc}
\end{equation}
The definition of the reward function remains unchanged.

\textbf{Topology improvement agent.}
For the connectivity improvement agent, there are three possible outcomes after executing an action: mesh quality improvement, no change (no flipping operation), and mesh quality degradation. Since we are using discrete actions, we can simply define the good and bad actions without considering the degree of goodness or badness. Therefore, unlike the continuous reward function used in node movement models, we assign constant rewards of 1, -0.1, and -1 to the above three cases, respectively.

\subsection{\add{Surface mesh smoothing}}\label{sec:SurfaceMesh}
\begin{figure}[tbp]
\centering
\includegraphics[width=\columnwidth]{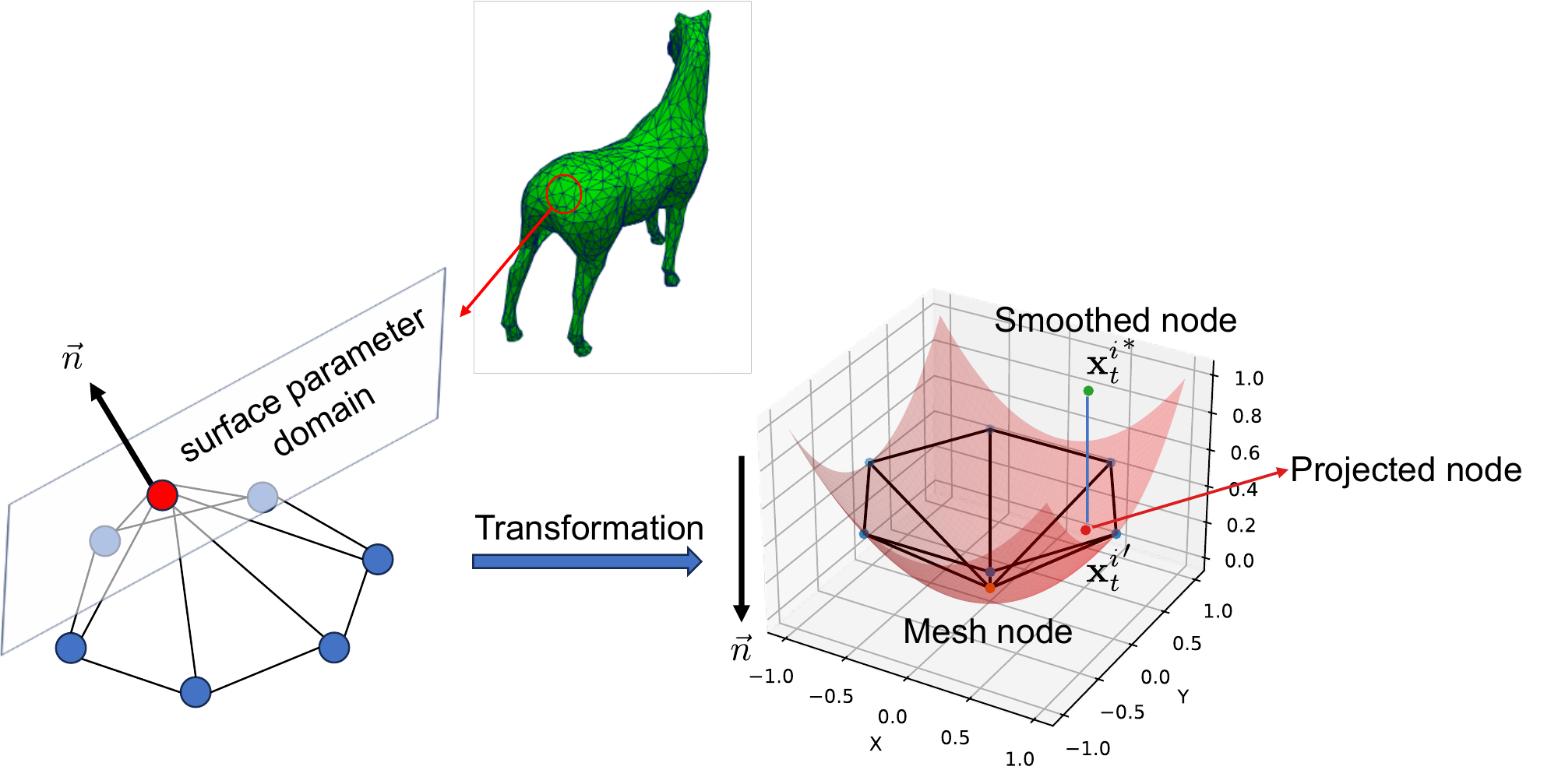}
\caption{\textit{Local surface fitting. The mesh node and its adjacent node are fitted using a quadratic surface, which is represented as \( z = a x^2 + b y^2 \). Here, \( x \) and \( y \) are the parameters of the surface, and the \( z \)-axis is aligned with the normal vector at the mesh node, determined by the area-weighted average of the normal vectors of the adjacent mesh elements.}}
\label{fig:Fitting}
\end{figure}

Extending the 2D node smoothing agent to 3D is not straightforward. The main challenge lies in ensuring that the mesh nodes remain on the original geometric surface, which means preserving original geometric features. Previous intelligent smoothing methods do not provide an explicit guarantee for this. For example, NN-Smoothing directly uses supervised learning to predict 3D mesh node coordinates, while DRL-Smoothing completely disregards the mesh nodes' adherence to the surface. To make our node smoothing agent achieve feature-preserving surface mesh smoothing, we propose a extra reward term based on local surface fitting. As shown in Figure~\ref{fig:Fitting}, we fit a local quadratic surface to the mesh node and its first-order neighbors after normalization. Let \( {\mathbf{x}_t^i}^* \) denote the smoothed mesh noded produced by the 3D node smoothing agent and \( {\mathbf{x}_t^i}^\prime \) the projected node on the fitted surface, then the extra surface fitting reward is defined as the negative Euclidean distance between \({\mathbf{x}_t^i}^*\) and \( {\mathbf{x}_t^i}^\prime \):
\begin{equation}
r_f \left( \mathbf{s}^i_t \right) = -  \|  {\mathbf{x}_t^i}^* -{\mathbf{x}_t^i}^\prime \|_2   .
\end{equation}
This reward term is added as an extra component to Equation~\ref{eq:rewardfunc}, forming the reward function for 3D surface mesh smoothing. It is worth noting that the projection method used here directly projects to the surface parameter domain rather than solving for the shortest distance from \( \mathbf{x}_t^{i*} \) to the surface. Despite its simplicity, experiments show that this method is sufficient to ensure the adherence of the mesh nodes to the surface. As for the connectivity enhancement method, since it only changes the connectivity of the mesh nodes without altering their coordinates, extending it to 3D only requires changing the input features to three dimensions.

\subsection{Model training}\label{sec:modeltraining}
\begin{table*}[tbp]
\footnotesize
\centering
\begin{tabular}{cccccc}
    \toprule
    \textbf{Model} & \textbf{Supervised data} &\textbf{Episode length} & \makecell[c]{\textbf{Requires traditional } \\ \textbf{method output}}    & \textbf{Surface smoothing}   & \textbf{ Connectivity Optimization} \\
    \midrule
    NN-Smoothing          & \Checkmark  & 1 & \XSolidBrush & \Checkmark & \XSolidBrush  \\
    GMSNet   & \XSolidBrush &1 & \XSolidBrush   & \XSolidBrush & \XSolidBrush  \\
    DRL-Smoothing & \XSolidBrush & 1 & \Checkmark & \Checkmark & \XSolidBrush  \\
    \textbf{GNNRL-Smoothing} & \XSolidBrush & 2 & \XSolidBrush & \Checkmark & \Checkmark  \\
    \bottomrule& 
\end{tabular}
\caption{\textit{Comparisons of the intelligent smoothing methods. Compared to other intelligent smoothing methods, the approach proposed in this paper neither requires high-quality supervised data nor depends on traditional smoothing algorithms. Moreover, it achieves feature-preserving surface mesh smoothing.}}
\label{tab:params}
\end{table*}
In our proposed approach, we utilize two intelligent agents: a node smoothing agent and a connectivity improvement agent, responsible for node smoothing and edge flipping, respectively. We train these agents independently using two classic reinforcement learning algorithms: TD3 for the node smoothing agent and D3QN for the connectivity improvement agent. Due to space limitations, we do not provide detailed explanations of these algorithms. While the previous sections focused on the implementation of the models, including the actor and Q-Network, this section delves into the specific implementation details related to the environment, covering state transitions and environment resets.

The process of mesh smoothing typically involves iterating nodes on the mesh, raising the question of how to define state transitions. In the DRL-Smoothing method, given the current state and action, the next state is randomly selected by moving to the next node and its adjacent nodes on the mesh. However, this method of state transition is problematic for reinforcement learning because the transition of states changes when the indices of mesh nodes change. This means that there is  \textit{no state transition probability distribution} $p( \mathbf{s}_{t+1}| \mathbf{s}_{t}, \mathbf{a}_t)$,  and it does not constitute a Markov Decision Process, which is the foundation of reinforcement learning. As shown in Figure~\ref{fig:RWDRLwoLap}, in this scenario, the model heavily relies on traditional methods for smoothing.
In the GNNRL-Smoothing, we apply the action directly to the node to be smoothed and output the modified mesh and its neighboring node positions as the state. In this case, there is a potential pattern of state transitions. Similarly, the state transition in the connectivity improvement agent follows the same approach.

Another issue to consider is when to terminate an episode, as it can potentially be infinite in length. One possibility is to terminate the episode when an action results in invalid elements, while another involves setting a maximum threshold for \( T \). Theoretically, the optimal smoothing strategy is independent of episode length $T$. Analyzing from the perspective of policy gradient method, it can be derived from Equation~\ref{eq:PG}. Let $T \in \mathbb{N}$,   the objective function of policy gradient method is:
\begin{align}
J\left(\pi\right)  &=\sum_\tau R\left(\tau\right) p\left(\tau| \theta\right)  \\
&= \sum_\tau \sum_{t=0}^{T-1} \gamma^t r_t p\left(\tau| \theta\right)  \\
&= \sum_\tau \sum_{t=0}^{T-1} \gamma^t \left(\gamma \phi\left(\mathbf{s}_{t+1}\right)  - \phi\left(\mathbf{s}_{t}\right) \right)  p\left(\tau| \theta\right)  \\
&= \sum_\tau \left(\gamma^T \tilde{\phi}\left(\mathbf{s}_{T}\right) -\phi\left(\mathbf{s}_0\right) \right)  p\left(\tau| \theta\right)  \\
&\le \sum_\tau \left(\gamma^T \tilde{\phi}\left(\mathbf{s}^*| \mathbf{s}_0\right) -\phi\left(\mathbf{s}_0\right) \right)  p\left(\tau| \theta\right),  \label{eq:lesseq}
\end{align}
where \( \tilde{\phi}(\mathbf{s}^*| \mathbf{s}_0) \) represents the best mesh quality when the node at the optimal state $\mathbf{s}^*$ given the initial state \( \mathbf{s}_0 \), and the equality in Equation \ref{eq:lesseq} is achieved when the final state is optimal.
It can be observed that for a given \( T \), the policy corresponding to the upper bound of the objective function is when the final state resides at the optimal mesh position. The only difference for different \( T \) values lies in the number of steps taken to reach the optimal node position. However, if \( T \) is too large, the model could potentially reach the optimal position at any step, significantly widening the exploration space. Therefore, a smaller \( T \) is more appropriate. In our model, we adopt a maximum episode length of 2.
In Section \ref{subsec:review}, we pointed out that GMSNet uses a policy gradient method with \( T = 1 \).
The situation is similar for  \( T  \), although the two use different reward functions, with the upper bound of its optimization objectives being:
\begin{align}
J(\pi) &=\sum_\tau R(\tau)p(\tau| \theta) \\
&= \sum_\tau \sum_{t=0}^{T-1} \gamma^t r_t p(\tau| \theta) \\
&= \sum_\tau \sum_{t=0}^{T-1} \gamma^t  \phi(\mathbf{s}_{t+1})  p(\tau| \theta) \\
&\le \sum_\tau T \tilde{\phi}(\mathbf{s}^*| \mathbf{s}_0)p(\tau| \theta) \label{eq:GMSNetlesseq},
\end{align}
where Equation~\ref{eq:GMSNetlesseq}  indicates that the accumulated reward obtained should be proportional to $T$. The experimental results and discussions are presented in Section \ref{sec:ExpEpLength}.

Finally, when an episode terminates due to negative volume elements or exceeding the maximum length, we resample new nodes as the initial state. \add{We have now thoroughly detailed all the design elements of our proposed reinforcement learning-based mesh smoothing method. The reinforcement learning training processes based on TD3 and D3QN are well-documented in the literature from extensive studies, so we will not elaborate further here due to space constraints.}
\begin{figure}[tbp]
\centering
\subfloat[]{\includegraphics[width=0.48\columnwidth]{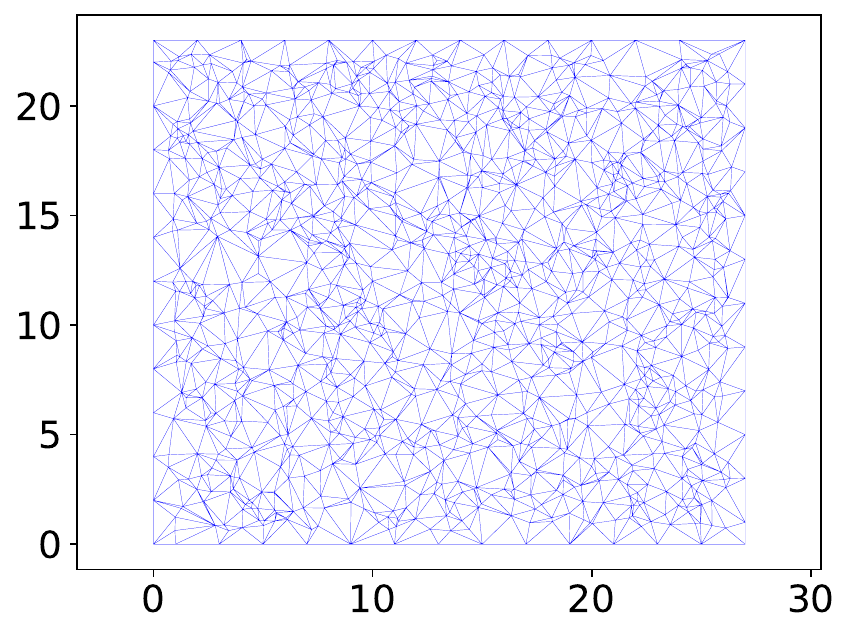}}\hspace{5pt}
\subfloat[]{\includegraphics[width=0.48\columnwidth]{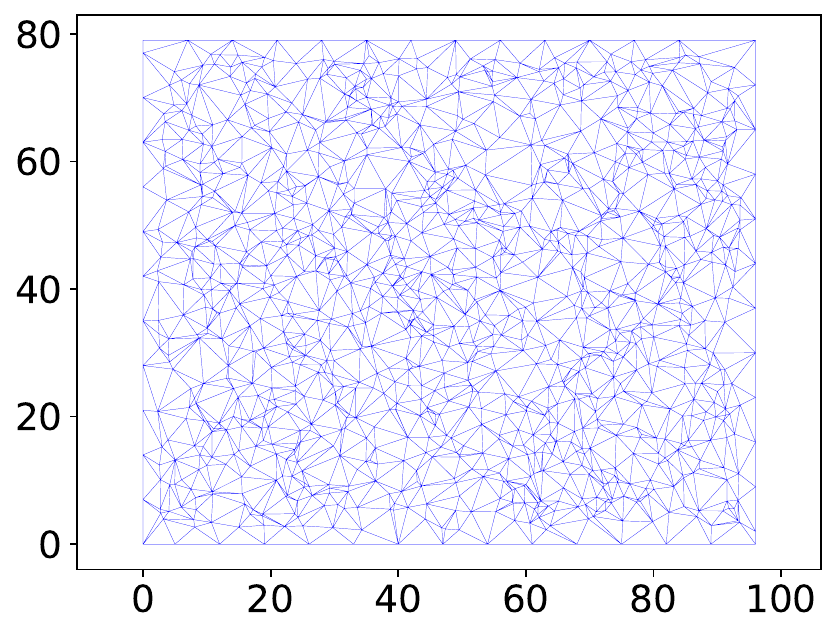}}\\
\caption{\textit{The input mesh samples to train the smoothing agents. We randomly generate mesh nodes within a square region and create a mesh using Delaunay triangulation. The mesh dataset consists of 20 meshes, with 56,089 mesh nodes available for sampling. When the environment is reset, we randomly select a mesh node and its adjacent mesh nodes as the initial state.}}
\label{fig:ExpData}
\end{figure}

From the perspective of reinforcement learning, we compared the configurations of all intelligent smoothing methods, as shown in Table \ref{tab:params}. It can be seen that our method has the least dependency on prior data or knowledge compared to other methods.

\section{Experiments}\label{sec:exp}
In this section,  we first compare the performance of GNNRL-Smoothing with traditional smoothing methods and intelligent smoothing methods in planar mesh smoothing. Next, we conducted ablation experiments to demonstrate the effectiveness of action design, reward function design, and episode length selection. \add{Finally, we evaluate the performance of our proposed model on 3D surface mesh smoothing and validate the effectiveness of the reward function based on local surface fitting.}

\subsection{Baselines and experimental setup}

% 搜索的参数
\begin{table}[tbp]
\centering
\begin{threeparttable}
    \begin{tabular}{ccc}
        \toprule
        \textbf{Parameter}  & \textbf{TD3}        & \textbf{D3QN}              \\
        \midrule
        Learning Rate       & 1e-3                & LogU\tnote{1} (1e-3, 1e-6) \\
        Batch Size          & 64                  & 64                         \\
        Action Noise        & U\tnote{2} (0, 0.5) & -                          \\
        Target Policy Noise & U(0, 0.5)           & -                          \\
        Replay Buffer Size  & LogU(1e3, 1e6)      & LogU(1e3, 1e6)             \\
        $\epsilon$-greedy   & -                   & U(0,1)                     \\
        Discount Rate       & 0.99                & 0.995                      \\
        \bottomrule
    \end{tabular}
    \begin{tablenotes}
        \footnotesize
        \item[1] Loguniform distribution
        \item[2] Uniform distribution
    \end{tablenotes}
\end{threeparttable}
\caption{\textit{Parameter search spaces for TD3 and D3QN.}}
\label{tab:TD3ParamSpace}
\end{table}
% 模型平滑前后绘图
\begin{figure*}[htbp]
\centering
\subfloat{\includegraphics[width=0.25\columnwidth]{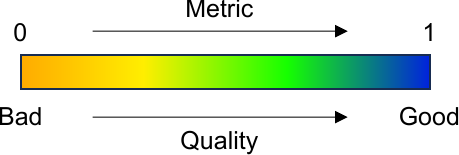}}\\ \setcounter{subfigure}{0}
\subfloat[Case 1 before and after smoothing]{\includegraphics[height=0.5\columnwidth]{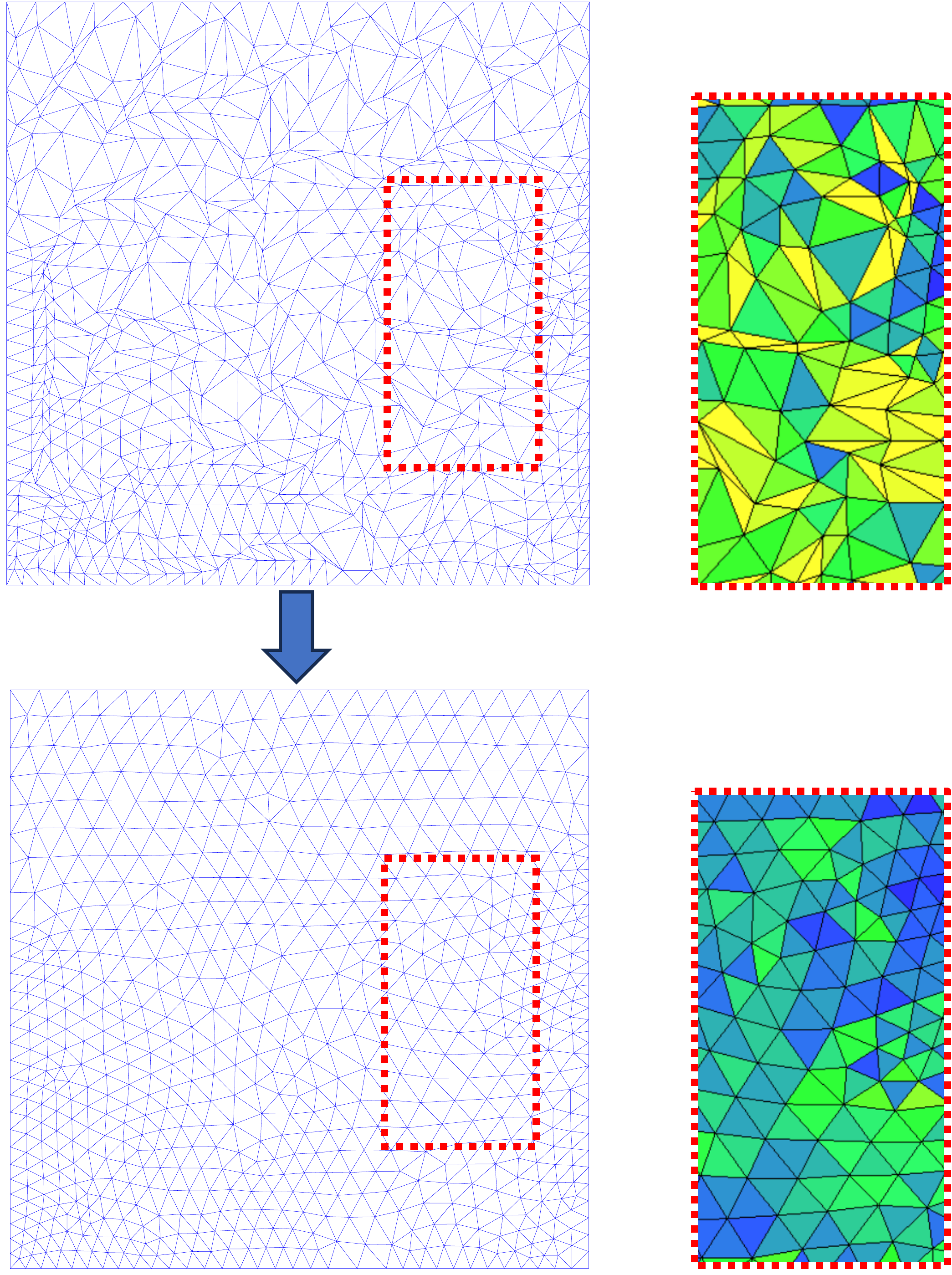}}\qquad
\subfloat[Case 2 before and after smoothing]{\includegraphics[height=0.5\columnwidth]{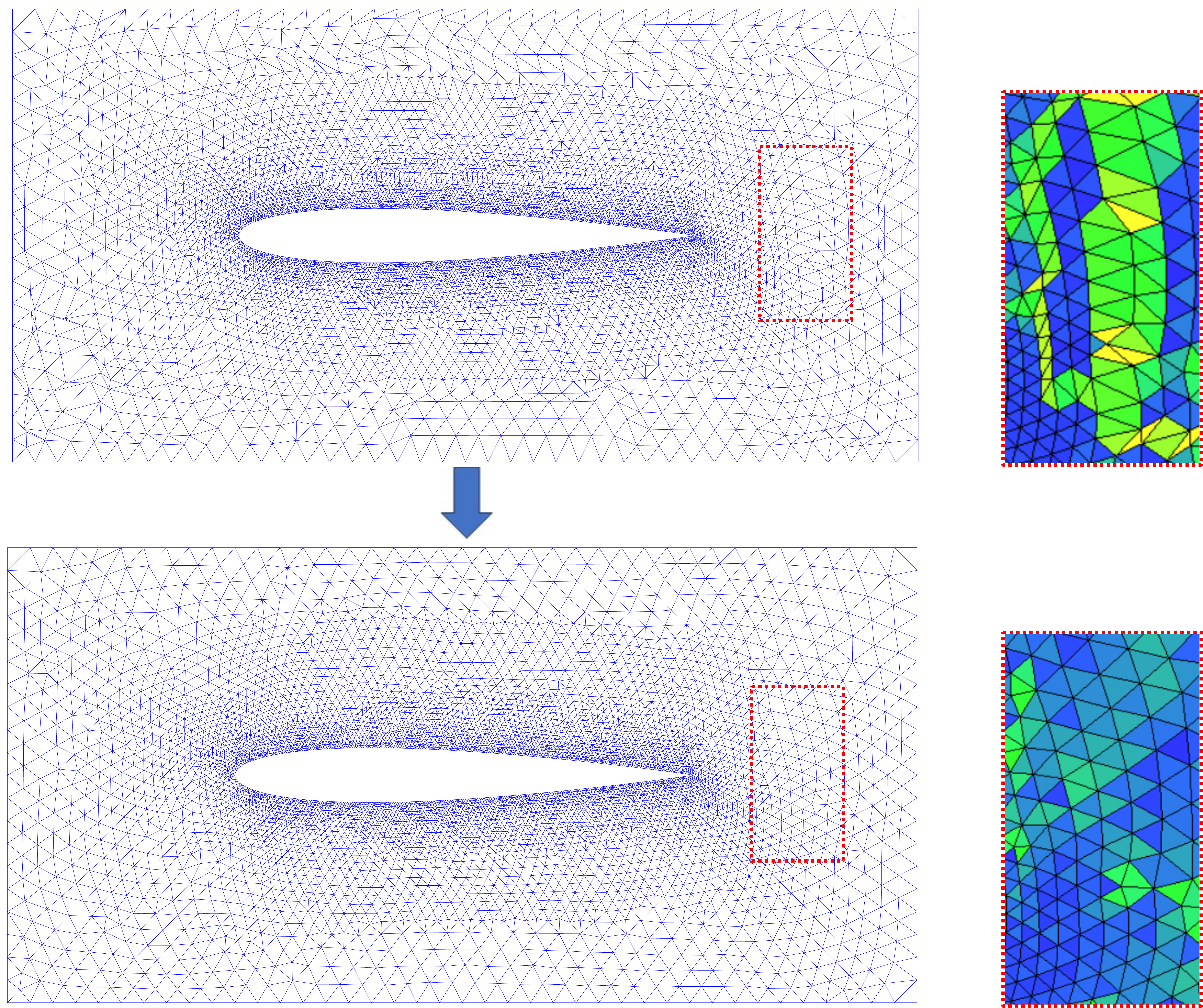}}\qquad \\
\subfloat[Case 3 before and after smoothing]{\includegraphics[height=0.5\columnwidth]{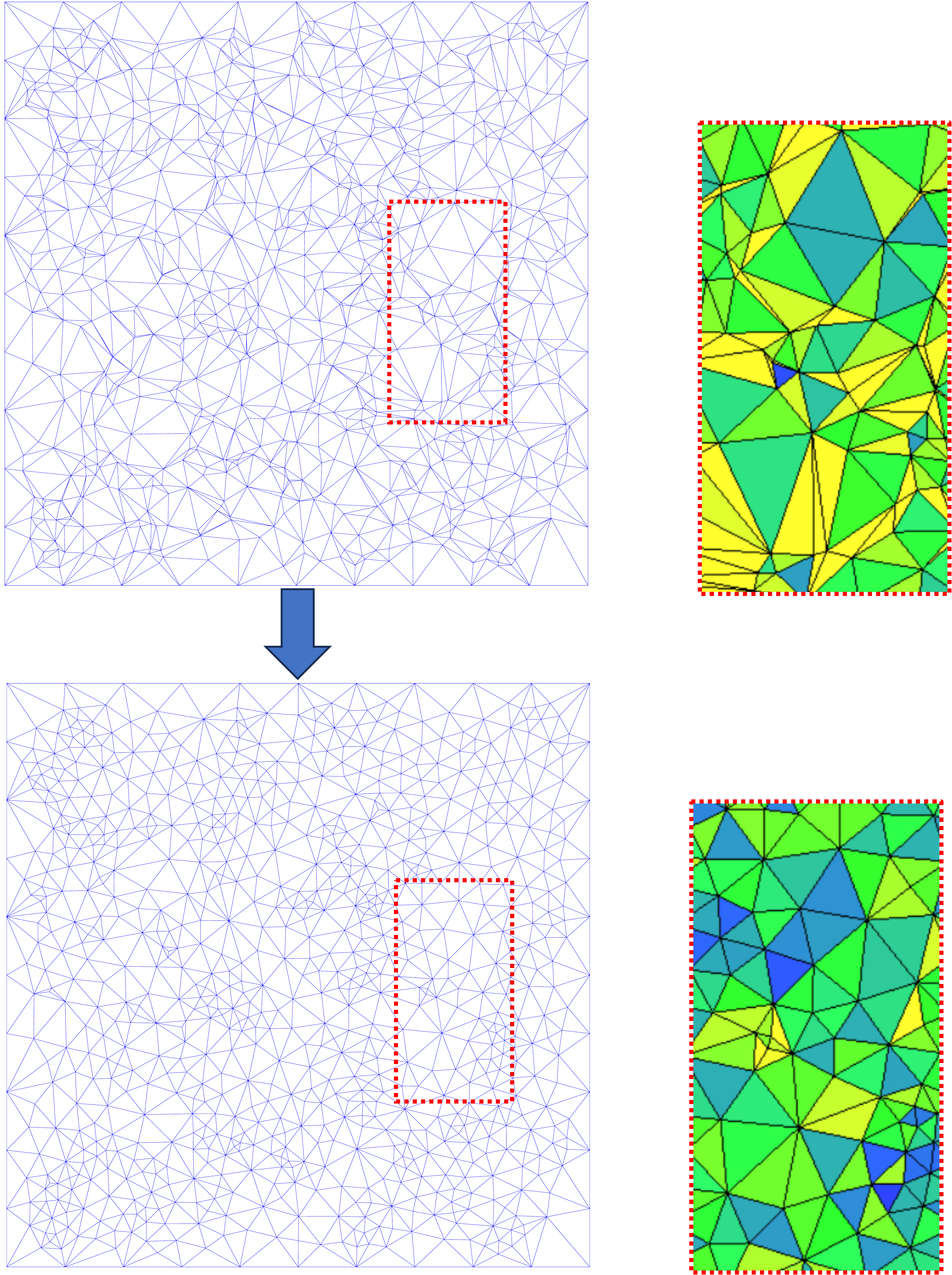}}\qquad
\subfloat[Case 4 before and after smoothing]{\includegraphics[height=0.5\columnwidth]{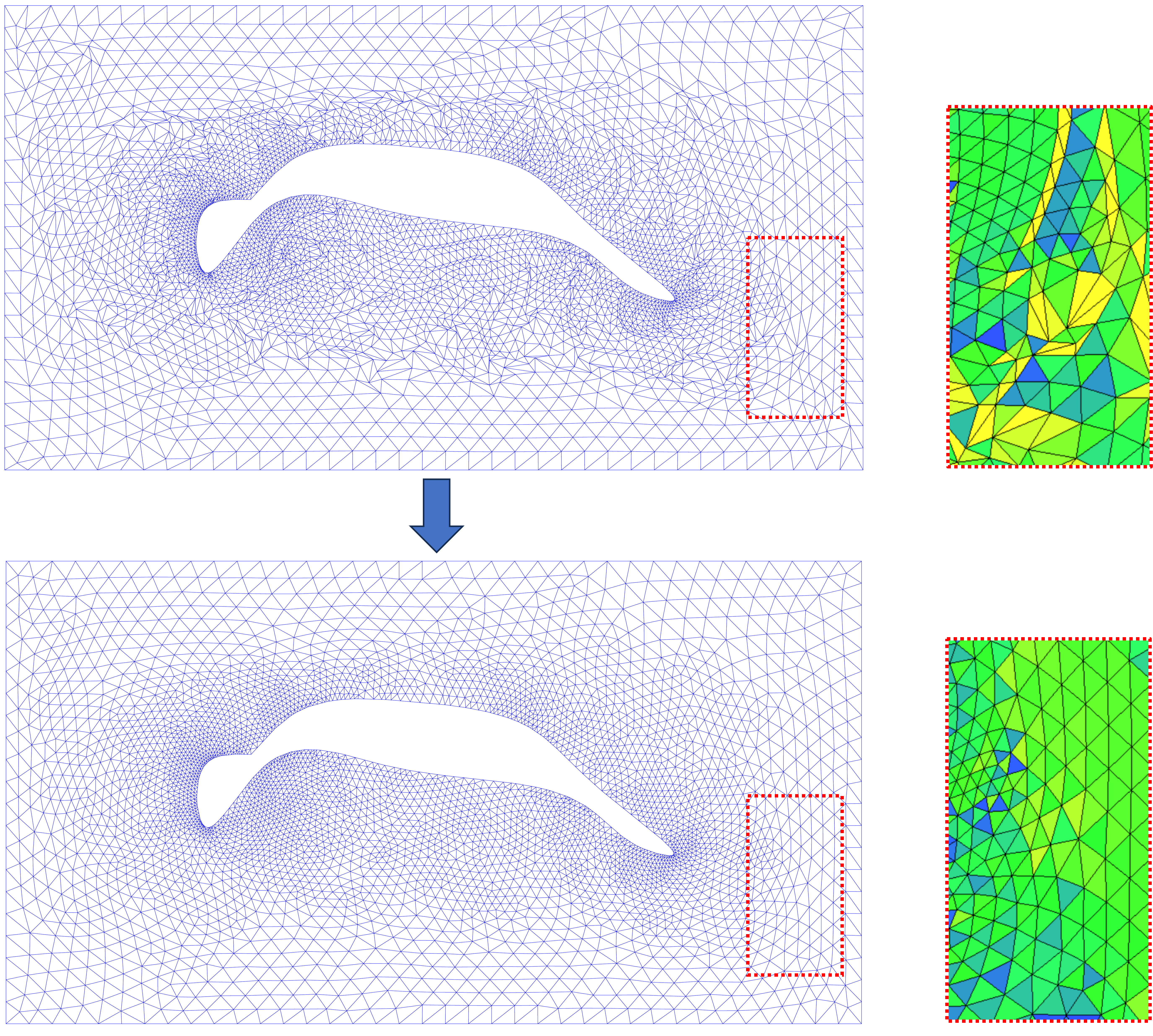}}\qquad \\
\caption{\textit{Results of mesh smoothing using GNNRL-Smoothing on the test cases. In the detailed view, a color scheme is used to indicate the quality of the mesh, with blue representing high quality elements and yellow representing low quality elements, as shown in the legend.}}
\label{exp:CaseBeforeAfter}
\end{figure*}
% 性能列表
\begin{table*}[tbp]
\centering
\begin{threeparttable}
    \begin{tabular}{lcccccccc}
        \toprule
        & \multicolumn{2}{c}{\textbf{Case 1}} & \multicolumn{2}{c}{\textbf{Case 2} } & \multicolumn{2}{c}{\textbf{Case 3}} & \multicolumn{2}{c}{\textbf{Case 4}} \\
        \cmidrule(lr){2-3} \cmidrule(lr){4-5} \cmidrule(lr){6-7} \cmidrule(lr){8-9}
        & $\tilde{q}_{\min} \uparrow$ & $\tilde{q}_{\textrm{mean}} \uparrow$ & $\tilde{q}_{\min}$ & $\tilde{q}_{\textrm{mean}}$ & $\tilde{q}_{\min}$ & $\tilde{q}_{\textrm{mean}}$ & $\tilde{q}_{\min}$ & $\tilde{q}_{\textrm{mean}}$ \\
        \midrule
        Origin                                                       & 0.129                       & 0.790                                & 0.007                 & 0.968                       & 0.006                 & 0.691                       & 0.066                 & 0.869                       \\
        Laplacian Smoothing                                          & 0.529                       & {\color{blue} 0.920}                 & 0.498                 & {\color{blue} 0.937}        & 0.258                 & {\color{red} 0.784} & 0.392 & {\color{red} 0.880} \\
        Angle-based Smoothing                                        & 0.496                       & 0.885                                & 0.408                 & 0.927                       & 0.265                 & 0.754                       & 0.325                 & 0.860                       \\
        GETMe Smoothing                                              & {\color{blue} 0.627}        & 0.817                                & {\color{blue} 0.650}  & 0.919                       & {\color{red} 0.343} & 0.701 & {\color{green} 0.455} & 0.814 \\
        Opt. Smoothing\tnote{1}                                      & 0.558                       & {\color{red} 0.917}                  & {\color{green} 0.539} & {\color{blue} 0.937} & {\color{blue} 0.392} & {\color{green} 0.785} & {\color{blue} 0.483} & {\color{green} 0.881} \\
        \midrule
        GMSNet                                                       & 0.513                       & {\color{green} 0.918}                & 0.507                 & {\color{green} 0.936}       & 0.250                 & {\color{blue} 0.786} & 0.365 & 0.873 \\
        NN-Smoothing                                                 & 0.503                       & 0.916                                & 0.499                 & {\color{red} 0.933}         & 0.164                 & 0.780                       & 0.398                 & {\color{blue}0.882}         \\
        DRL-Smoothing                                                & 0.441                       & 0.859                                & 0.480                 & 0.927                       & 0.250                 & 0.764                       & 0.322                 & 0.855                       \\
        \midrule
        \textbf{GNNRL-Smoothing}                                     & {\color{red} 0.562}         & 0.886                                & 0.506                 & 0.923                       & 0.252                 & 0.735                       & {\color{red} 0.406} & 0.861  \\
        $\text{\textbf{GNNRL-Smoothing}}_{\text{\textbf{withflip}}}$ & {\color{green} 0.578} & 0.884 & {\color{red} 0.520} & 0.922  & {\color{green} 0.391} & 0.772 & {\color{red} 0.406} & 0.865  \\
        \bottomrule
    \end{tabular}
    \begin{tablenotes}
        \item[1] Optimization-based smoothing
    \end{tablenotes}
\end{threeparttable}
\caption{\textit{Comparison of the performance of smoothing methods. Using only the node smoothing agent, GNNRL-Smoothing achieved competitive smoothing performance evaluated using the mesh quality metric \( \tilde{q}_{\min}  \), where higher values ($\uparrow$) indicate better quality. By further integrating the connectivity improvement agent, the quality of meshes with poor topological connections can be significantly improved as illustrated by case 3 in Figure~\protect\ref{fig:ElementDis}. The optimal, second, and third results are marked in {\color{blue} blue}, {\color{green} green}, and {\color{red} red}, respectively.}}
\label{tab:SmoothingPerformance}
\end{table*}
The baseline models included in our study consist of traditional smoothing methods: Laplacian smoothing, Angle-based smoothing, Optimization-based smoothing, and GETMe smoothing. Additionally, three intelligent smoothing models were included: NN-Smoothing model, GMSNet model, and DRL-Smoothing model. The traditional smoothing methods were configured following the specifications outlined in the GMSNet paper. The NN-Smoothing model and GMSNet model were trained and tested using the parameters provided in their original papers, while the DRL-Smoothing model utilized the original code provided by the authors. Similar to GMSNet, GNNRL-Smoothing was trained on randomly generated meshes (as shown in Figure~\ref{fig:ExpData}), and the model's parameters were searched through hyperparameter optimization. For the TD3 algorithm, we employed the Optuna \cite{akibaOptunaNextgenerationHyperparameter2019a} framework to optimize the parameters including action noise (Gaussian noise), target policy noise, and the size of the experience buffer. We tuned the D3QN algorithm with the same framework by optimizing the $\epsilon$-greedy policy's random sampling probability $\epsilon$, the learning rate, and the experience buffer size. The model's hyperparameters and the search space for the optimized hyperparameters are summarized in Table \ref{tab:TD3ParamSpace}. During training, we gradually decayed the noise as training progresses to enable the model to better exploit the existing experience.

\subsection{Planar mesh smoothing}
We tested the smoothing performance of all mesh smoothing methods on four mesh cases and calculated the minimum and average quality metric $\tilde{q}=\dfrac{1}{\textrm{aspect ratio}}$ of the mesh elements before and after smoothing . \add{It is worth noting that although different mesh quality metrics can be used to evaluate mesh quality, they are often equivalent for triangular meshes \cite{knuppAlgebraicMeshQuality2001}. For simplicity, we use \( \tilde{q} \) as the sole metric in this study.} The meshes used for testing were not included in the training dataset, which helps demonstrate the generalization capability of the intelligent smoothing models. For each mesh, we conducted five experiments and reported the average results. Figure~\ref{exp:CaseBeforeAfter} shows the mesh cases before and after smoothing using GNNRL-Smoothing; Table \ref{tab:SmoothingPerformance} presents a comprehensive performance comparison of various smoothing methods applied to different mesh cases. Due to space constraints, the comparison of intelligent smoothing methods is provided in Figure~\ref{fig:AllModels} in Appendix \ref{app:AllModels}.

\subsubsection{Smoothing performance}
% 翻转模型效果
\begin{figure}[tbp]
\centering
\includegraphics[width=0.7\linewidth]{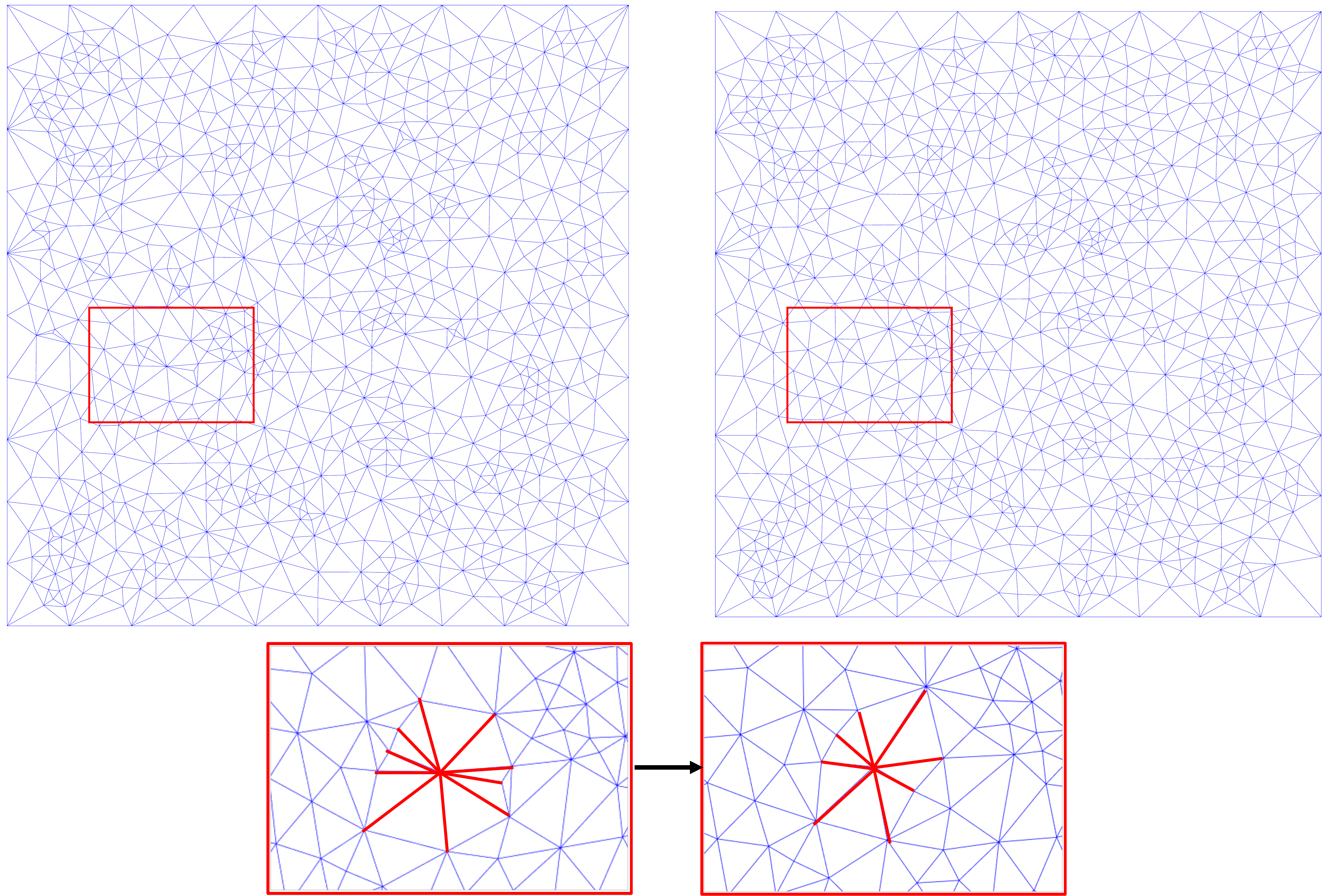}
\caption{\textit{The smoothing result of case 3 for GNNRL-Smoothing with (right) and without (left) the connectivity improvement agent. It can be seen that the connectivity improvement agent reduces unreasonable topological connection, making the node degree distribution more reasonable.}}\label{fig:edgeflip}
\end{figure}
Firstly, as shown in Figure~\ref{exp:CaseBeforeAfter}, the GNNRL-Smoothing model significantly improves the orthogonality and smoothness of the mesh. It achieves superior mesh smoothing quality even for highly distorted meshes such as case 3. Secondly, from Table \ref{tab:SmoothingPerformance}, intelligent smoothing methods have achieved highly competitive results compared to traditional smoothing methods. In some cases, such as NN-Smoothing in case 4 with a $\tilde{q}_{\text{mean}}$ of 0.882, and GMSNet in case 3 with a $\tilde{q}_{\text{mean}}$ of 0.786, they even surpass the optimization-based methods. The performance of DRL-Smoothing is slightly inferior to that of the GMSNet and NN-Smoothing models. A possible explanation for this is that the DRL-Smoothing model was trained on a limited set of mesh samples and cannot generalize well to unseen samples, such as highly distorted mesh elements. Additionally, it relies on the output of traditional methods.

Our proposed model achieves either the second or thrid-best \( \tilde{q}_{\text{min}} \) values across all cases, \add{which is only inferior to GETMe Smoothing and Opt. Smoothing. However, in GETMe Smoothing, the improvement is achieved by repeatedly optimizing the worst mesh elements, while in Opt. Smoothing, it is obtained through computationally demanding procedures.}
In terms of \( \tilde{q}_{\text{mean}} \) , the propoesed model's performance is competitive when compared to both traditional methods and other intelligent smoothing models. The slightly inferior results in \( \tilde{q}_{\text{mean}} \) could be attributed to the use of the min function in our reward function; however, this is justified since the goal of mesh optimization is to improve the quality of the poor elements.

\begin{figure}[tbp]
\centering
\includegraphics[width=\linewidth]{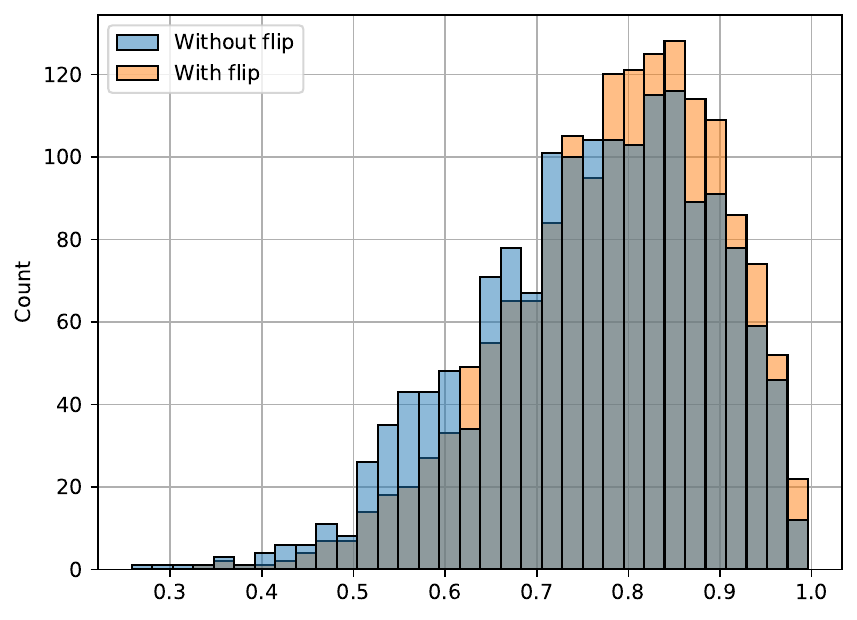}
\caption{\textit{Mesh quality distribution of case 3 with and without connectivity improvement agent. }}\label{fig:ElementDis}
\end{figure}

When further combined with connectivity improvement agent, the GNNRL-Smoothing model significantly improves its ability to handle meshes with poor topologies as shown in Figure~\ref{fig:edgeflip}, which improves the minimum mesh quality from 0.252 to 0.391 by flipping undesirable edges. As shown in Figure~\ref{fig:ElementDis}, the connectivity improvement agent could also reduce the proportion of low-quality mesh elements while increasing the proportion of high-quality mesh elements.
\begin{table}
\centering
\footnotesize
\begin{tabular}{ccc}
    \toprule
    \textbf{Method}                     & \textbf{Time (s)} $\downarrow	$ & \textbf{Speedup (times)} $\uparrow$ \\
    \midrule
    Opt. Smoothing                      & 3.93e-2           & 1.00             \\
    Laplacian Smoothing                 & 8.20e-4           & 48.00            \\
    Angle-based Smoothing               & 2.73e-3           & 14.42            \\
    GETMe Smoothing                     & 3.23e-3           & 12.18            \\
    \midrule
    NN-Smoothing                        & 4.52e-3           & 8.70             \\
    GMSNet                              & 3.74e-3           & 10.53            \\
    DRL-Smoothing                       & 4.29e-3           & 9.18             \\
    \midrule
    \textbf{Node smoothing agent}       & 5.50e-3           & 7.16             \\
    \textbf{Topology improvement agent} & 4.38e-3           & 8.99             \\
    \bottomrule
\end{tabular}
\captionof{table}{\textit{Efficiency comparison of smoothing methods.}}
\label{tab:GPUCPUTime}
\end{table}

\subsubsection{Efficiency}
The runtime efficiency of various smoothing methods was measured through experiments. The experiments were conducted on an Intel(R) Core(TM) i7-9700KF CPU @ 3.60GHz, with an NVIDIA RTX TITAN as the GPU. All code was implemented in Python and ran on the GPU. The experimental results are shown in Table \ref{tab:GPUCPUTime}. From Table \ref{tab:GPUCPUTime}, it can be seen that the Laplacian Smoothing has the highest speedup compared to the optimization-based methods due to its simplicity; all intelligent methods have higher efficiency compared to optimization-based methods. Our proposed model, due to its more complex architecture compared to MLP, shows a slight decrease in efficiency. Meanwhile, the time overhead of the connectivity improvement agent is similar to that of the node smoothing agent, without introducing significant additional time overhead. This allows the model to efficiently optimize both the node positions and the connectivity of the mesh simultaneously.
\begin{figure}[tbp]
	\centering
	\subfloat[GNNRL-Smoothing]{\includegraphics[width=0.4\columnwidth]{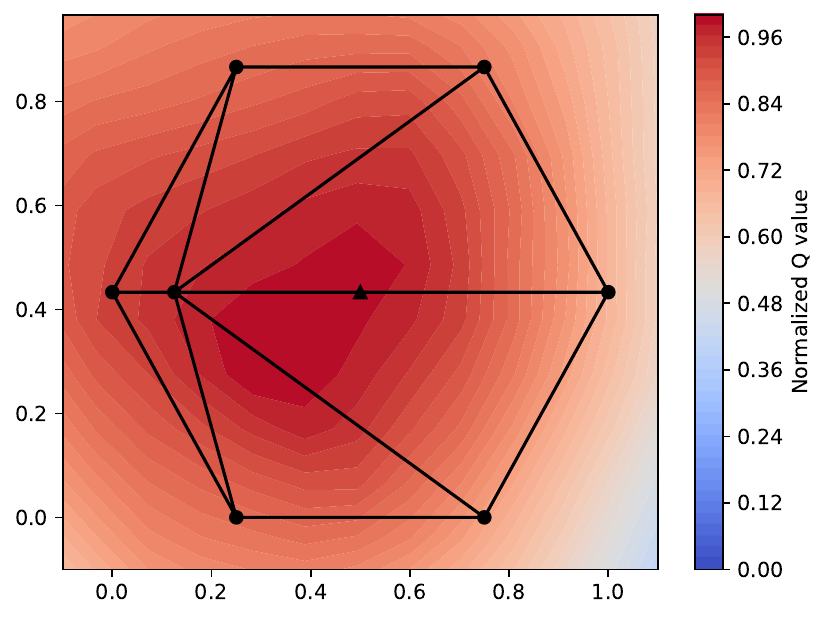}}\hspace{5pt}
	\subfloat[GNNRL-Smoothing]{\includegraphics[width=0.40\columnwidth]{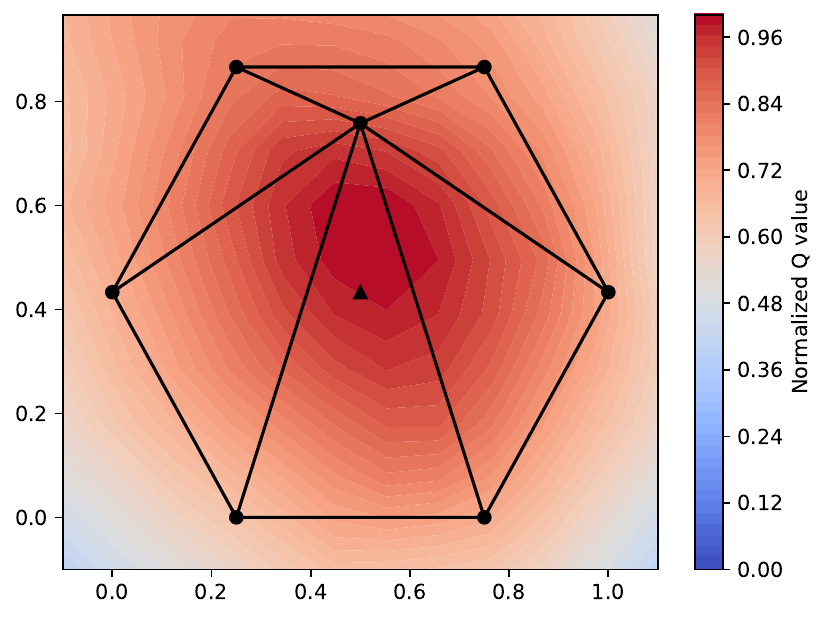}}\\
	\subfloat[DRL-Smoothing ]{\includegraphics[width=0.4\columnwidth]{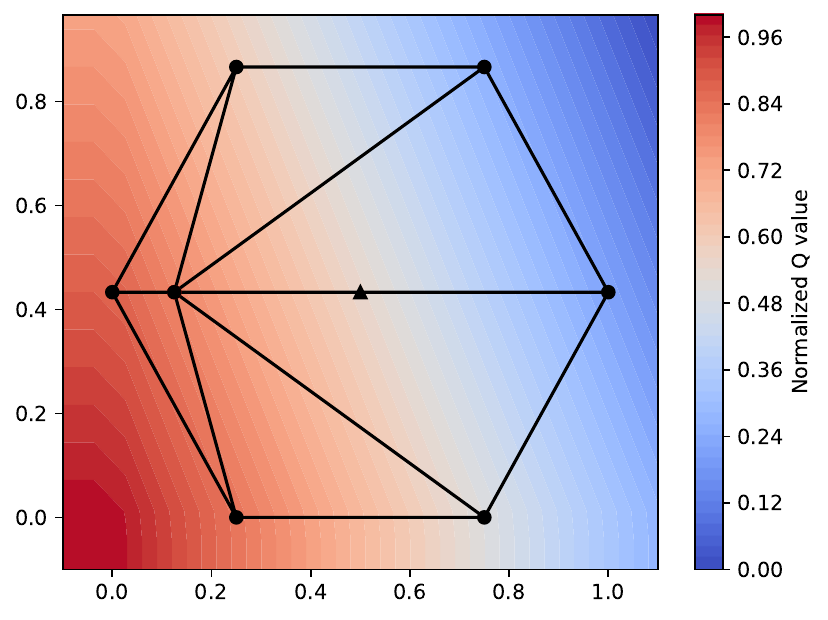}}\hspace{5pt}
	\subfloat[DRL-Smoothing ]{\includegraphics[width=0.4\columnwidth]{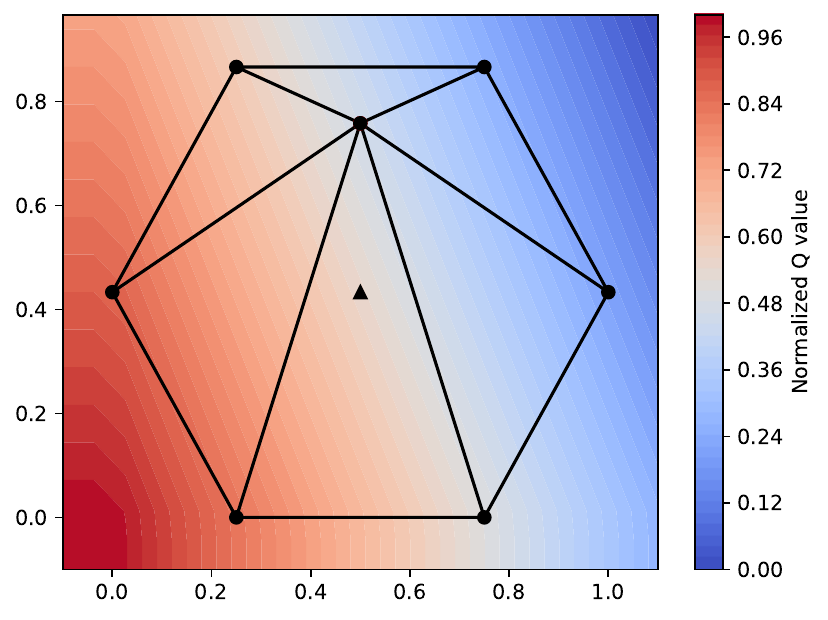}}\\
	\caption{\textit{Visualization of the critic models for GNNRL-Smoothing and DRL-Smoothing.
		The triangular node indicates the optimal node position. The color in the figure shows the changes in normalized Q values, with high Q values displayed in red and low Q values displayed in blue. GNNRL-Smoothing correctly assigns high Q values to positions near the optimal node and low Q values to the invalid regions.}}
	\label{fig:GNNRLvsDRL}
\end{figure}

\subsubsection{\add{GNNRL-Smoothing vs DRL-Smoothing}}

GNNRL-Smoothing and DRL-Smoothing both utilize the Actor-Critic architecture for training, allowing us to understand the learned policy by visualizing the critic model's estimation of state-action values. As shown in Figure~\ref{fig:GNNRLvsDRL}, a simple mesh node and its adjacent nodes, marked with black dots, represent the current state \(\mathbf{s}\), and the optimal node position is indicated by a red triangle. The plot shows the updated node position \((x, y)\) along with the normalized Q value \(q\) (form 0 to 1). It demonstrates that the critic model of GNNRL-Smoothing successfully assigns high Q values when the updated node is near the optimal point and low Q values in invalid regions. On the other hand, DRL-Smoothing incorrectly places high Q values outside the valid region. This further demonstrates that our proposed model successfully models the smoothing process.

\subsection{Ablation study}

\subsubsection{Action design}\label{sec:ExpAction}

\begin{figure}[htbp]
	\begin{minipage}[t]{0.5\linewidth}
		\centering
		\includegraphics[width=\linewidth]{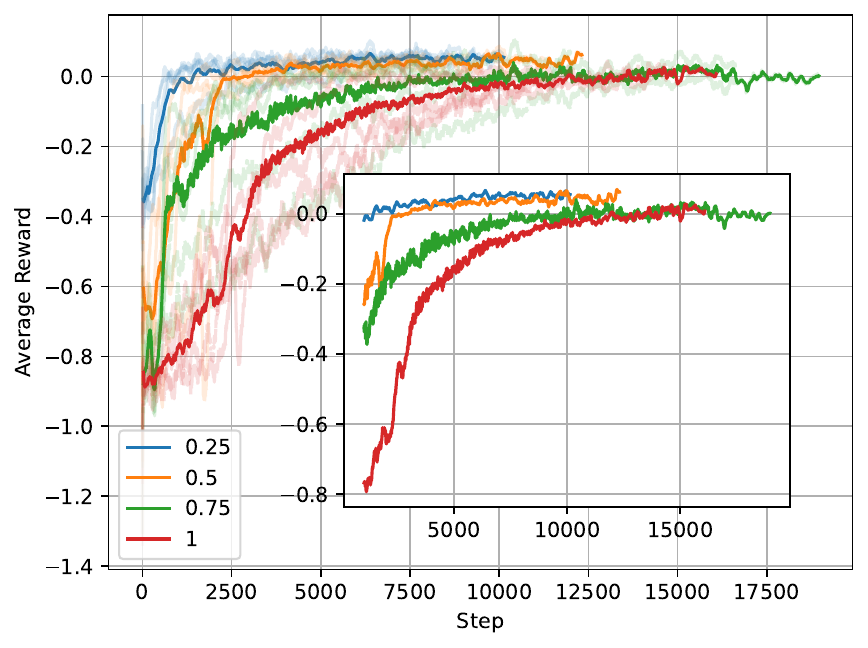}
		\caption{\textit{The Reward-Step curves of the model under different action ranges. The Average Reward of the top-5 trials for each action range is averaged. As shown in the figure, smaller action ranges achieve higher scores and fast convergence.}}
		\label{fig:ActionPlot}
	\end{minipage} \quad
	\begin{minipage}[t]{0.5\linewidth}
		\centering
		\includegraphics[width=\linewidth]{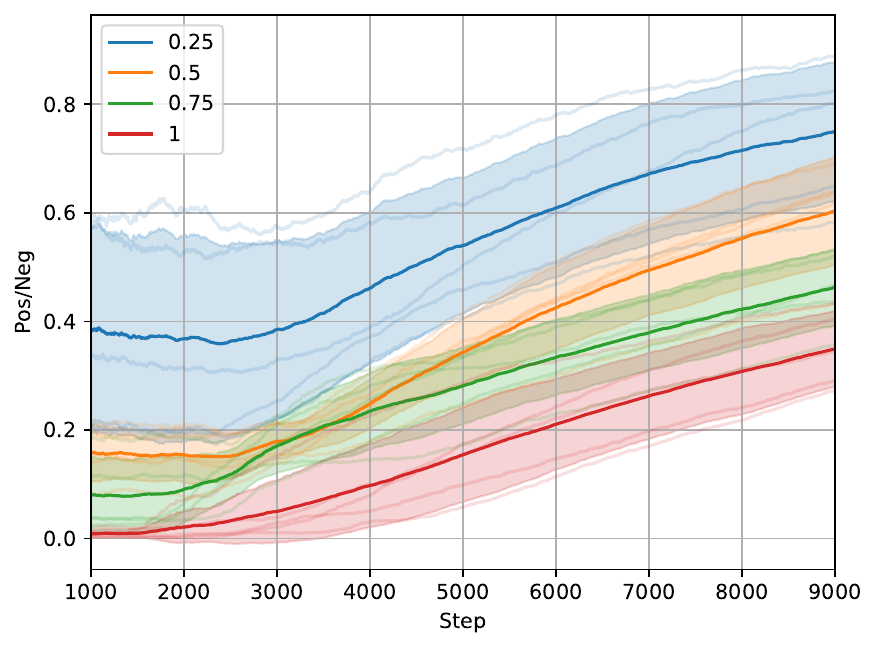}
		\caption{\textit{The relative proportion of samples with positive reward to those with negative reward. Smaller action ranges make the model less likely to generate invalid mesh elements, thus improving sample efficiency during model training.}}
		\label{fig:PosVsNeg}
	\end{minipage}
\end{figure}

In the action design of the node smoothing agent, we constrained the action range to a narrow range (-0.25, 0.25). We highlighted the potential benefits of employing a smaller action range for facilitating model training and convergence. Here, we demonstrate through experiments that a smaller action range improves model performance by increasing \textit{sample efficiency}, which means it can generate more samples with positive rewards to facilitate agent learning.
Firstly, agents with different action ranges (0.25, 0.5, 0.75, and 1) were trained using the same parameter configuration. For efficiency, we only tuned the parameter exploration noise through grid search. Exploration noise was sampled at 10 points within the range of 0 to 0.5, and for each sample point, 5 experiments with different seeds were conducted, totaling 200 experiments. For each action range, we show the top-5 average rewards with a state window size of 1000, as shown in Table \ref{tab:ActionRanges}, and the training curves are depicted in Figure~\ref{fig:ActionPlot}. The results indicate that a smaller action range yields higher rewards and better convergence speed.
Secondly, we monitored the relative proportion of samples with positive rewards to those with negative rewards during training, as shown in Figure~\ref{fig:PosVsNeg}. It can be observed that a smaller action range results in a higher proportion of samples that help the model learn effective strategies. Conversely, as the action range increases, the proportion of positive rewards decreases, leading to slower model convergence.
% 动作范围对模型性能影响
\begin{table}[tbp]
\centering
\begin{tabular}{ccc}
    \toprule
    \textbf{Action range} & \textbf{Average Reward ($\times$ 1e-2)}                \\
    \midrule
    (-0.25, 0.25)         & \textbf{8.8  {\fontsize{6pt}{6pt}\selectfont $\pm$ 1.0}} \\
    (-0.5, 0.5)           & 7.4  {\fontsize{6pt}{6pt}\selectfont $\pm$ 0.7} \\
    (-0.75, 0.75)         & 5.3  {\fontsize{6pt}{6pt}\selectfont $\pm$ 3.1} \\
    (-1,1)                & 3.3  {\fontsize{6pt}{6pt}\selectfont $\pm$ 0.9}          \\
    \bottomrule
\end{tabular}
\caption{\textit{Summary of experiment results with different action ranges. For each action range, the top-5 results were selected and averaged. As the action range gradually increases, the Average Reward obtained by the model gradually decreases.}}
\label{tab:ActionRanges}
\end{table}
\begin{table*}[tbp]
	\footnotesize
	\centering
	\begin{threeparttable}
		\begin{tabular}{ccccc}
			\toprule
			& {\textbf{Original} \textbf{reward}}           & {\textbf{Adaptive} \textbf{penalty}}          & {\textbf{Adv.-based} \textbf{reward}}\tnote{1} & \makecell{\textbf{Adv.-based} \textbf{reward}+ \\ \textbf{Adaptive} \textbf{penalty}} \\ \midrule
			{Case 1} & 0.807 \fontsize{6pt}{6pt}\selectfont{± 0.001} & 0.818 \fontsize{6pt}{6pt}\selectfont{± 0.002} & 0.809 \fontsize{6pt}{6pt}\selectfont{± 0.001} & \textbf{0.853 \fontsize{6pt}{6pt}\selectfont{± 0.001}} \\
			{Case 2} & 0.900 \fontsize{6pt}{6pt}\selectfont{± 0.000} & 0.900 \fontsize{6pt}{6pt}\selectfont{± 0.000} & 0.889 \fontsize{6pt}{6pt}\selectfont{± 0.000} & \textbf{0.920 \fontsize{6pt}{6pt}\selectfont{± 0.000}} \\
			{Case 3} & 0.595 \fontsize{6pt}{6pt}\selectfont{± 0.000} & 0.649 \fontsize{6pt}{6pt}\selectfont{± 0.074} & 0.595 \fontsize{6pt}{6pt}\selectfont{± 0.000} & \textbf{0.660 \fontsize{6pt}{6pt}\selectfont{± 0.089}} \\
			{Case 4} & 0.789 \fontsize{6pt}{6pt}\selectfont{± 0.025} & 0.827 \fontsize{6pt}{6pt}\selectfont{± 0.001} & 0.824 \fontsize{6pt}{6pt}\selectfont{± 0.000}  & \textbf{0.848 \fontsize{6pt}{6pt}\selectfont{± 0.001}} \\
			\bottomrule
		\end{tabular}
		\begin{tablenotes}
			\item[1] Advancement-based reward
		\end{tablenotes}
	\end{threeparttable}
	\caption{\textit{The smoothing performance of agents trained with different reward functions. For each mesh, we conducted ten experiments and provided the average and standard deviation of each result. The best result is highlighted in bold.}}
	\label{tab:RewardFunc}
\end{table*}

\subsubsection{Reward design}\label{sec:ExpReward}
For the node smoothing agent, we made two modifications to the reward function: introducing an advancement-based reward term and an adaptive penalty for invalid elements. We conducted experiments to determine whether these changes lead to performance improvements in the agent. We trained the models using different reward functions in the node smoothing agent and employed the same hyperparameter optimization method as in Section~\ref{sec:ExpAction}. The experimental results are shown in Table \ref{tab:RewardFunc}. From these results, it is evident that both proposed enhancements have improved the performance of agents in mesh smoothing. Overall, the adaptive penalty method has shown a more significant improvement in smoothing performance, while the reward reshape method has brought a relatively smaller enhancement. Combining both methods leads to the best smoothing performance, validating the effectiveness of our design.

\subsubsection{Episode length}\label{sec:ExpEpLength}
\begin{figure}[tbp]
\footnotesize
\begin{minipage}[b]{0.5\linewidth}
    \centering
    \begin{tabular}{cc}
        \toprule
        \makecell{\textbf{Episode}  \textbf{Length}} & \makecell{\textbf{Average}  \textbf{Reward}}         \\ \midrule
        2                                          & 0.769 \fontsize{6pt}{6pt}\selectfont{$\pm$} 0.018  \\
        4                                          & 1.493 \fontsize{6pt}{6pt}\selectfont{$\pm$} 0.047  \\
        8                                          & 3.009 \fontsize{6pt}{6pt}\selectfont{$\pm$} 0.040  \\
        16                                         & 5.878 \fontsize{6pt}{6pt}\selectfont{$\pm$} 0.186  \\
        32                                         & 12.114 \fontsize{6pt}{6pt}\selectfont{$\pm$} 0.231 \\ \bottomrule
    \end{tabular}
    \captionof{table}{\textit{The mean top-5 average rewards under different maximum episode lengths. As Equation~\protect\ref{eq:GMSNetlesseq} illustrates, the upper bound of rewards achievable by the model is linearly related to the episode length.}}
    \label{tab:eplength}
\end{minipage}\quad
\begin{minipage}[b]{0.5\linewidth}
    \centering
    \includegraphics[width=\linewidth]{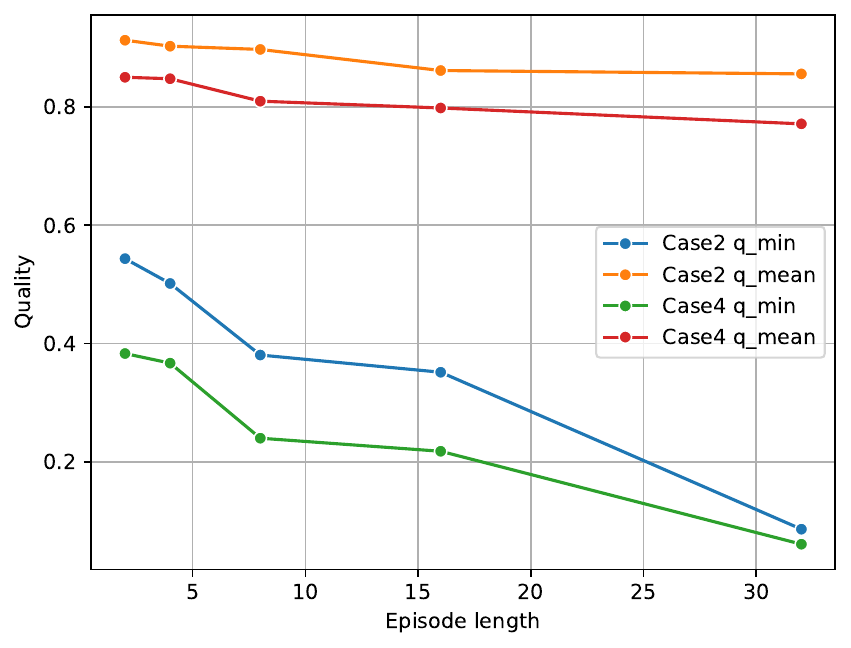}
    \captionof{figure}{\textit{The performance of agents trained with different episode lengths. As the episode length increases, the model's performance gradually deteriorates.}}
    \label{fig:eplength}
\end{minipage}
\end{figure}

To investigate the impact of different episode lengths on model performance, we conducted experiments with episode lengths of 2, 4, 8,16 and 32, respectively. The experimental results on case 2 and case 4 are shown in Figure~\ref{fig:eplength}. It can be observed that the models were trainable with different episode lengths. However, as the episode length increased, there was a slight decrease in the smoothing performance of the models. Although we note that the optimal strategy corresponding to different episode lengths remains the same, the training difficulty varies with different lengths, with shorter episode lengths being easier to train. The agents trained with GMSNet reward have also been shown in Table \ref{tab:eplength}. The average reward is indeed linearly related to the episode length, consistent with Equation~\ref{eq:GMSNetlesseq}. The experimental results demonstrate the correctness of the episode length we adopted, while also providing insight into selecting effective Episode lengths under different forms of reward functions.

\subsection{\add{Surface mesh smoothing}}
\subsubsection{Smoothing performance}
\begin{figure*}[htbp]
	\centering
%	\subfloat{\includegraphics[width=0.25\columnwidth]{Figs/Legend.pdf}}\\ \setcounter{subfigure}{0}
	\subfloat[Case 1 before and after smoothing]{\includegraphics[height=0.5\columnwidth]{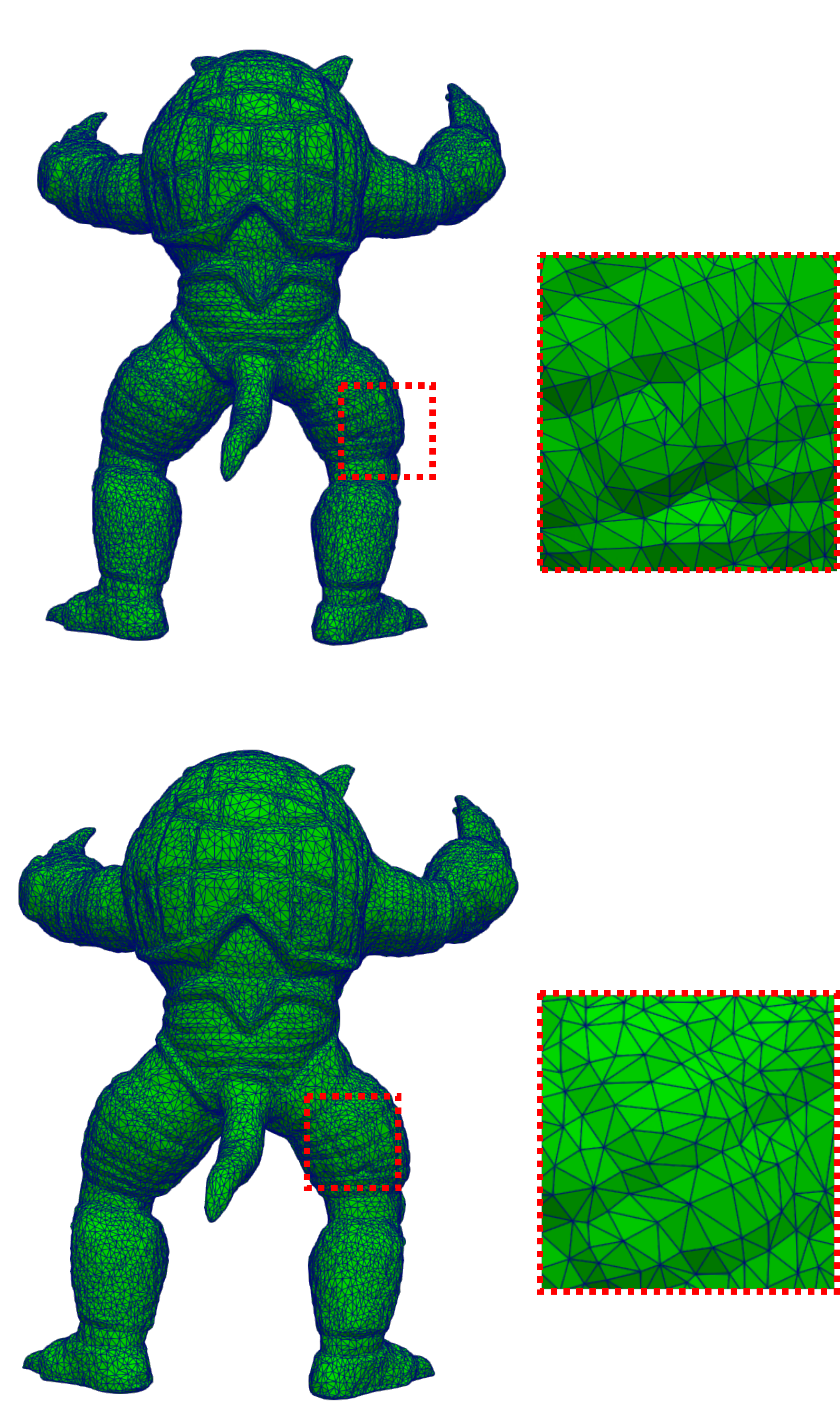}}\qquad
	\subfloat[Case 2 before and after smoothing]{\includegraphics[height=0.5\columnwidth]{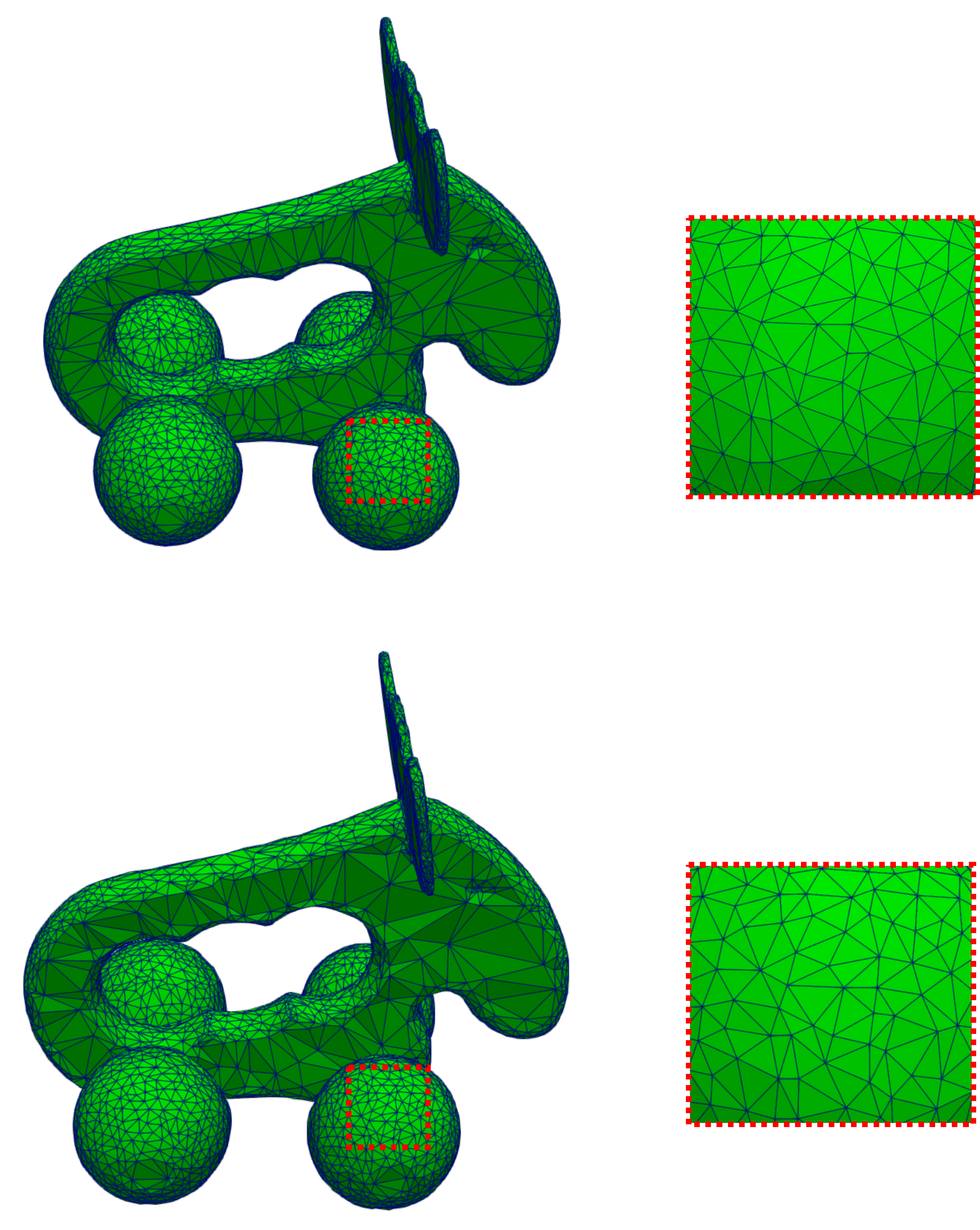}}\qquad 
	\subfloat[Case 3 before and after smoothing]{\includegraphics[height=0.5\columnwidth]{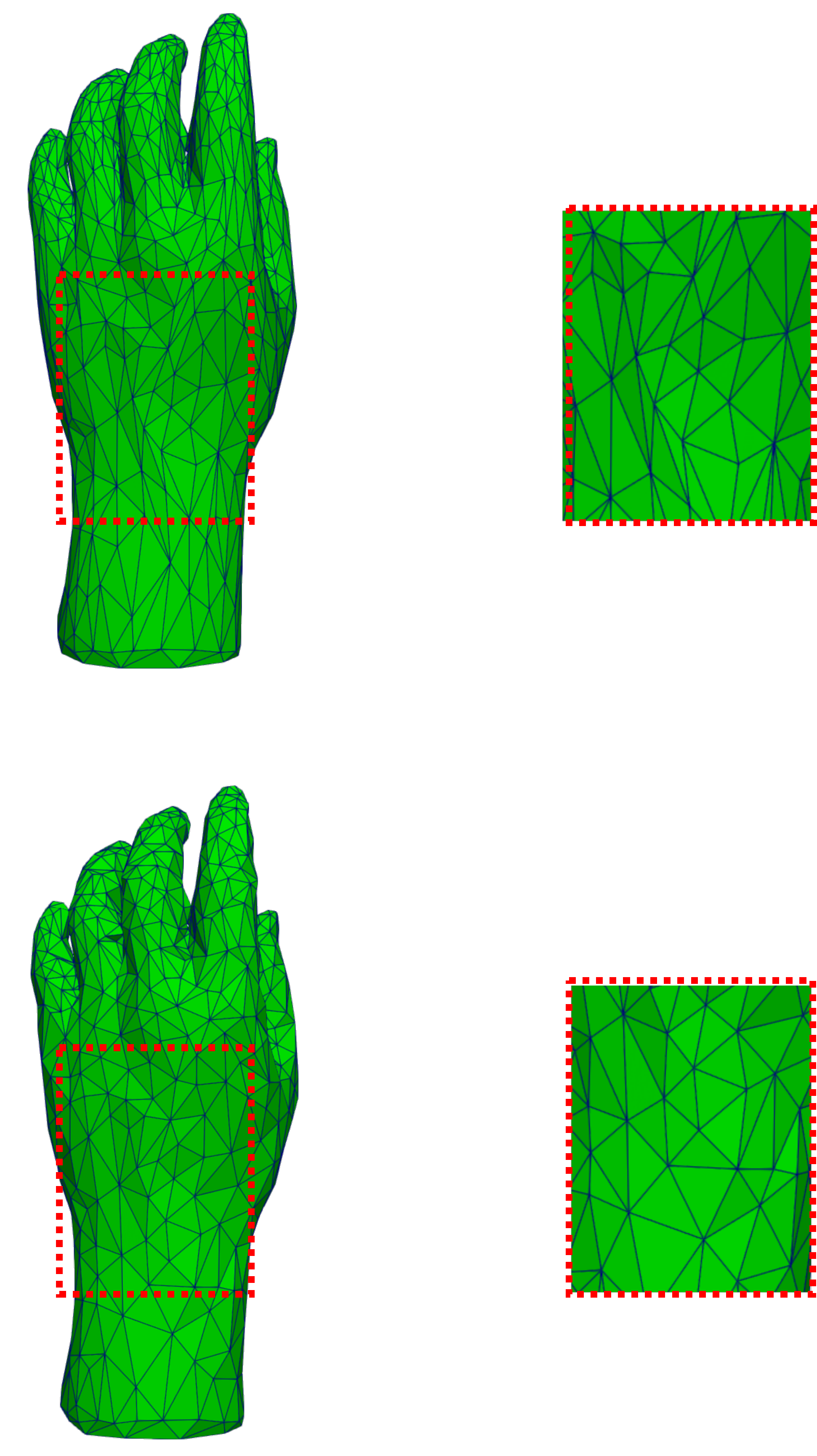}}\qquad
	\subfloat[Case 4 before and after smoothing]{\includegraphics[height=0.5\columnwidth]{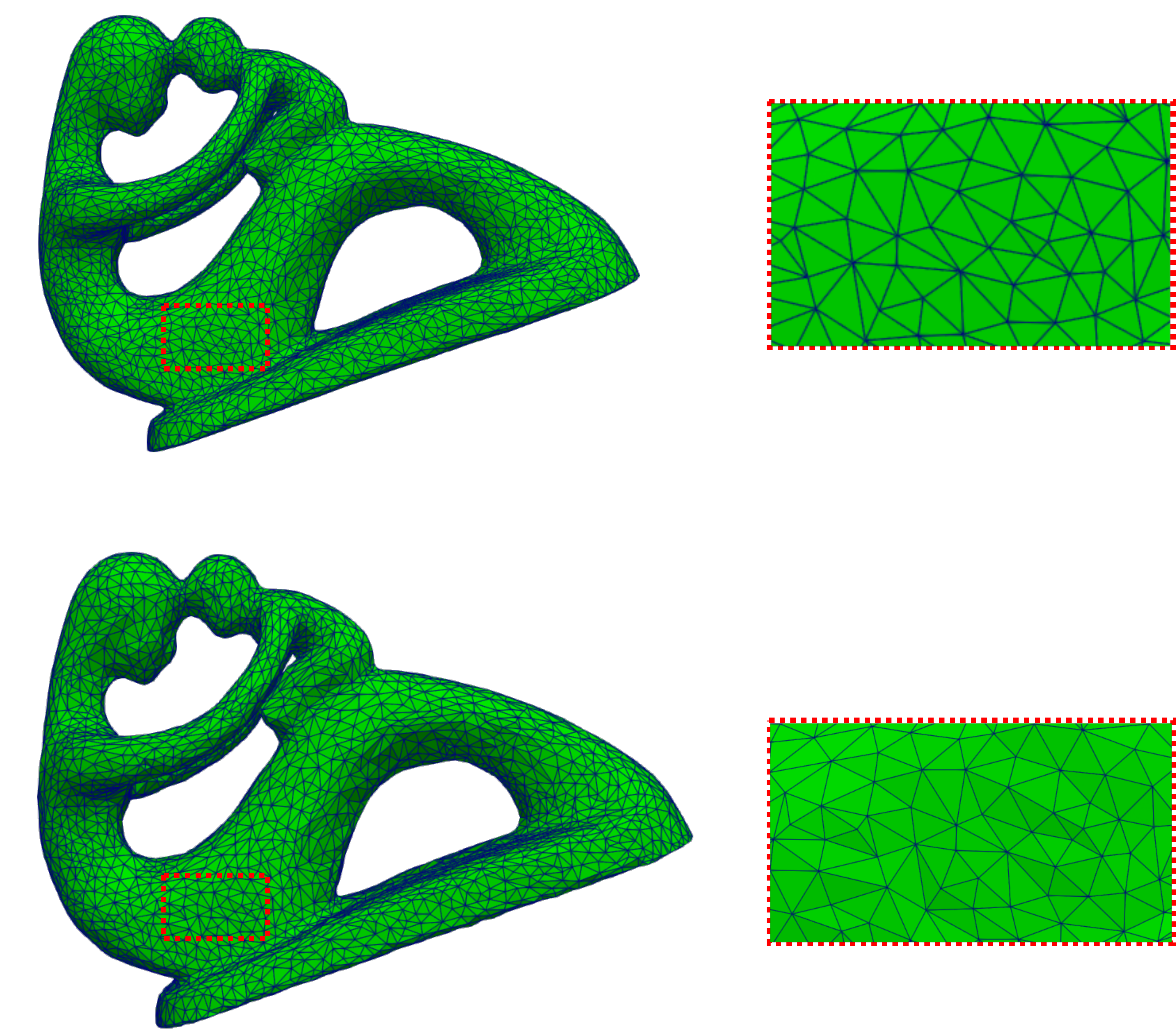}}\qquad 
	\caption{\textit{Results of surface mesh smoothing using GNNRL-Smoothing on the test cases.}}
	\label{exp:SurfaceCase}
\end{figure*}

To evaluate the performance of our proposed method for surface mesh smoothing, we selected four test cases from the AIM@SHAPE library. Our method was applied to each mesh for 10 smoothing iterations, without any additional operations to ensure mesh adherence to the surface after smoothing. The meshes before and after smoothing are shown in Figure~\ref{exp:SurfaceCase}, and the corresponding mesh quality metrics are presented in Table~\ref{tab:meshquality}. As shown in Figure~\ref{exp:SurfaceCase}, despite multiple smoothing operations performed by the proposed model on complicated surface meshes, the smoothed meshes still preserve the original geometric features and shape. Furthermore, the model is capable of smoothing and optimizing surface meshes across various local surface geometries without requiring additional operations, such as node projection. Table~\ref{tab:meshquality} demonstrates the improvement in mesh quality, as indicated by the $\tilde{q}_{\textrm{mean}}$ and mean min angle metrics, before and after smoothing. This highlights the applicability and potential of the proposed method for surface mesh smoothing.
\begin{table}[tbp]
	\centering
	\begin{tabular}{ccccc}
		\toprule
		 & \multicolumn{2}{c}{\textbf{$\tilde{q}_{\textrm{mean}}$}} & \multicolumn{2}{c}{\textbf{Mean min angle}} \\
		\cmidrule(lr){2-3} \cmidrule(lr){4-5}
		\textbf{Case} &\textbf{Before} & \textbf{After} & \textbf{Before} & \textbf{After} \\
		\midrule
		Case1 & 0.33 & 0.70 & 35.66  & 36.94 \\
		Case2 & 0.34 & 0.70 & 35.41 &  37.16 \\
		Case3 & 0.35 & 0.68 & 30.07 &  35.84 \\
		Case4 & 0.35 & 0.72 & 36.95 &  38.59 \\
		\bottomrule
	\end{tabular}
	\caption{\textit{Surface mesh quality  before and after smoothing.}}
	\label{tab:meshquality}
\end{table}

\subsubsection{Necessity of the surface fitting reward $r_f$}
\begin{figure}[tbp]
	\centering
	\includegraphics[width=\linewidth]{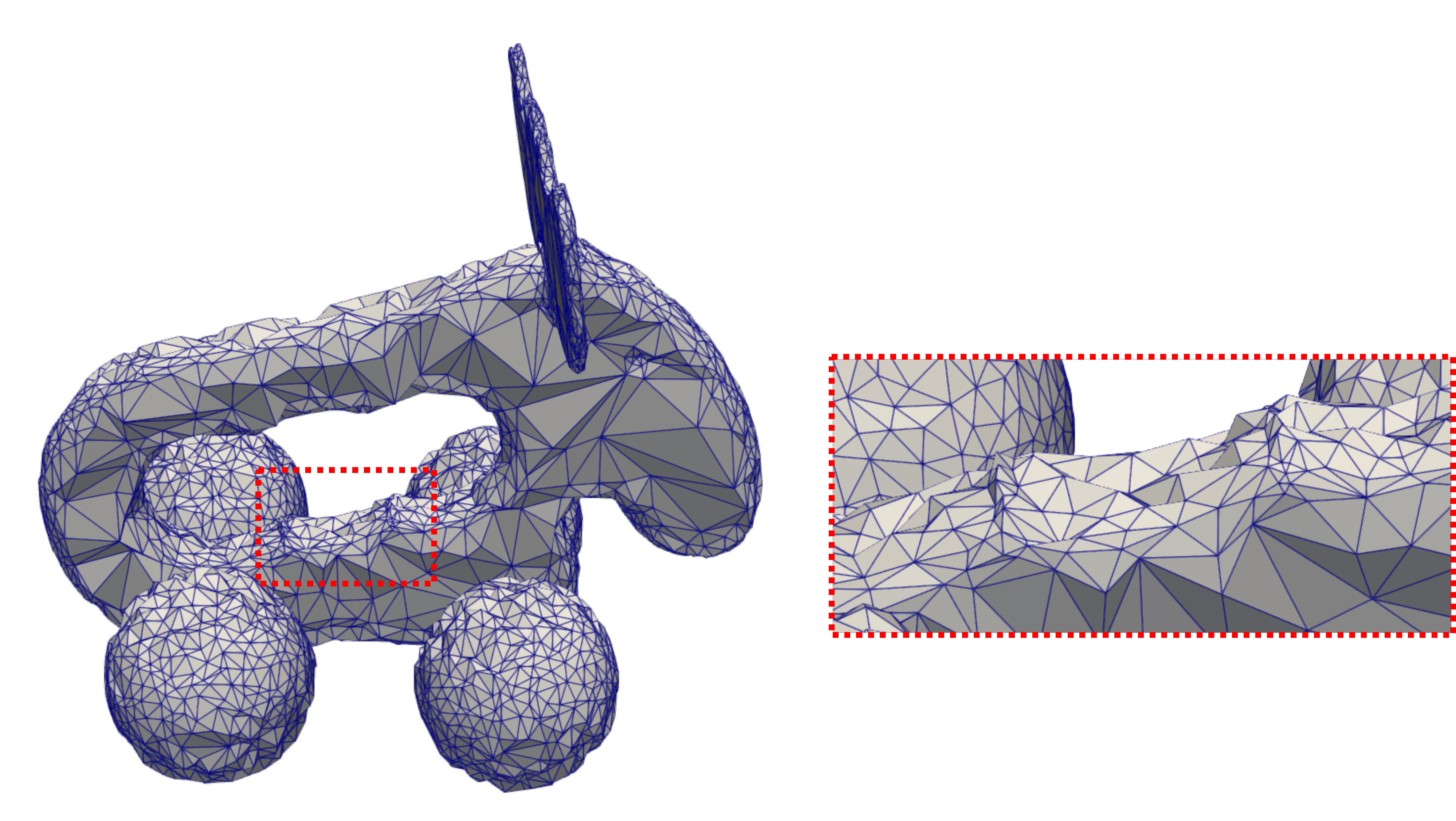}
	\caption{\textit{The surface mesh smoothing results of the agent trained \textbf{without} the surface fitting reward. It shows that, in local areas, the movement of mesh nodes disrupts  the original geometric shape.}}\label{fig:NoSurfaceRewad}
\end{figure}

To extend the proposed method to 3D, we introduced a surface fitting reward \( r_f \) based on local surface fitting in Section~\ref{sec:SurfaceMesh}. Here, we experimentally demonstrate the necessity of this reward for effective surface mesh smoothing. We trained the surface mesh node smoothing agent under the same configuration, both with and without the surface fitting reward. The results of the model without the surface fitting reward are shown in Figure~\ref{fig:NoSurfaceRewad}, illustrating its impact on the mesh smoothing performance. We observe that many mesh nodes deviate significantly from the original geometric surface. The reason is evident: without \( r_f \) as a soft constraint, the agent attempts to modify the coordinates of the mesh nodes in any direction to improve mesh quality, without considering the preservation of the original geometric shape. This outcome underscores the necessity of \( r_f \) in surface mesh smoothing.

\section{Limitations and future work}\label{sec:limit}
Although our proposed smoothing method has achieved significant results, there are several issues that need to be addressed. 
Firstly, the method of adding extra reward terms for  surface mesh smoothing agent essentially serves as a soft constraint. However, due to the inherent uncertainty of neural networks, this approach may not be suitable for applications requiring high mesh quality. Alternative hard constraint methods, such as node projection method, can address this issue but are often complex to implement. Therefore, exploring more effective intelligent approaches to ensure that smoothing algorithms preserve the geometric features of surface meshes is worthwhile.

Secondly, the performance of reinforcement learning-based models is highly influenced by hyperparameters, and even the random seed can affect the model's performance. This issue is a fundamental problem in reinforcement learning and requires further development in its field. However, when training smoothing agents, certain operations such as truncating invalid episodes and improving the network architecture can enhance model stability and accelerate convergence. Further exploration and application of reinforcement learning in the field of mesh smoothing are also needed.

\section{Conclusion}\label{sec:con}
In this paper, we reveal the reliance of existing intelligent smoothing methods on prior data and knowledge from the perspective of reinforcement learning. Building on this, we further integrated graph neural networks with reinforcement learning and propose a mesh node smoothing agent and a connectivity improvement agent. We evaluated the proposed model on both 2D and 3D meshes. The experimental results demonstrate that our model effectively preserves features while smoothing complex 3D surface meshes. Additionally, it outperforms other intelligent smoothing methods for 2D meshes, achieving a speed that is 7.16 times faster than traditional optimization-based smoothing method. Moreover, the connectivity improvement agent significantly enhances the quality distribution of the mesh.
 This work demonstrates the feasibility of reinforcement learning in the field of intelligent mesh smoothing and provides valuable insights for future extensions to surface mesh smoothing and more advanced intelligent mesh connectivity optimization.

\section*{CRediT authorship contribution statement}

\textbf{Zhichao Wang}: Conceptualization, Methodology, Software, Investigation, Visualization, Writing - Original Draft, Writing - Review \& Editing.
\textbf{Xinhai Chen}: Investigation, Visualization, Writing - Review \& Editing, Funding Acquisition.
\textbf{Chunye Gong}: Writing - Review \& Editing.
\textbf{Bo Yang}: Writing - Review \& Editing.
\textbf{Liang Deng}: Writing - Review \& Editing.
\textbf{Yufei Sun}: Writing - Review \& Editing.
\textbf{Yufei Pang}: Writing - Review \& Editing.
\textbf{Jie Liu}: Supervision, Software, Writing - Review \& Editing, Funding Acquisition.

\section*{Acknowledgements}
This research was partially supported by the National Natural Science Foundation of China (12402349), the Natural Science Foundation of Hunan Province (2024JJ6468), the Youth Foundation of the National University of Defense Technology (ZK2023-11), and the National Key Research and Development Program of China (2021YFB0300101).

\section*{Declaration of competing interest}

The author(s) declared no potential conficts of interest with respect to the research, author- ship, and/or publication of this article.

\section*{Data availability}
The data used in this study are available from the corresponding author upon reasonable request.

\bibliographystyle{eg-alpha-doi}
\bibliography{cas-refs}
\onecolumn
\appendix
%\section{Appendix}

\section{The training of GMSNet.}\label{app:GMSNet}
As shown in Alg. \ref{alg:GMSNet}, during each epoch, $\mathcal{B}$ mesh nodes are randomly selected on each mesh for gradient descent, and a \textit{Shift Truncation} operation is employed to ensure the validation of the mesh.
\begin{algorithm}[htbp]
\small
\centering
\caption{\textit{Training of GMSNet}} \label{alg:GMSNet}
\begin{algorithmic}[1]
    \STATE {Mesh dataset $\mathbf{M}=\{{\mathcal{M}_i}\}_{i=1}^n$, MetricLoss $\mathcal{L}$}
    \STATE {Training epochs $N$, batch size $\mathcal{B}$, learning rate $\alpha$}
    \STATE {GMSNet parameters $\mathbf{W}$}
    \FOR{ $j \gets  1 ~\mathrm{to}~ N $ }
    \FOR{mesh $\mathcal{M}_i$ in $\mathbf{M}$}
    \STATE{Sample $\mathcal{B}$ mesh nodes from $\mathcal{M}_i$: \( \{\boldsymbol{x}_i\}_{i=1}^{\mathcal{B}} \)}
    \STATE{Compute the optimized node position: $\boldsymbol{x}_i^*=\mathrm{GMSNet}(\boldsymbol{x}_i, \mathbf{S}(\boldsymbol{x}_i))$ for $\boldsymbol{x}_i
    \in \{\boldsymbol{x}_i\}_{i=1}^{\mathcal{B}}$}
    \STATE{Shift truncate $\Delta \boldsymbol{x}_i$ if $\Delta \boldsymbol{x}_i$ results in negative volume elements}
    \ENDFOR
    \STATE{Update model parameters: $\mathbf{W} \gets $ $\mathbf{W}-\frac{\alpha}{\mathcal{B}}\sum_{i=1}^{\mathcal{B}}\nabla\mathcal{L}(\boldsymbol{x}_i^*, \mathbf{S}(\boldsymbol{x}_i^*))$}
    \ENDFOR
\end{algorithmic}
\end{algorithm}

\section{Model architectures}\label{app:network}
Here we provide the architecture parameters for the node smoothing agent and the connectivity improvement agent in Table~\ref{tab:network}. We use lists to represent the stacked networks. For example, [2, 4, 8] represents an MLP with dimensions 2, 4, and 8. All MLPs use the ReLU activation function.
\begin{table}[H]
\centering
\begin{tabular}{cc}
    \toprule
    \textbf{Model}                      & \textbf{Architecture }                   \\
    \midrule
    \textbf{Node smoothing agent}       &                                          \\
    $\textrm{MLP}_e$                    & [2, 32, 64]                              \\
    $\textrm{GT}$                       & [$\textrm{GT}_1$,  $\textrm{GT}_2$]        \\
    $\textrm{MLP}_d$                    & [64, 64, 64]                             \\
    Actor network                       & [64, 32, 2]                              \\
    Critic network                      & [66, 32, 16, 1]                          \\
    \midrule
    \textbf{Topology improvement agent} &                                          \\
    Mesh encoder                        & The same as that in node smoothing agent \\
    V-Network                           & [64, 32, 1]                              \\
    A-Network                           & [64, 32, 1]                              \\
    \bottomrule
\end{tabular}
\caption{\textit{Network architectures of node smoothing agent and Topology improvement agent.}} \label{tab:network}
\end{table}

\section{Comparison of intelligent mesh smoothing models}\label{app:AllModels}
\begin{figure}[H]
\centering
\subfloat{\includegraphics[width=0.25\columnwidth]{Figs/Legend.pdf}}\\ \setcounter{subfigure}{0}
\subfloat[Case 1 after NN-smoothing]{\includegraphics[height=0.22\columnwidth]{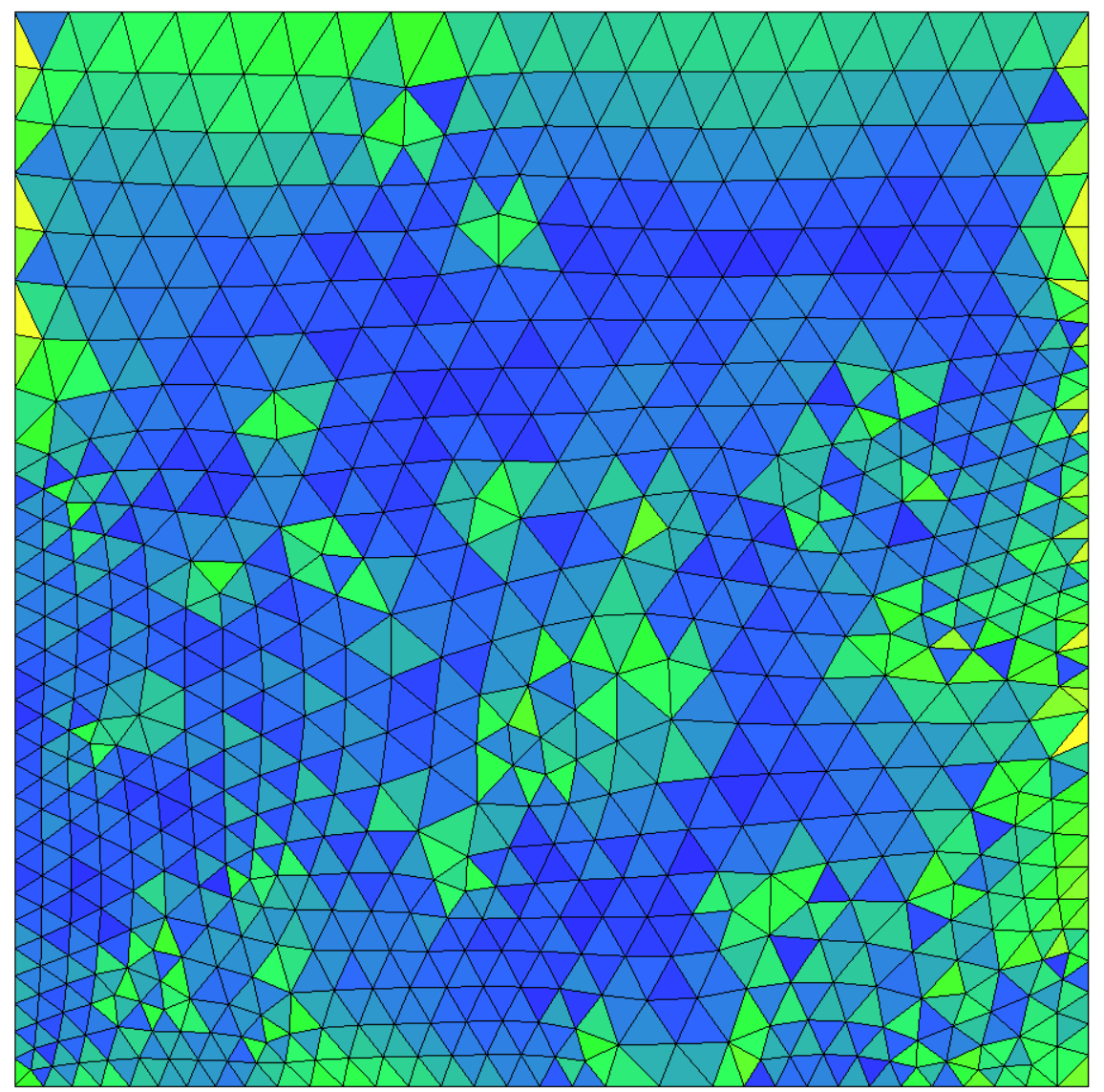}}\quad
\subfloat[Case 1 after GMSNet]{\includegraphics[height=0.22\columnwidth]{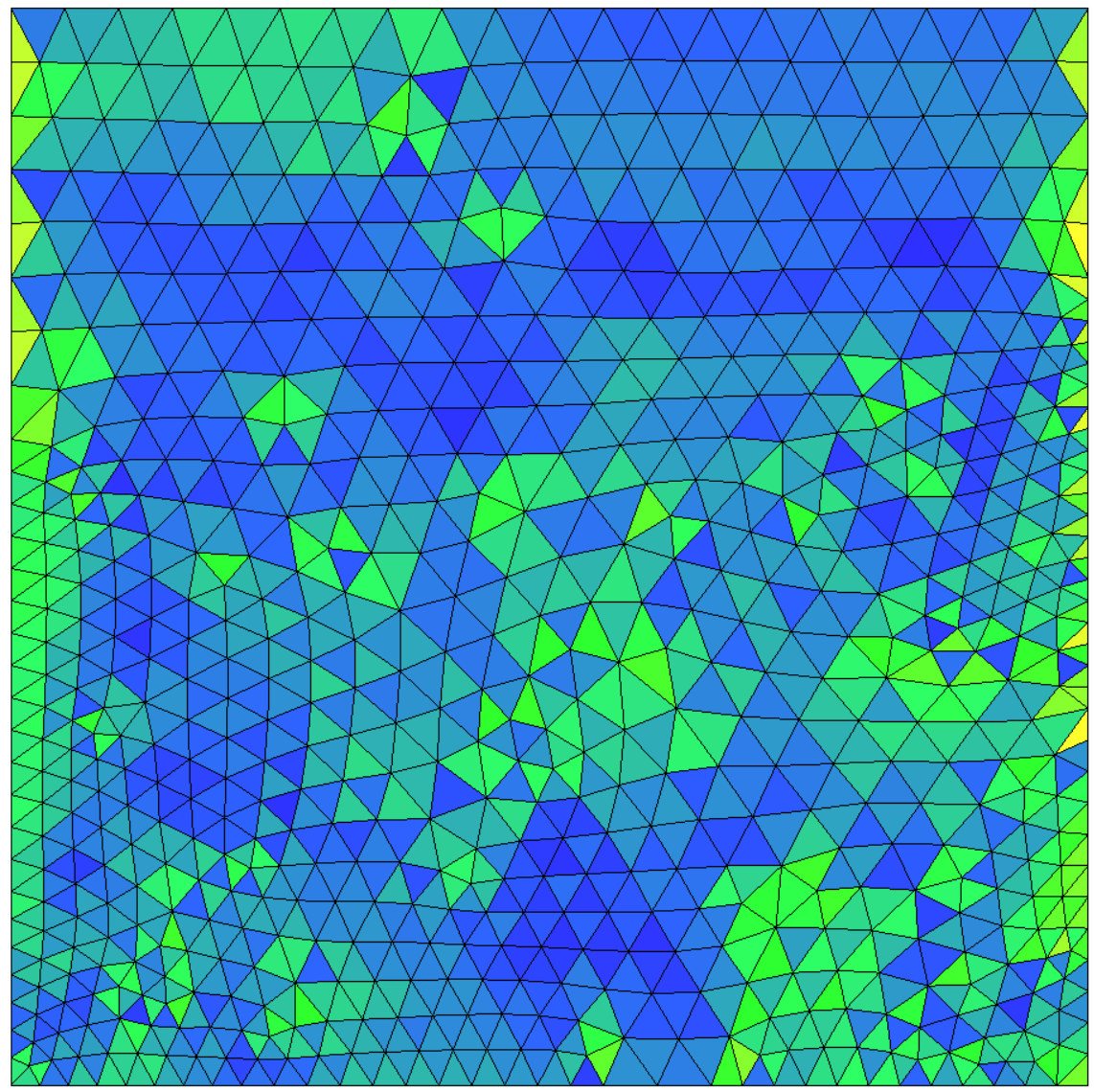} }\quad
\subfloat[Case 1 after DRL-Smoothing]{\includegraphics[height=0.22\columnwidth]{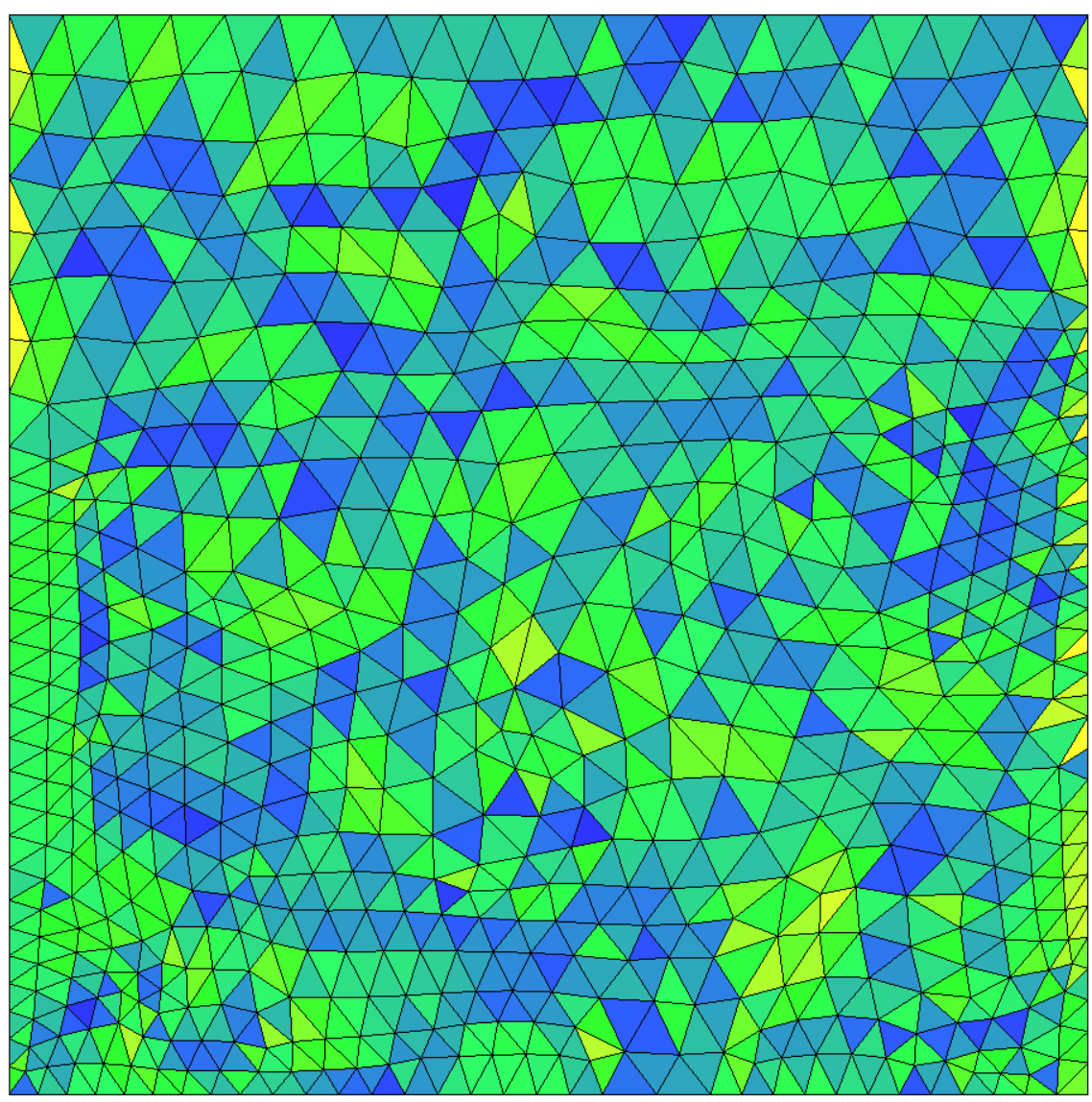} \label{fig:Case1DRL}}\quad
\subfloat[Case 1 after GNNRL-Smoothing]{\includegraphics[height=0.22\columnwidth]{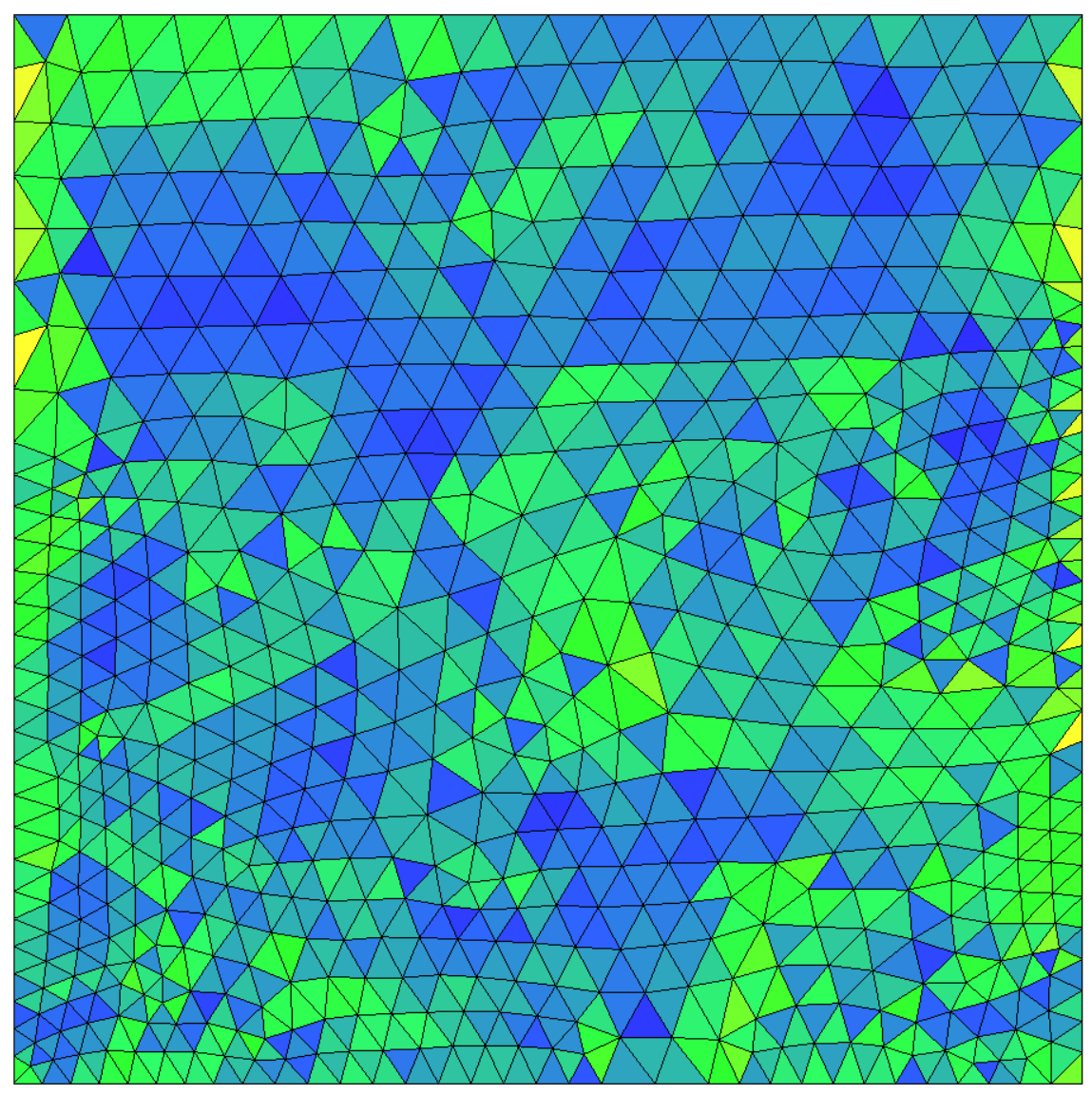}}\quad \\

\subfloat[Case 2 after NN-smoothing]{\includegraphics[width=0.22\columnwidth]{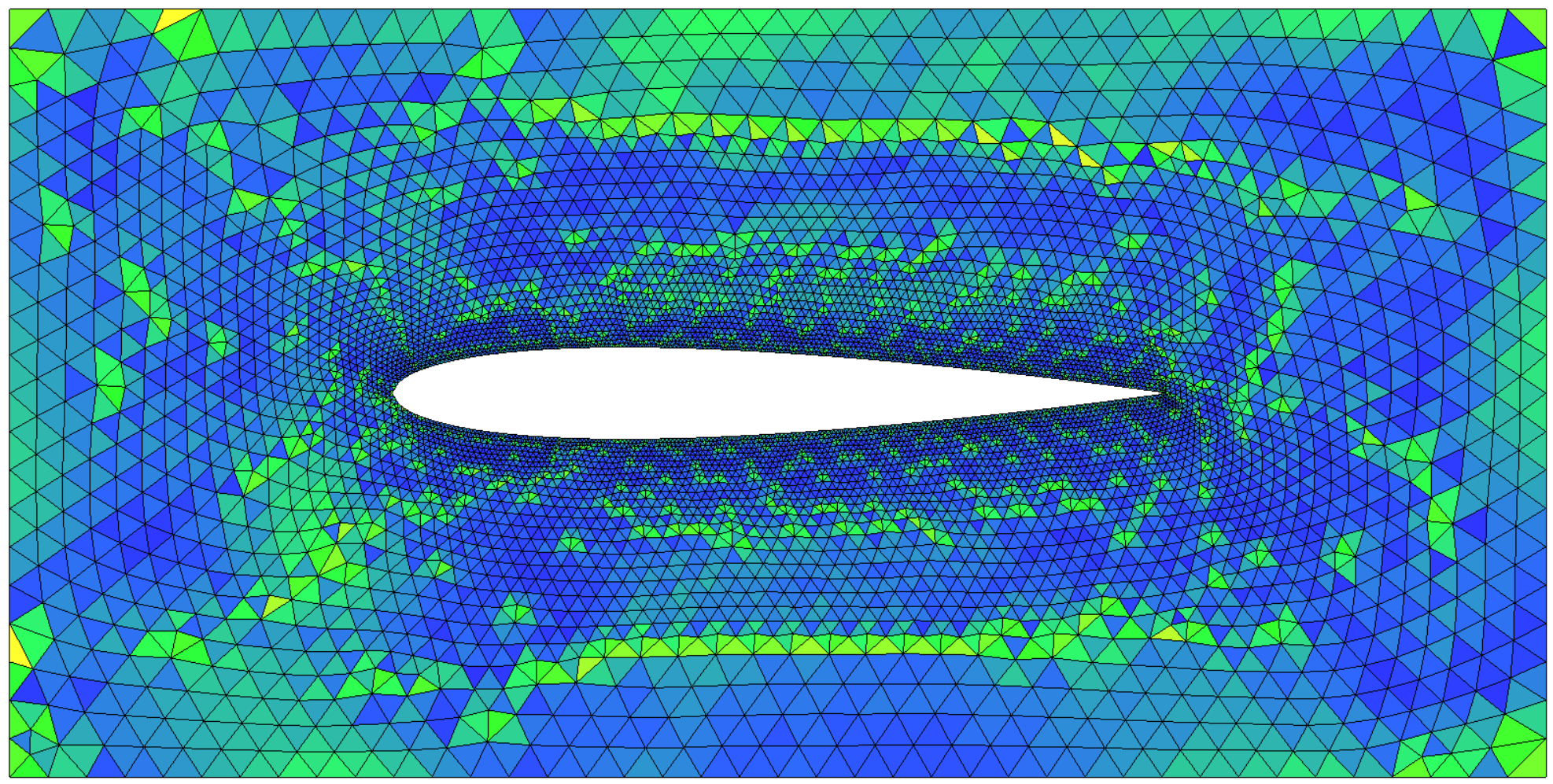}}\quad
\subfloat[Case 2 after GMSNet]{\includegraphics[width=0.22\columnwidth]{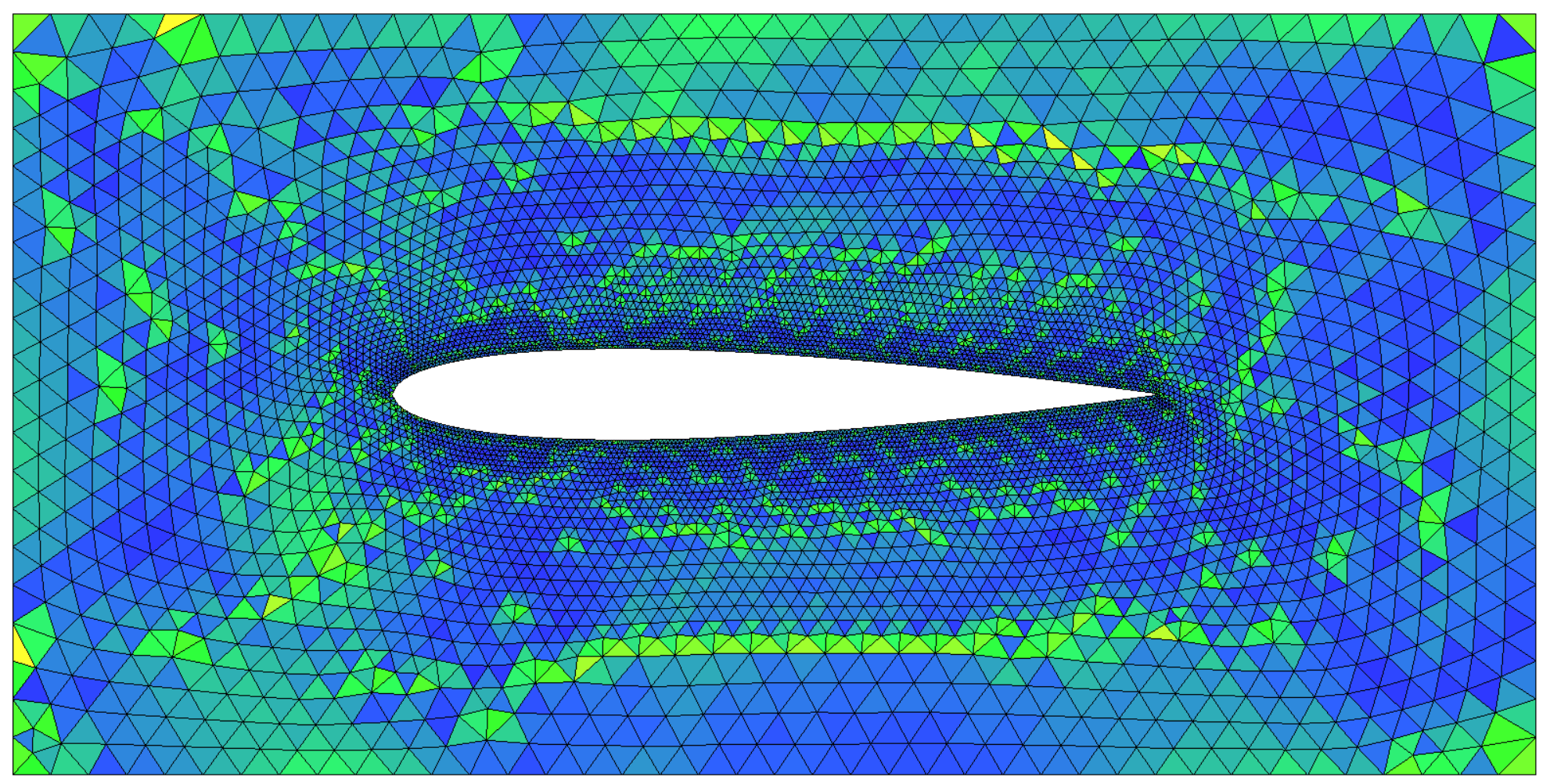}}\quad
\subfloat[Case 2 after DRL-Smoothing]{\includegraphics[width=0.22\columnwidth]{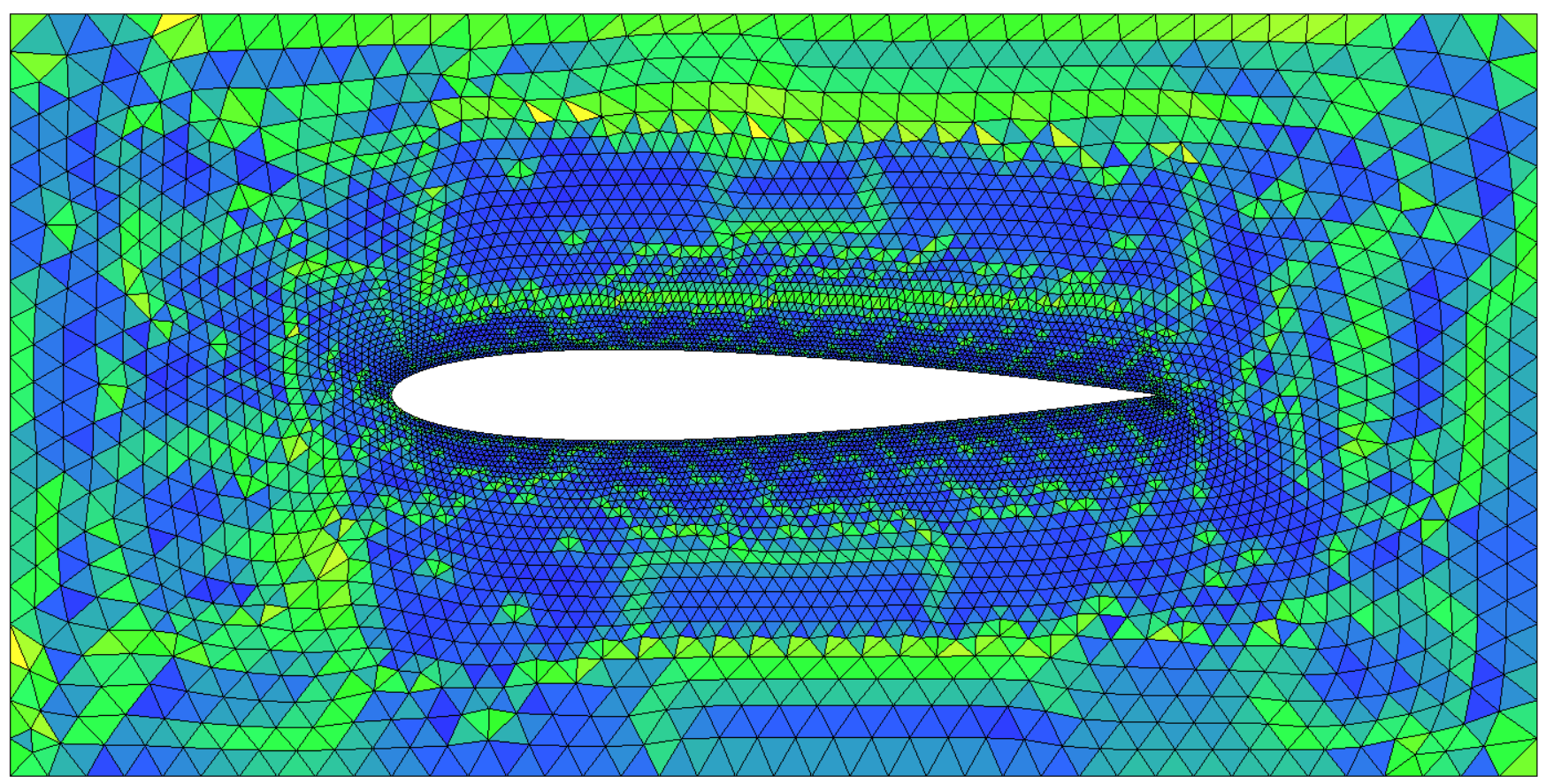} \label{fig:Case2DRL}}\quad
\subfloat[Case 2 after GNNRL-Smoothing]{\includegraphics[width=0.22\columnwidth]{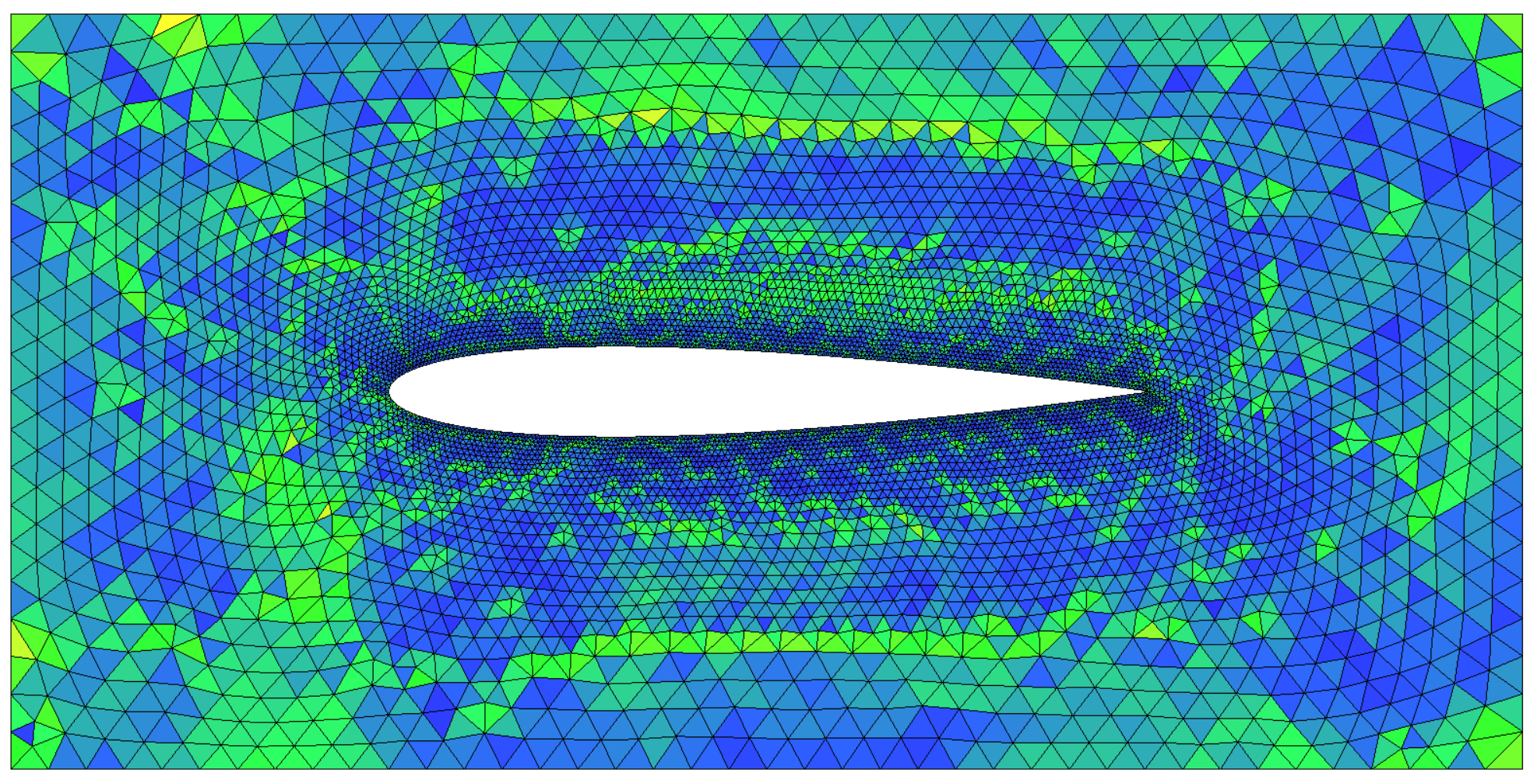}}\quad \\

\subfloat[Case 3 after NN-smoothing]{\includegraphics[height=0.22\columnwidth]{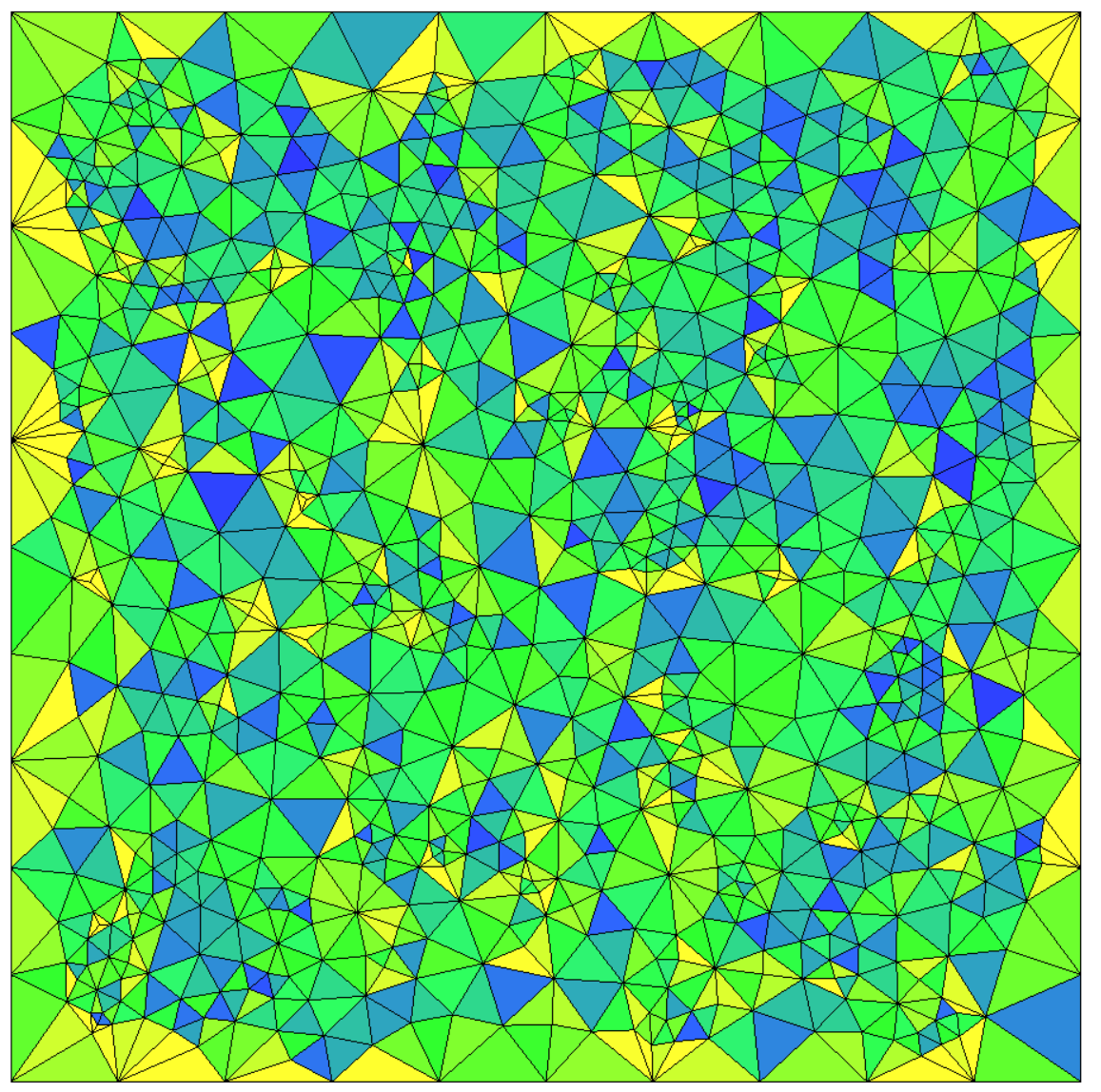}}\quad
\subfloat[Case 3 after GMSNet]{\includegraphics[height=0.22\columnwidth]{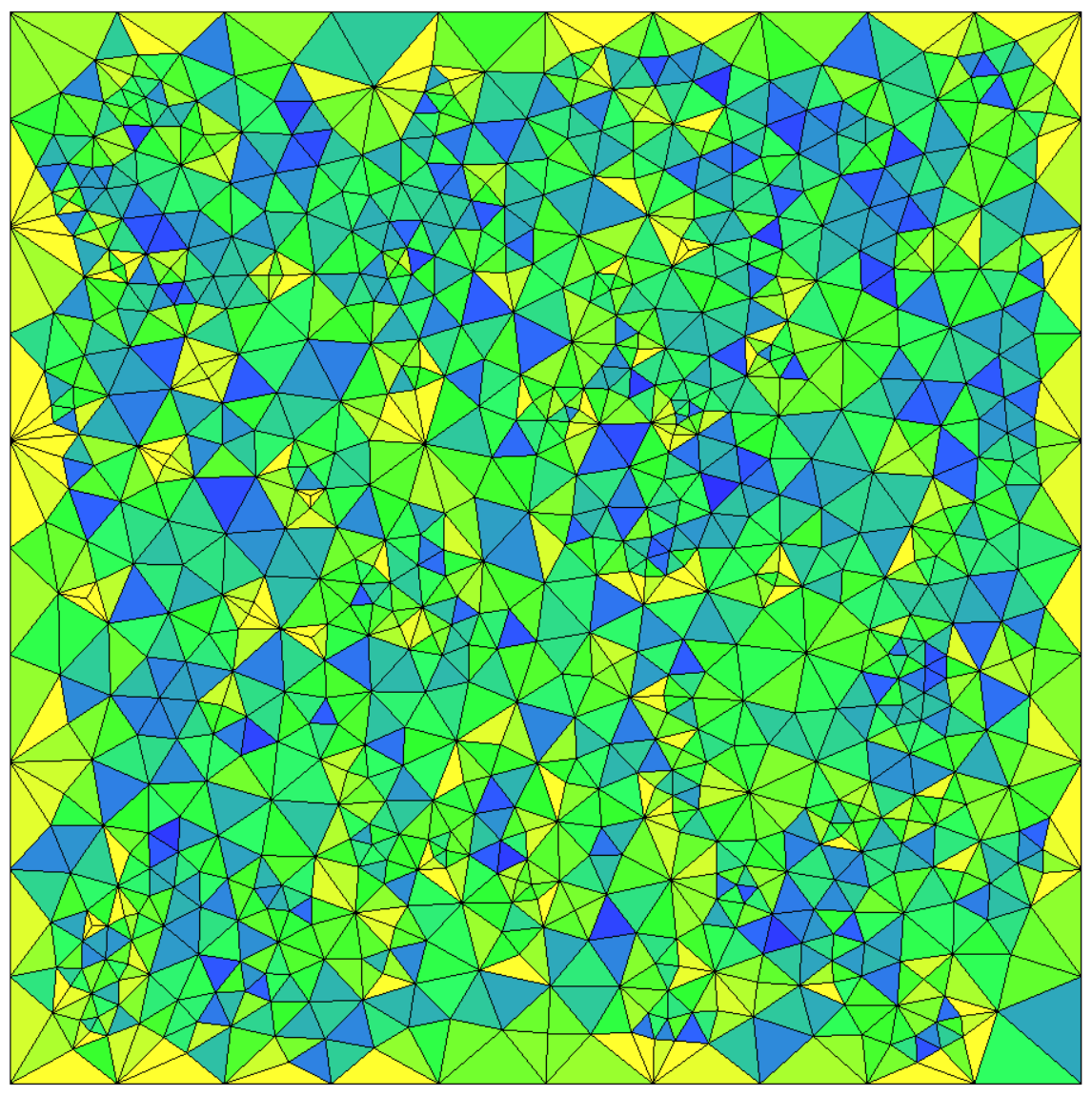}}\quad
\subfloat[Case 3 after DRL-Smoothing]{\includegraphics[height=0.22\columnwidth]{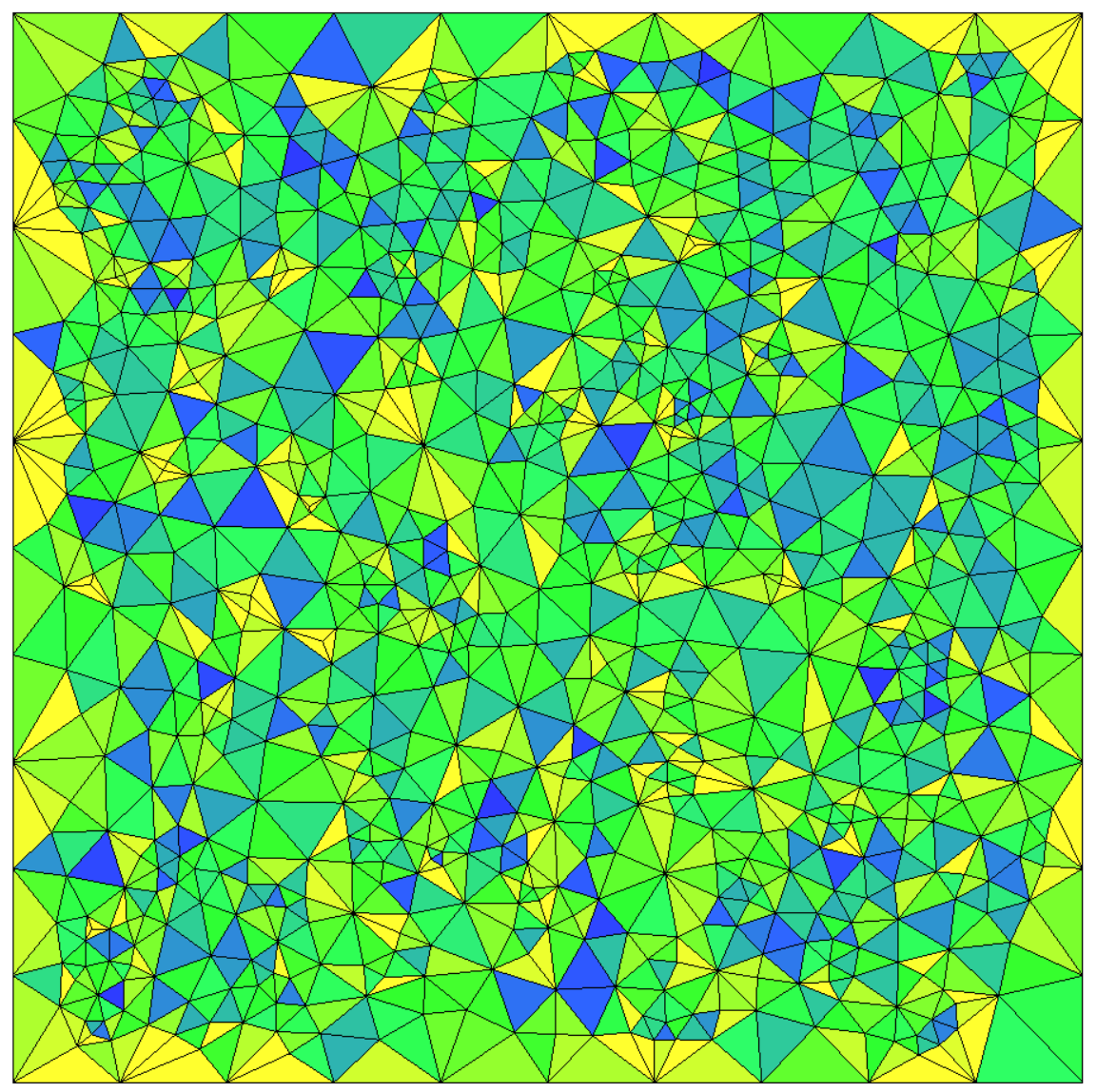}}\quad
\subfloat[Case 3 after GNNRL-Smoothing]{\includegraphics[height=0.22\columnwidth]{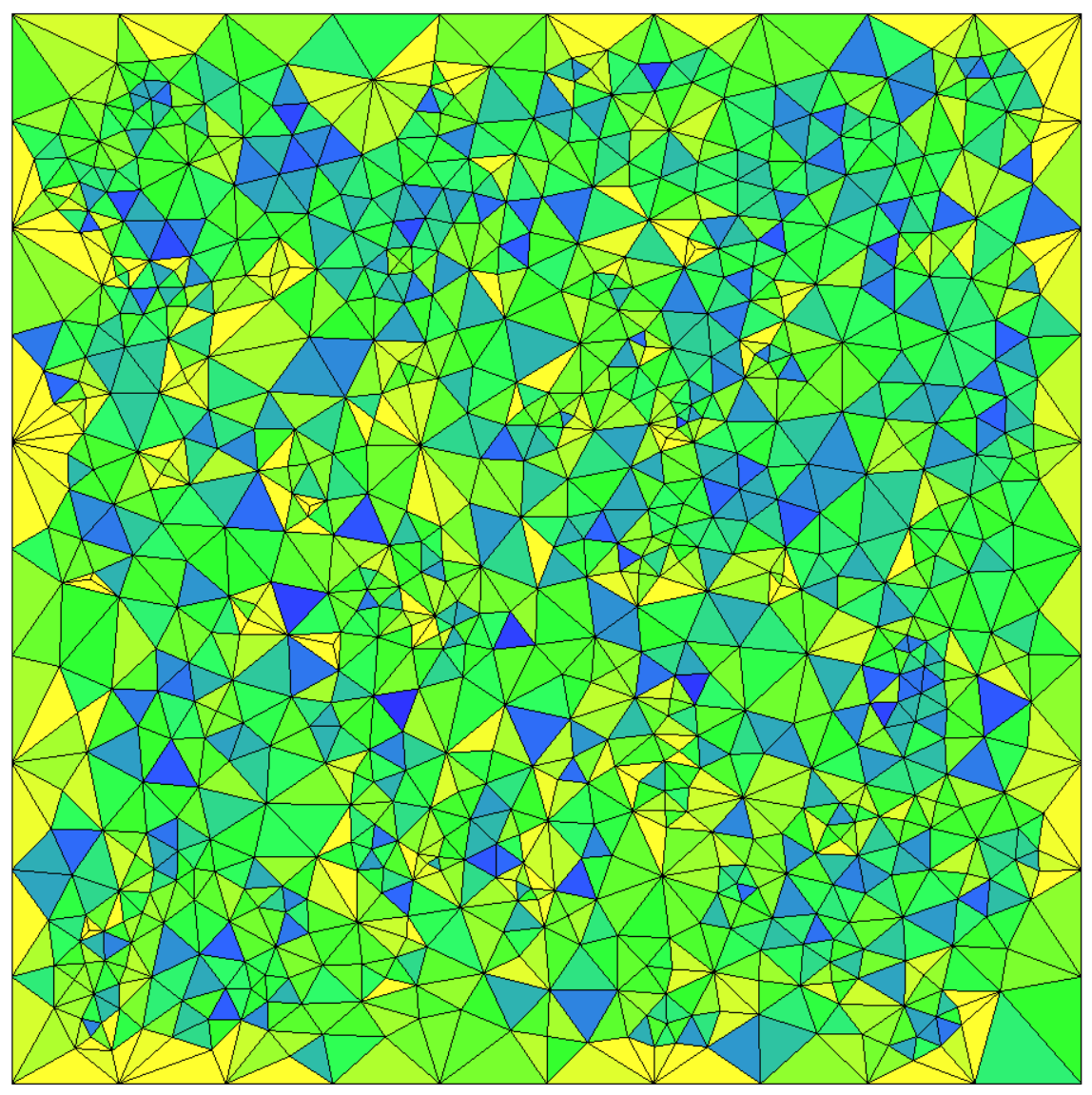}}\quad \\

\subfloat[Case 4 after NN-smoothing]{\includegraphics[width=0.22\columnwidth]{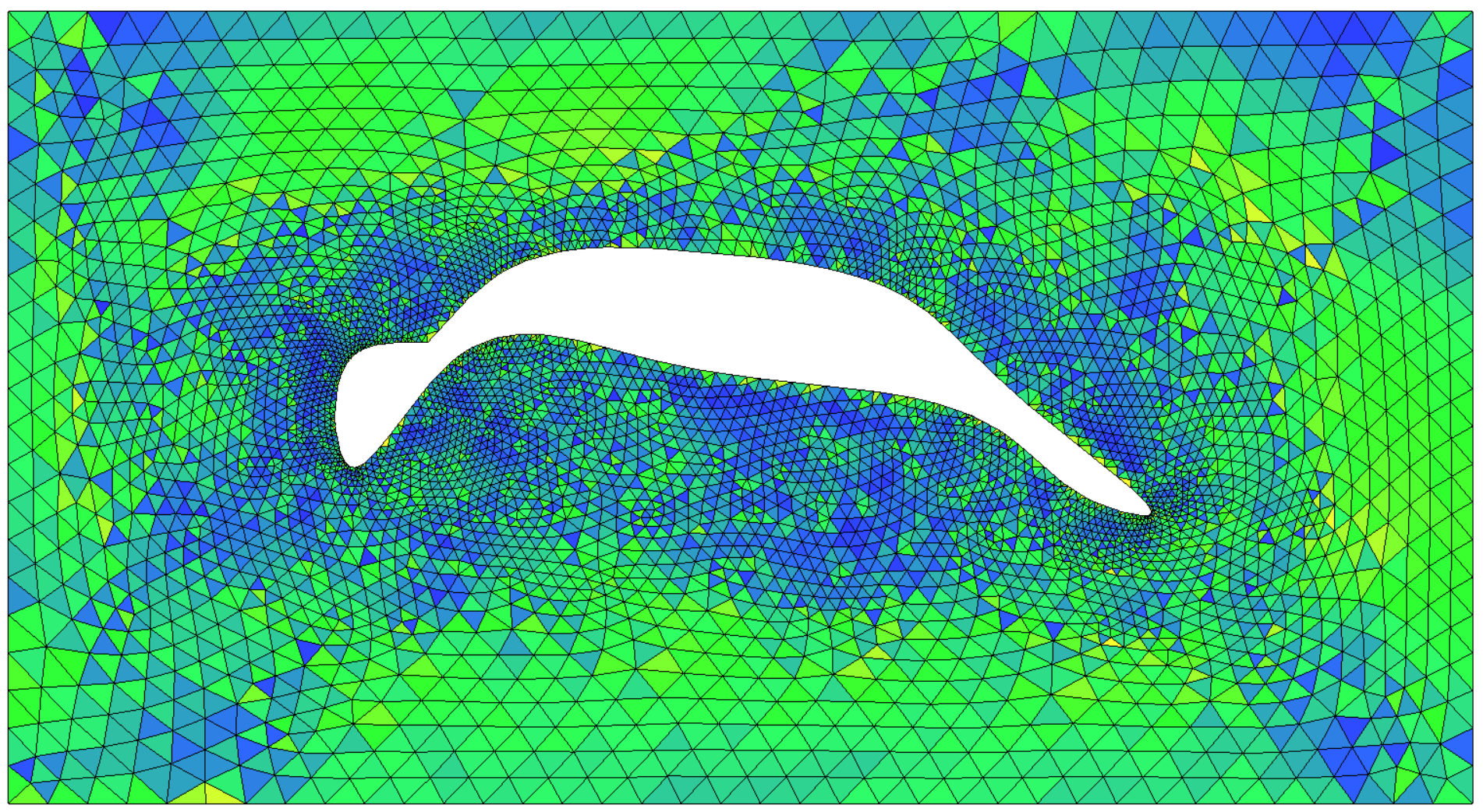}}\quad
\subfloat[Case 4 after GMSNet]{\includegraphics[width=0.22\columnwidth]{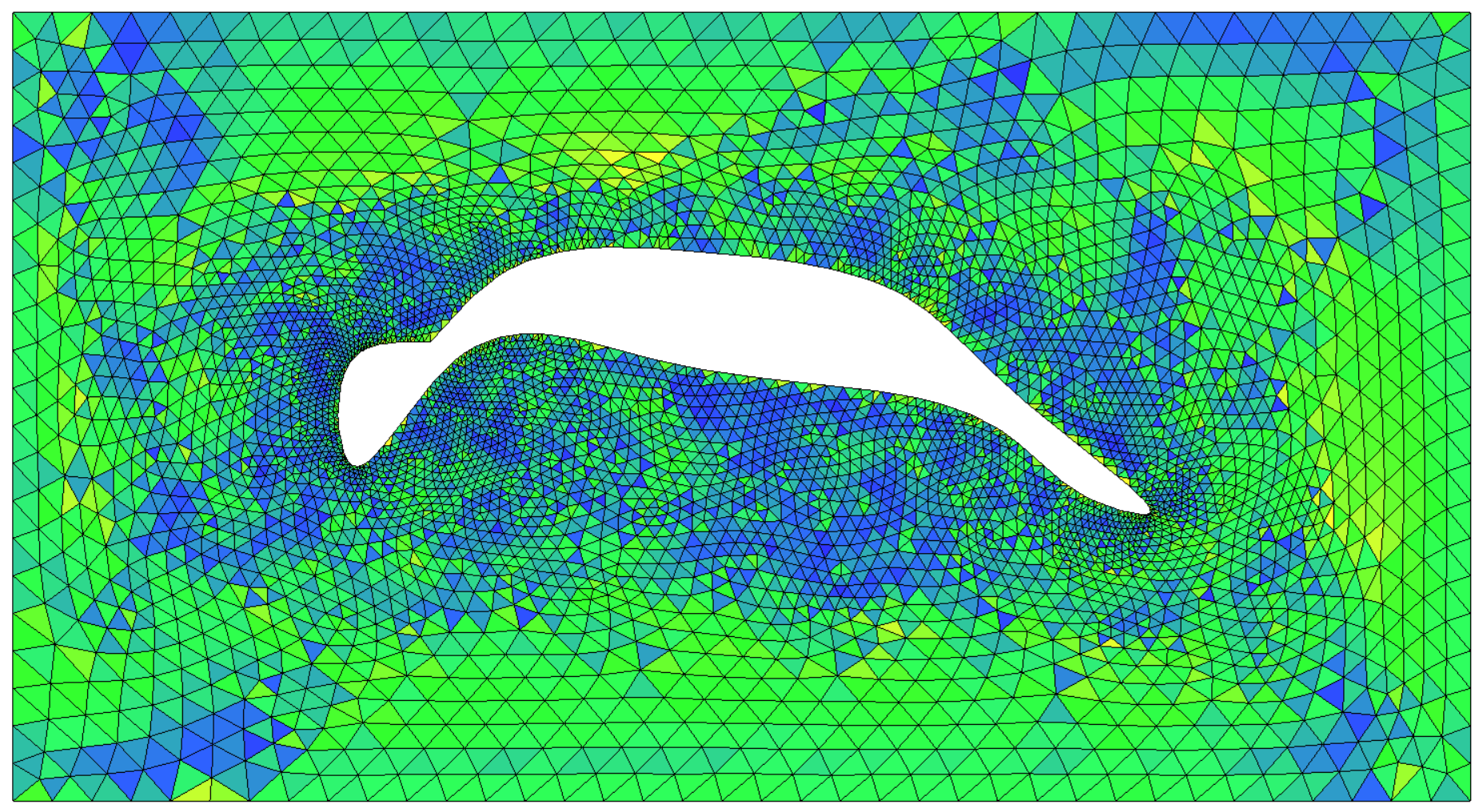}}\quad
\subfloat[Case 4 after DRL-Smoothing]{\includegraphics[width=0.22\columnwidth]{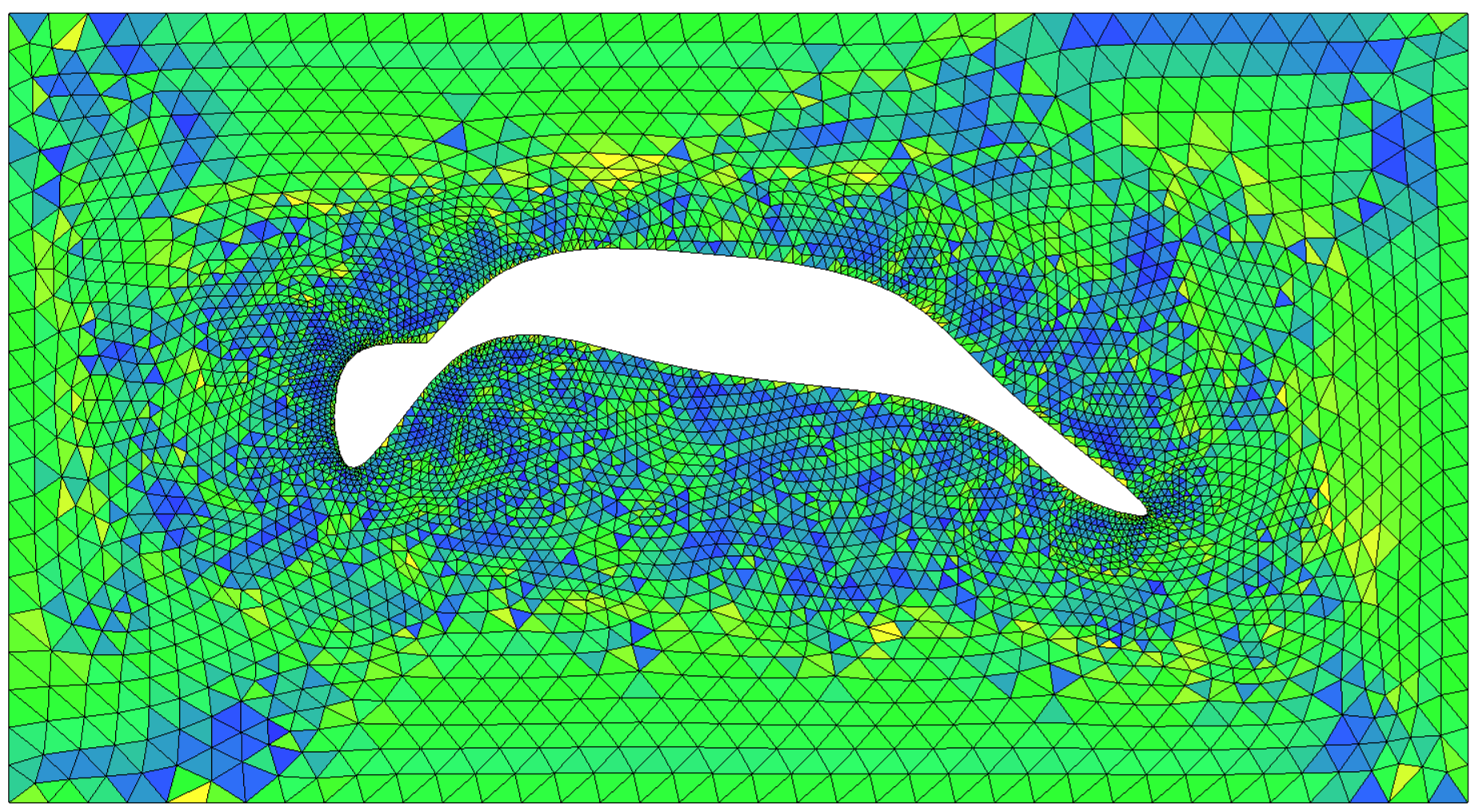}}\quad
\subfloat[Case 4 after GNNRL-Smoothing]{\includegraphics[width=0.22\columnwidth]{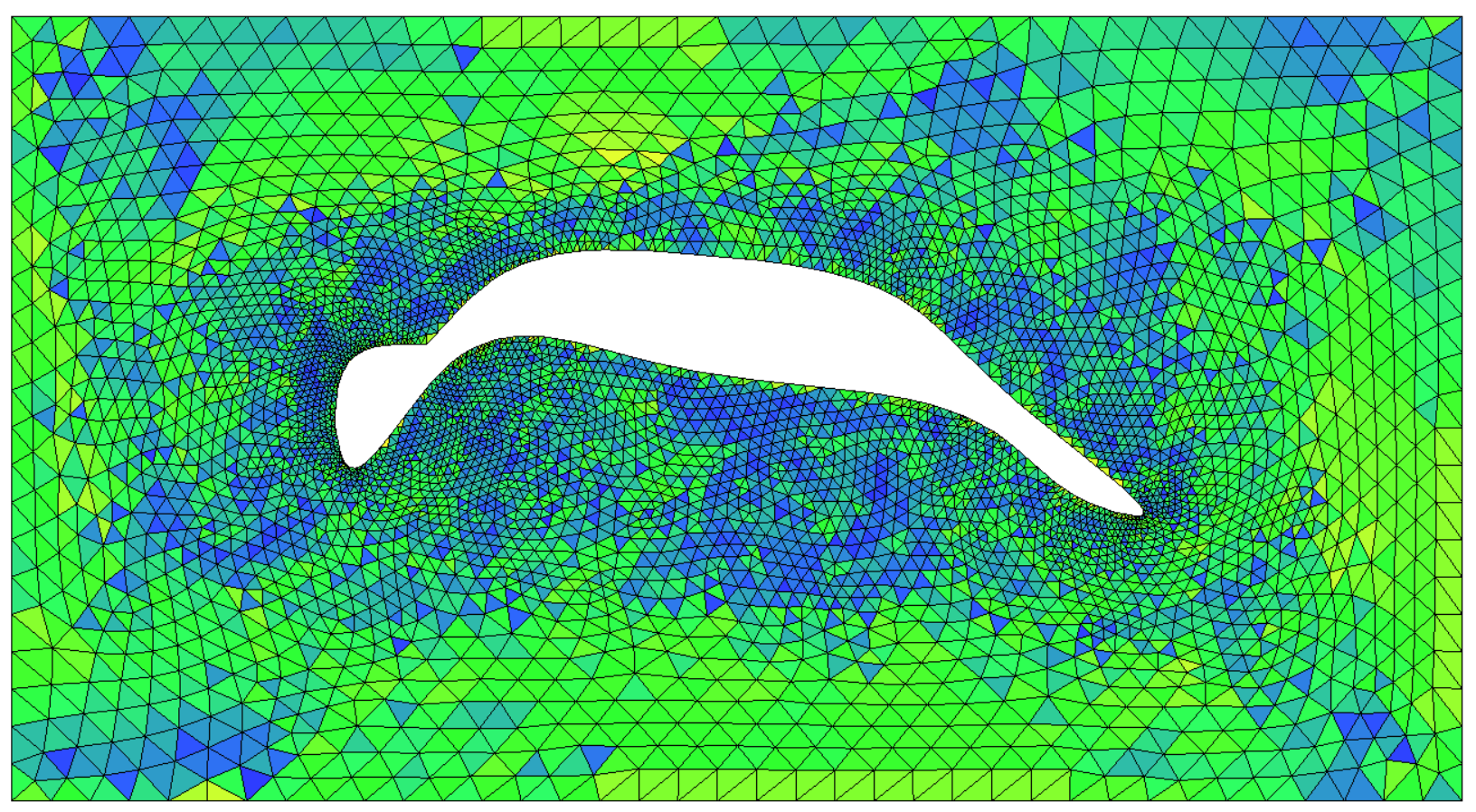}}\quad \\
\caption{\textit{Results of intelligent mesh smoothing models on the test cases. }}
\label{fig:AllModels}
\end{figure}

In Figure~\ref{fig:AllModels}, we compare the smoothing performance of intelligent mesh smoothing methods on test cases. In each case, the intelligent smoothing methods significantly improve the quality of the mesh elements. The NN-Smoothing, GMSNet, and GNNRL-Smoothing models achieve similar smoothing results, whereas the mesh elements smoothed by DRL-Smoothing exhibit some distortion, as shown in Figure~\ref{fig:Case1DRL} and \ref{fig:Case2DRL}.

\clearpage

\end{document}